\documentclass[sigconf]{acmart}

\usepackage{preamble}

\usepackage{hyperref}       
\usepackage{url}            
\usepackage{booktabs}       
\usepackage{amsfonts}       
\usepackage{nicefrac}       
\usepackage{microtype}      
\usepackage{xcolor}         
\usepackage[capitalize,noabbrev]{cleveref}
\crefname{algocf}{Algorithm}{Algorithms}
\Crefname{algocf}{Algorithm}{Algorithms}
\usepackage{multirow}
\usepackage{graphicx}
\usepackage{subcaption}
\usepackage{wrapfig}
\usepackage{enumitem}
\usepackage{adjustbox}
\usepackage[version=4]{mhchem}
\usepackage[linesnumbered,ruled,vlined]{algorithm2e}
\usepackage{bbding}
\usepackage{float}

\def\SM{Appendix}

\AtBeginDocument{%
  }

\setcopyright{acmlicensed}
\copyrightyear{2025}
\acmYear{2025}
\setcctype{by}
\acmConference[MM '25]{Proceedings of the 33rd ACM International Conference on Multimedia}{October 27--31, 2025}{Dublin, Ireland}
\acmBooktitle{Proceedings of the 33rd ACM International Conference on Multimedia (MM '25), October 27--31, 2025, Dublin, Ireland}
\acmDOI{10.1145/3746027.3755729}
\acmISBN{979-8-4007-2035-2/2025/10}

\begin{document}

\title{MIPS: a Multimodal Infinite Polymer Sequence Pre-training Framework for Polymer Property Prediction}


\author{Jiaxi Wang}
\affiliation{%
  \institution{Department of Electronic Engineering, Tsinghua University}
  \city{Beijing}
  \country{China}}
\email{wjx20@mails.tsinghua.edu.cn}

\author{Yaosen Min}
\affiliation{%
  \institution{Zhongguancun Institute of Artificial Intelligence}
  \city{Beijing}
  \country{China}}
\email{yaosenmin@zgci.ac.cn}

\author{Xun Zhu}
\affiliation{%
  \institution{Department of Electronic Engineering, Tsinghua University}
  \city{Beijing}
  \country{China}}
\email{zhu-x24@mails.tsinghua.edu.cn}

\author{Miao Li}
\authornote{Corresponding authors.}
\affiliation{%
  \institution{Department of Electronic Engineering, Tsinghua University}
  \city{Beijing}
  \country{China}}
\email{miao-li@tsinghua.edu.cn}

\author{Ji Wu}
\authornotemark[1]
\affiliation{%
  \institution{Department of Electronic Engineering \& College of AI, Tsinghua University \\
  Beijing National Research Center for Information Science and Technology}
  \city{Beijing}
  \country{China}}
\email{wuji\_ee@tsinghua.edu.cn}





\renewcommand{\shortauthors}{Trovato et al.}

\begin{abstract}
Polymers, composed of repeating structural units called monomers, 
are fundamental materials with a wide range of applications in daily life and industry. 
Accurate property prediction for polymers is essential for their design, development, and application. 
However, existing modeling approaches, which typically represent polymers by the constituent monomers, 
struggle to capture the whole properties of polymer, since the properties change during the polymerization process. 
In this study, we propose a Multimodal Infinite Polymer Sequence (MIPS) pre-training framework, 
which represents polymers as infinite sequences of monomers and 
integrates both topological and spatial information for comprehensive modeling. 
From the topological perspective, 
we generalize message passing mechanism (MPM) and graph attention mechanism (GAM) to infinite polymer sequences. 
For MPM, we demonstrate that applying MPM to infinite polymer sequences is equivalent to 
applying MPM on the induced star-linking graph of monomers. 
For GAM, we propose to further replace global graph attention with localized graph attention (LGA). 
Moreover, we show the robustness of the "star linking" strategy through an adversarial evaluation method 
named Repeat and Shift Invariance Test (RSIT). 
Despite its robustness, "star linking" strategy exhibits limitations when monomer side chains contain ring structures, 
a common characteristic of polymers, as it fails the Weisfeiler-Lehman~(WL) test.
To overcome this issue, we propose backbone embedding to enhance the capability of MPM and LGA on infinite polymer sequences. 
From the spatial perspective, we extract 3D descriptors of repeating monomers to capture spatial information. 
Finally, we design a cross-modal fusion mechanism to unify the topological and spatial information. 
Experimental validation across eight diverse polymer property prediction tasks reveals that MIPS achieves state-of-the-art performance. 
Ablation studies further comfirm the efficacy of our infinite polymer sequence modeling approach and multimodal pre-training framework.\footnote{Our code is available at \url{https://github.com/wjxts/MIPS}}
\end{abstract}

\ccsdesc[500]{Applied computing~Bioinformatics}
\ccsdesc[500]{Computing methodologies~Unsupervised learning}

\keywords{Multimodal Pre-training, Polymer Property Prediction, Infinite Polymer Sequence Modeling}



\maketitle


\section{Introduction}
\begin{figure*}[t]
  \captionsetup[subfigure]{justification=centering}
  \centering
  \includegraphics[width=\textwidth]{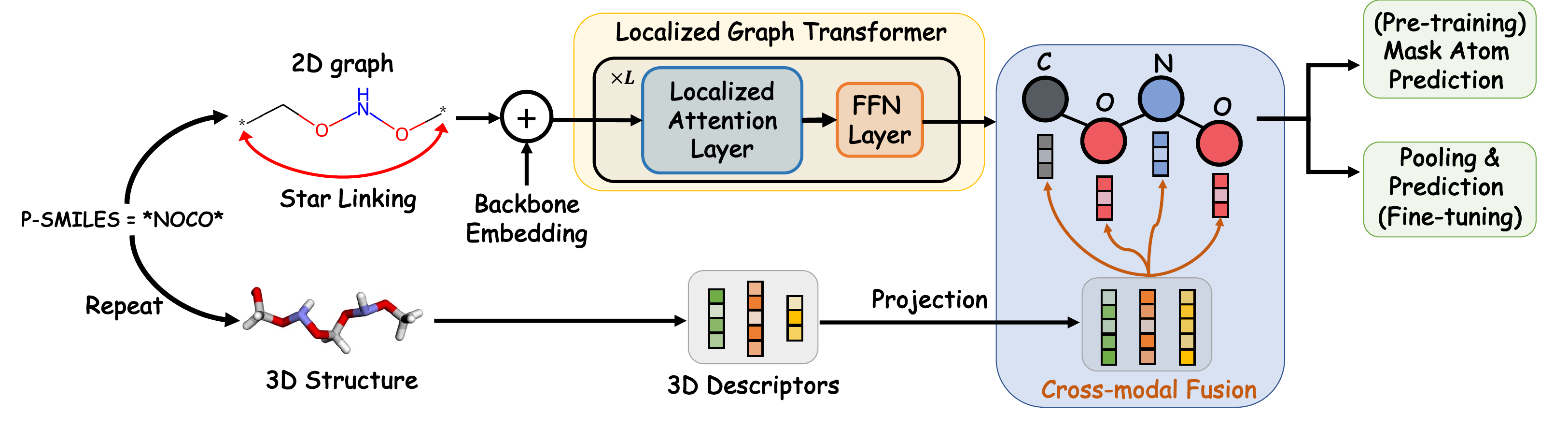}
  \caption{Pipeline of MIPS pre-training framework.}
  \label{fig:pipeline}
\end{figure*}
Polymers constitute a fundamental class of materials widely utilized in daily life~\cite{Namazi2017PolymerLife} 
and industrial applications, such as energy storage~\cite{Luo2018PolymerEnergy1}, drug design~\cite{Pasut2007PolymerDrug}, and optoelectronics~\cite{Munshi2021PolymerOptic}, 
due to their versatile properties and functionalities. 
Precisely determining the properties of polymers is crucial for their design, development, and deployment, 
as it allows researchers to optimize performance for targeted applications. 
While conventional approaches to obtaining polymer properties, including experiments and simulations, 
provide accurate results, they typically remain time-consuming and resource-intensive.
In contrast, machine learning approaches have emerged as a promising alternatives, 
significantly reducing time and cost, and therefore facilitating efficient screening of extensive polymer libraries for specific applications~\cite{Le2012QSPR}.
These approaches compute polymer descriptors as input features for traditional machine learning models.
Inspired by the success of deep learning in natural language processing and computer vision~\cite{LeCun2015DL}, 
researchers apply neural networks to automatically learn features from raw polymer data~\cite{Rahman2021AML,Simine2020PredictingOS}. 
In practice, the amount of labeled data are often limited. 
Self-supervised learning (SSL) methods have been further developed to leverage the unlabeled data 
which improeves the generalization ability~\cite{Kuenneth2022polyBERT,Wang2024MMPolymer,Pei2023PPP}. 
Despite these advancements, given the inherent complexity of polymers due to structural variations and the impact of the polymerization process, 
developing effective and reliable modeling approaches remains a significant challenge.
Existing polymer modeling methods primarily rely on monomer representations~(P-SMILES\cite{Kuenneth2022polyBERT}). 
However, due to the polymerization process, the properties of polymers differ from those of their constituent monomers~\cite{Odian2004Polymer,Fried2014Polymer}. 
In this work, we present a Multimodal Infinite Polymer Sequence (MIPS) pre-training framework
which models polymers as infinite sequences of monomers and 
effectively integrates both topological and spatial information via a cross-modal fusion mechanism. 
Our contributions can be summarized as follows: 
\begin{itemize}
\item 
We propose to model polymers as infinite sequence of monomers, 
and accordingly extend message passing mechanism (MPM) and graph attention mechanism (GAM). 
We extend MPM to infinite polymer sequences by "star linking" strategy and further replace GAM with localized graph attention (LGA). 
\item 
We introduce Repeat and Shift Invariance Test (RSIT) to access the robustness of polymer modeling methods. 
Experimental results demonstrate that existing polymer modeling methods, including "star keep", "star remove", and "star substitution", 
suffer significant performance degradation under RSIT. 
In contrast, our "star linking" is resistant to RSIT. 
\item  
We theoretically prove that WL test, as well as the MPM and LGA, can not distinguish the defined twin polymer graphs. 
We propose backbone embedding to improve the expressive power of MPM and LGA in distinguishing twin polymer graphs. 
\item 
Our proposed MIPS framework effectively integrates topological and spatial polymer features.
Extensive evaluation on eight polymer property prediction tasks demonstrate that MIPS outperforms existing methods, achieving state-of-the-art performance. 
Comprehensive ablation and interpretation analyses reveal that MIPS can learn meaningful polymer representations. 
\end{itemize}



\section{Related Work}
\begin{figure*}[t]
  \captionsetup[subfigure]{justification=centering}
  \centering
  \begin{subfigure}[b]{0.96\columnwidth}
    \centering
    \centerline{\includegraphics[width=\textwidth]{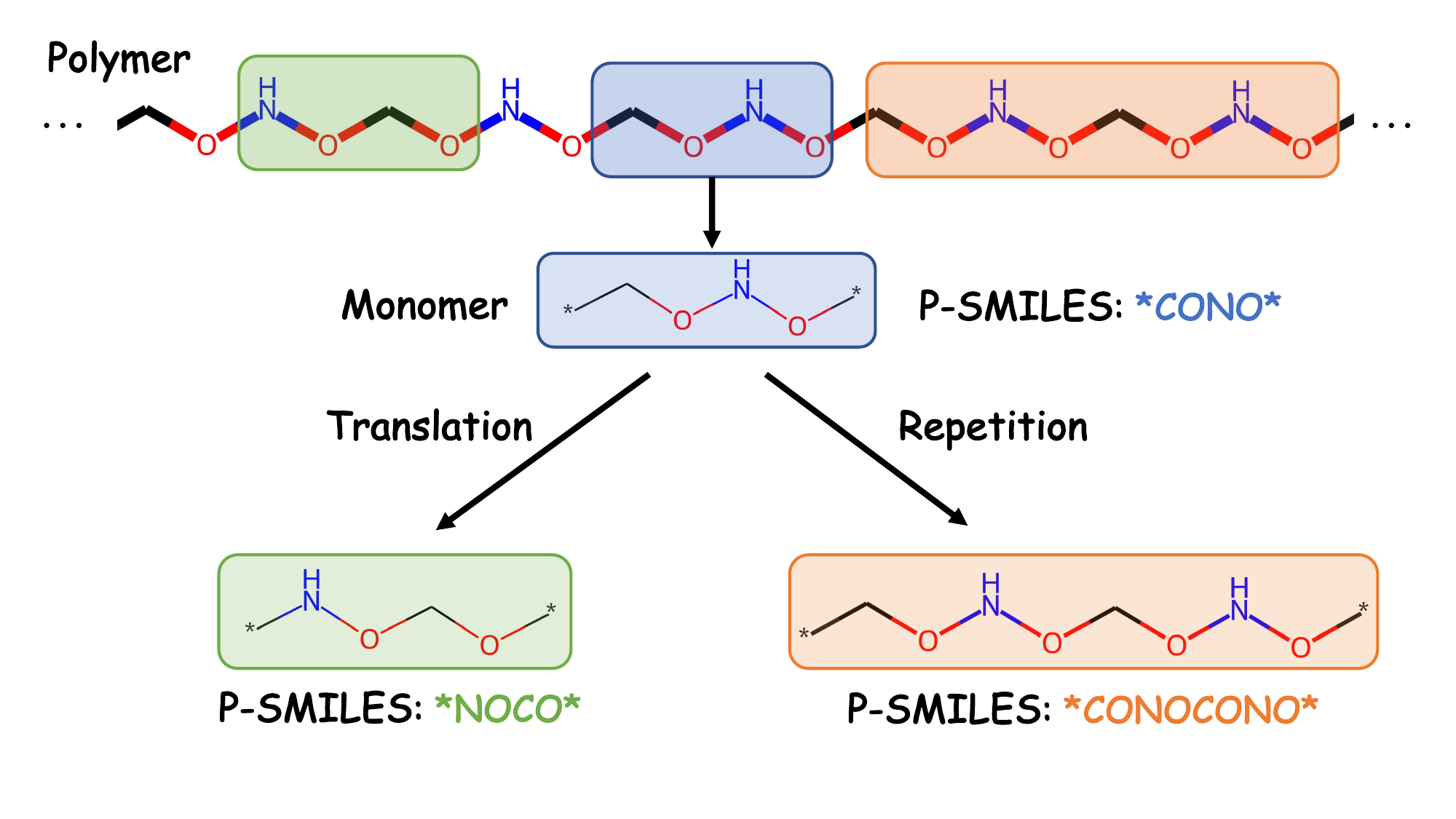}}
  \end{subfigure}
  \hfill
  \begin{subfigure}[b]{0.96\columnwidth}
    \centering
    \includegraphics[width=\textwidth]{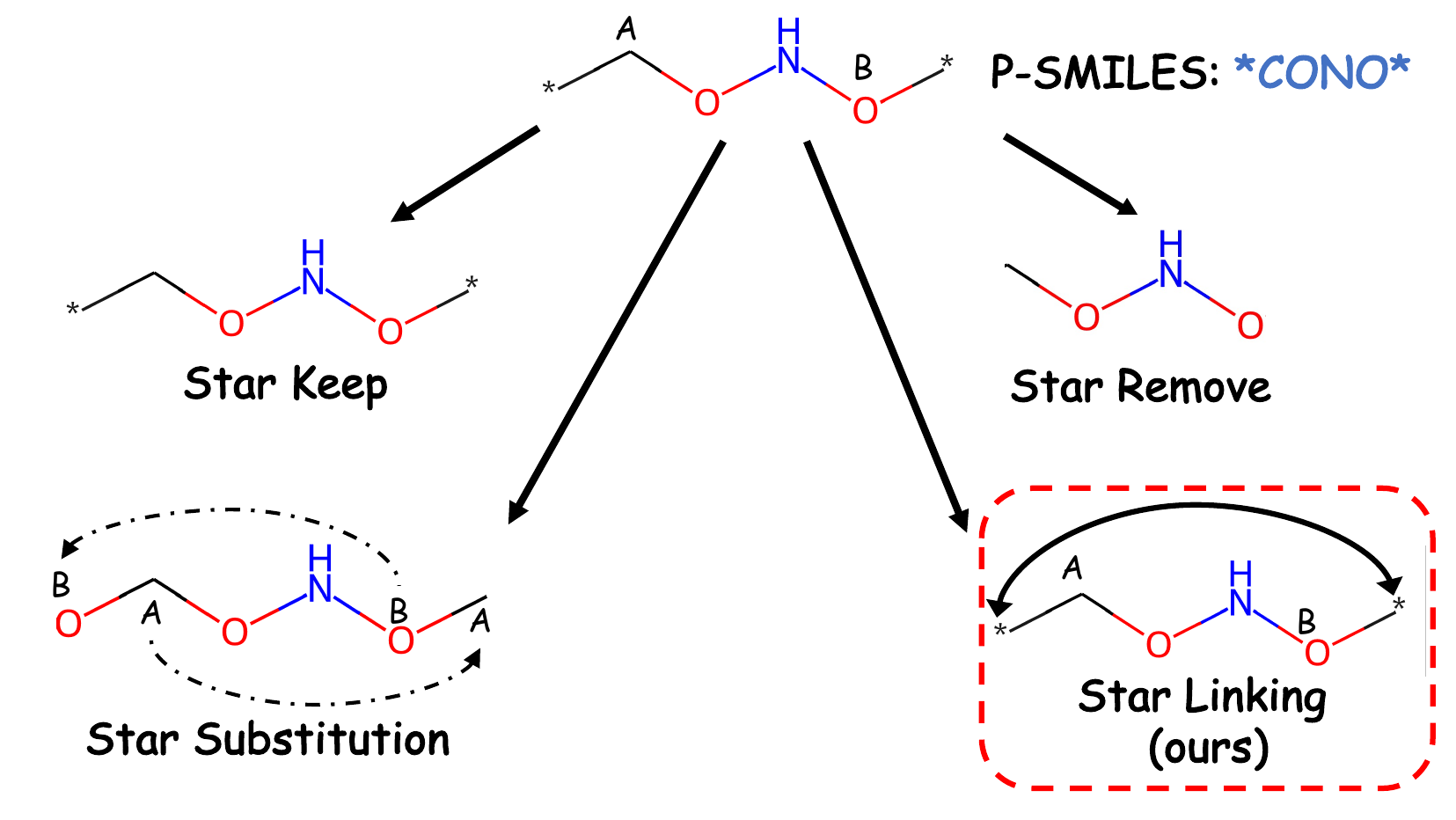}
  \end{subfigure}
  \caption{Left: How polymers are represented by P-SMILES, along with examples of translation and repetition.
  The underlying polymer of P-SMILES is unchanged under translation and repetition. 
  Right: Three existing polymer modeling methods: "star keep", "star remove", and "star substitution". 
  Our proposed "star linking" strategy is specifically designed to model infinite polymer sequences.}
  \label{fig:preliminary}
\end{figure*}
\subsection{Polymer Representation Learning}
Acquiring effective polymer representation is essential for accurately predicting polymer properties~\cite{Xu2022TransPolymer,Kuenneth2022polyBERT}. 
Classic quantitative structure-property relationship (QSPR) approaches~\cite{Le2012QSPR} rely on handcrafted features, 
and utilize traditional machine learning models, such as 
support vector machines~\cite{Cortes1995SVM}, random forests~\cite{Leo2001RF}, principle component regression~\cite{Jolliffe1982PCR}, etc. 
However, these techniques require extensive domain expertise for designing task-specific features, limiting their adaptability and generalization across different tasks.
With the rise of deep learning, neural networks have been utilized to automatically learn features from raw polymer data. 
Leveraging polymer representations, such as P-SMILES~\cite{Kuenneth2022polyBERT}, 
convolutional neural networks, recurrent neural networks and transformer models 
have been employed to establish effective mappings between polymer sequences and their corresponding properties.
Despite their advances, the efficacy of these deep learning models largely depends on 
the availability of sufficiently large and labeled datasets, which are often scarce in practical scenarios.
Motivated by the success of pretraining-finetuning paradiams in natural language processing~\cite{Devlin2019Bert} 
and contrastive learning strategies in computer vision~\cite{Ting2020SimCLR}, 
recent approaches have introduced masked-token prediction~\cite{Kuenneth2022polyBERT,Pei2023PPP} and 
contrastive methods~\cite{Wang2024MMPolymer} into polymer representation learning.
Due to the importance of 3D structure information in deciding polymer property~\cite{Wang2023A3PT,Hannes2022Infomax}, 
MMPolymer~\cite{Wang2024MMPolymer} further combine 3D information and 1D P-SMILES representation to enhance polymer property prediction. 
However, existing techniques predominantly rely on monomer-level P-SMILES, 
whose properties often differ from the resultant polymer structures after polymerization. 
To address these limitations and leverage multimodal information more effectively, 
we propose a Multimodal Infinite Polymer Sequence (MIPS) pre-training framework.
MIPS represents polymers explicitly as infinite sequences of monomers and incorporates both topological and spatial information. 
Experimental results demonstrate that MIPS significantly outperforms existing methods across eight polymer property prediction tasks, 
setting new the state-of-the-art performance. 
Furthermore, ablation and interpretation studies reveal that MIPS can learn meaningful polymer representations.

\subsection{Molecular Self-supervised Learning}
Existing polymer representation learning methods primarily rely on monomer-level P-SMILES~\cite{Kuenneth2022polyBERT}, 
which allows monomers to be treated as special molecular instances. 
Under this perspective, a variety of molecular self-supervised learning (SSL) methods 
can be seamlessly adapted to polymer domain~\cite{Pei2023PPP}.
SSL leverages large volumes of unlabeled data to learn powerful feature representations. 
Molecular SSL methods can be broadly categorized according to the chosen molecular input form: 
string-based, graph-based, 3D-based, and hybrid approaches. 
String-based methods, such as MolBERT~\citep{Fabian2020MolBERT} and MolGPT~\citep{Bagal2022MolGPT}, 
regard molecules as sequences of characters by SMILES~\cite{Weininger1988SMILES} and 
utilize transformer architectures with mask-based token prediction to capture molecular representation. 
Graph-based methods incorporate graph-level SSL strategies, such as masked atom/bond prediction~\cite{Rong2020GROVER} 
and graph contrastive learning based on molcular augmentations~\cite{Wang2022MolCLR}.
Recognizing the importance of spatial information, 
3D-based methods have been developed with denoising autoencoders~\cite{Zaidi2023Denoising} and 
equivariant neural networks~\cite{Satorras2021EGNN} specifically designed to learn robust 3D structural representations.
Additionally, since complementary information is contained within 2D topology and the 3D geometry, 
multiple hybrid methods have emerged~\cite{Geng2023UniMol,Hannes2022Infomax,Lin2024CrossView}
that combine both 2D and 3D representations for richer feature extraction.
In this work, we extend the molecular SSL concepts to polymer representation learning. 
Our proposed MIPS framework represents polymers as infinite monomer sequences and 
simultaneously leverages topological and spatial information through a cross-modal fusion mechanism. 

\subsection{Expressive Power of Graph Neural Networks} 
The analysis of the expressive power of graph neural networks (GNNs) 
represents an active and significant area of research within theoretical graph learning~\cite{Zhang2025Expressive}.
Such analysis deepens our understanding of GNN capabilities and 
provides valuable insights and inspiration for developing more powerful architectures~\cite{Xu2019GIN}.
The seminal work of \citet{Xu2019GIN} establishs that the expressivity of GNNs in distinguishing non-isomorphic graphs 
is limited by the Weisfeiler–Lehman~(WL) test, 
and further introduced the Graph Isomorphism Network (GIN), which achieves this theoretical upper bound.
Subsequent studies have designed GNN variants capable of surpassing the WL test, thus achieving higher expressive power. 
For instance, K-GNN~\cite{Morris2019HigherOrderGNN} enhances GNN expressivity by 
generalizing to higher-order architectures based on the multidimensional k-WL algorithm.
$\mathcal{O}$-GNN~\cite{Zhu2023GNNIR}, GSN-v~\cite{Giorgos2023GNNCounting}, FragNet~\cite{Tom2024FragNet} 
explicitly incorporate the fragment bias into GNNs to improve representational capacity. 
In this work, we extend the previous analysis of GNN expressivity to infinite polymer graph sequences. 
We demonstrate that the standard WL test, as well as the message passing and localized graph attention mechanisms, 
can not distinguish between the defined twin polymer graphs. 
To address this limitation, we propose backbone embedding that 
enhances the expressive power of message passing and localized graph attention mechanisms for distinguishing twin polymer graphs. 
Experimental results confirm the effectiveness of the backbone embedding, 
especially in scenarios involving polymers that contain multiple ring structures.  

\subsection{Adversarial Attack}
Although deep learning has machieved remarkable progress across various fields~\cite{LeCun2015DL}, 
neural networks are vulnerable to small perturbations in input data~\cite{Szegedy2014Intriguing}. 
Such adversarial samples share semantic content identical to the original inputs, yet significantly degrade the performance of models, 
raising serious concerns about the reliability and robustness of deep learning approaches. 
In this work, we propose a Repeat and Shift Invariance Test (RSIT) to evaluate the robustness of polymer modeling methods. 
RSIT generates adversarial samples by randomly augmenting the polymer P-SMILES through sequence translation and repetition, without changing the underlying polymer. 
We show that existing polymer modeling methods, including "star keep", "star remove", and "star substitution"~\cite{Wang2024MMPolymer}, 
exhibit significant performance degradation under RSIT. 
We model the invariance of translation and repetition by directly modeling infinite polymer sequences. 
Our infinite polymer sequence modeling approach is resistant to RSIT and achieve superior empirical performance, 
underscoring its robustness and effectiveness. 

\section{Methods}

In this section, we introduce the Multimodal Infinite Polymer Sequence (MIPS) pre-training framework 
which models polymers as infinite sequences of monomers and integrates both topological and spatial information. 
The pipeline of MIPS is illustrated in \cref{fig:pipeline}. MIPS consists of three parts: 
topological structure encoder, spatial structure encoder, and cross-modal fusion.

\subsection{Topological Structure Encoder}


Polymer sequences, composed of repeating units called monomers, are usually represented by P-SMILES~\cite{Kuenneth2022polyBERT}, 
which encode the topological structures of monomers and two endpoints involved in polymerization.  
We show an illustration of P-SMILES in \cref{fig:preliminary}~(left). 
The star symbols in P-SMILES indicate the endpoints of the monomer. 
\textit{An important property of P-SMILES is that the polymer represented by P-SMILES remains invariant under translation and repetition of P-SMILES (shown in \cref{fig:preliminary}~(left)).} 
To model such invariance, we propose to directly build model on infinite polymer sequences 
since infinite polymer sequences are naturally invariant under the translation and repetition of P-SMILES.

\begin{algorithm}[t]
  \caption{Random Augment P-SMILES}
  \KwIn{P-SMILES $s$}
  \KwOut{Augmented P-SMILES $\hat{s}=\text{random\_augment}\left(s\right)$}
  $\hat{s} \gets s $\; 
  $p \gets \texttt{Bernoulli(0.5)}$\; 
  \If{$p == 1$}{
      $\hat{s} \gets \texttt{repeat}(\hat{s})$; \ \tcp{e.g., *CONO* $\rightarrow$ *CONOCONO*}
  }
  $\hat{s} \gets \texttt{random\_translation}(\hat{s})$; \ \tcp{e.g., *CONO* $\rightarrow$ *NOCO*}
  \Return $\hat{s}$
  \label{alg:random_augment}
\end{algorithm}

\subsubsection{Repeat and Shift Invariance Test}

To test how well the model can capture the translation and repetition invariance of P-SMILES, 
we propose Repeat and Shift Invariance Test (RSIT) (\cref{alg:rsit}). 
RSIT is an adversarial attack algorithm that generates adversarial samples by randomly augmenting the input P-SMILES through translation and repetition. 
We will show in the experiment section that existing modeling methods, including "star keep", "star remove", and "star substitution" (\cref{fig:preliminary}~(right)) 
exhibit significant performance degradation under RSIT, while our infinite polymer sequence modeling approach is resistant to RSIT. 

\subsubsection{Modeling Infinite Polymer Sequences}
In this subsection, we introduce how to model infinite polymer sequences 
using message passing mechanism (MPM) and graph attention mechanism (GTM). 
We define the monomer graph as $G = (V, E, \mathbf{X})$, where $V=\left\{v_i\right\}_{i=0}^{\abs{V}-1}$ is the list of atoms in the monomer, 
$E=\left\{e_i\right\}_{i=1}^{\abs{E}}$ is the list of bonds, $e_i=\left(v_{i1}, v_{i2}\right)$, and $\mathbf{X}\in \R^{d \times \abs{V}}$ is the atom feature matrix. 
Without loss of generality, we assume $v_0$ and $v_{\abs{V}-1}$ are the boundary nodes of the monomer to form polymer. 
The polymer graph $G^p = (V^p, E^p, \mathbf{X}^p)$ is the periodic repetition of the monomer graph $G$, 
where the $i$-th ($i \in \integers$) atom of $V^p$ has the same properties as the $(i \ mod \ \abs{V})$-th atom of $V$. 
Except for the bonds in each monomer, $E^p$ also contains the bonds between the boundary atoms of adjacent monomers, 
i.e., $v^{p}_{k\abs{V}}$ and $v^{p}_{k\abs{V}-1}$, $\forall \ k \in \integers$.
\subsubsection*{Message Passing Mechanism on Infinite Polymer Sequences} 
Denote the neighborhood of node $v$ as $\mathcal{N}(v_i)$, and the node feature of $v$ as $\mathbf{x}_v$.
The message passing mechanism~\cite{Kipf2017GCN,Xu2019GIN,Velickovic2018GAT,Hamilton2017GraphSAGE} updates the node feature of $v$ as follows:
\begin{equation}
  \mathbf{x}_v \gets \text{UPDATE}\left(\mathbf{x}_v, \text{AGG} \left( \{\mathbf{x}_u\}_{u \in \mathcal{N}(v)} \right) \right),
    \label{eq:mpm}
\end{equation}
where $\text{UPDATE}$ and $\text{AGG}$ serve as the update and message aggregation functions, respectively. 
Typical networks using message passing mechanism include Graph Convolutional Network (GCN)~\cite{Kipf2017GCN}, 
Graph Isomorphism Network (GIN)~\cite{Xu2019GIN}, Graph Attention Network (GAT)~\cite{Velickovic2018GAT}, GraphSAGE~\cite{Hamilton2017GraphSAGE}, etc.
Directly applying message passing mechanism on polymer graph is prohibitive due to the infinite number of nodes in the polymer graph. 
However, we will show that message passing on infinite polymer sequences is 
equivalent to message passing on the induced star-linking graph of monomers. 
All the proofs are provided in the appendex.  

\begin{algorithm}[t]
  \caption{Repeat and Shift Invariance Test (RSIT)}
  \KwIn{model $m$, number of trials $T$, sample $x$, label $y$, loss function $L$}
  \KwOut{adversarial loss $\texttt{adv\_loss}$, adversarial prediction $\texttt{adv\_predict}$}
  \For{$i = 1$ to $T$}{
      
      $\texttt{adv\_x} \gets \text{random\_augment}(x)$ (\cref{alg:random_augment})\;
      $\hat{y} \gets m.predict(\texttt{adv\_x})$\;
      $\texttt{loss} \gets L(\hat{y}, y)$\;
      \If{$\texttt{loss} >\texttt{adv\_loss}$}{
          $\texttt{adv\_loss} \gets \texttt{loss}$\;
          $\texttt{adv\_predict} \gets \hat{y}$\;
      }
  }
  \Return $\texttt{adv\_loss}$, $\texttt{adv\_predict}$
  \label{alg:rsit}
\end{algorithm}

\begin{definition}[Induced Star-Linking Graph $G^{*}$]
The induced star linking graph $G^{*} = (V^{*}, E^{*}, \mathbf{X}^{*})$ of the monomer graph $G=(V, E, \mathbf{X})$ 
is constructed by linking the boundary atoms of $G$, 
where $V^{*}=V$, $E^{*}=E \cup \{ \left(v_0, v_{\abs{V}-1} \right) \}$, $\mathbf{X}^{*} = \mathbf{X}$.
\end{definition}
We assume $G$, $G^p$, and $G^{*}$ share the same initial node features, i.e., $\mathbf{x}_i=\mathbf{x}_i^p=\mathbf{x}_i^*$, $\forall 0 \leq i < \abs{V}$.
\begin{proposition}
  \label{proposition:mpm_eq_star_linking}
  Under the same message passing mechanism (\cref{eq:mpm}), 
  the node feature of the $i$-th atom in the polymer graph $G^p$ 
  is the same as the node feature of the $i$-th atom in the induced star linking graph $G^{*}$, $\forall 0 \leq i < \abs{V}$.
\end{proposition}
All the proofs are provided in \SM.
\begin{theorem}
  \label{theorem:mpm_eq_star_linking}
  If the network is composed of message passing layers (\cref{eq:mpm}), nodewise transformations, and end with a mean pooling, 
  Then the output of the network on the polymer graph $G^p$ is the same as that on the induced star linking graph $G^{*}$. 
\end{theorem}
Using \cref{theorem:mpm_eq_star_linking}, 
we can directly apply the message passing mechanism on the induced star-linking graph $G^{*}$ to effectively model infinite polymer sequences. 
\subsubsection*{Localized Graph Attention Mechanism on Infinite Polymer Sequences} 
To improve the model capability and capture the long range dependency on graphs, graph attention mechanism is introduced~\cite{Vaswani2017Transformer,Ying2021Graphormer,Shi2022Graphormer}.
Given the graph $G=(V, E, \mathbf{X})$, the attention mechanism computes $\mathbf{Y} = \text{Attn}\left(V, E, \mathbf{X}\right)$ as follows:
\[
  \mathbf{Q} = \mathbf{W}^Q \mathbf{X}, \ \mathbf{K} = \mathbf{W}^K \mathbf{X}, \ \mathbf{V} = \mathbf{W}^V \mathbf{X},
\]
\[
  \mathbf{A} = \text{Softmax}\left( \frac{\mathbf{K}^T\mathbf{Q}}{\sqrt{d}} + \mathbf{A}^d + \mathbf{A}^p \right),
\]
$\mathbf{W}^Q$, $\mathbf{W}^K$ and $\mathbf{W}^V \in \R^{d\times d}$ are the projection weight matrices 
for query~($\mathbf{Q}$), key~($\mathbf{K}$) and value~($\mathbf{V}$) matrices, respectively. 
The softmax operation is applied column-wise to $\mathbf{A}$. 
$\mathbf{A}^d$ and $\mathbf{A}^p$ are distance attention and path attention.
Assume the distance between node $i$ and node $j$ in graph is $d_{ij}$, 
and the shortest path between node $i$ and node $j$ is $\mathbf{p}_{ij}=(i_0, \ldots, i_{d_{ij}})$, 
then $\mathbf{A}^d$ and $\mathbf{A}^p$ are computed as: 
\[
  A^d_{ij} = f_{\text{dist}}\left(d_{ij} \right)
\]
\[
  A^p_{ij} = f_{\text{path}} \left(\mathbf{p}_{ij}\right),
\]
where $f_{\text{dist}}$ and $f_{\text{path}}$ are functions with learnable parameters and output scalar.
\begin{equation}
  \mathbf{Y} =\mathbf{V}\mathbf{A}
  \label{eq:global_attn}
\end{equation}
  

Directly applying global graph attention mechanism to the polymer graph is infeasible due to its infinite number of nodes.
To overcome the infinite receptive field issue, 
we propose to replace the global graph attention mechanism with localized graph attention (LGA). 
LGA computes $\hat{\mathbf{Y}} = \text{LocalAttn}\left(V, E, \mathbf{X}\right)$ by
substituting \cref{eq:global_attn} with the following equation:
\begin{equation}
  \begin{aligned}
    \mathbf{\hat{A}}_{ij} & = \mathbf{A}_{ij}*\mathbf{1}_{\{d_{ij} < d_{{thres}}\}}\\
  \hat{\mathbf{Y}} & = \mathbf{V}\mathbf{\hat{A}}.
  \label{eq:local_attn}
\end{aligned}
\end{equation}
$\mathbf{1}_{\{d_{ij}\leq d_{{thres}}\}}$ equals to 1 if the distance between node $i$ and node $j$ is less than $d_{{thres}}$, otherwise 0.

Similar to message passing mechanism, we will show that localized graph attention on infinite polymer sequences is equivalent to localized graph attention on monomer with star linking strategy.

\begin{theorem}
  \label{theorem:attn_eq_star_linking}
  Assume that the distance between boundary atoms in monomer graph $G$ is larger than $2d_{{thres}}-1$.
  If the network consists of localized graph attention layers, nodewise transformations, and concludes with a mean pooling, 
  then its output on the polymer graph $G^p$ will be identical to that on the induced star-linking graph $G^{*}$. 
\end{theorem}

If the distance between boundary atoms in monomer graph $G$ is less than $2d_{{thres}}-1$, 
we first repeat the monomer graph $G$ until the distance exceeds $2d_{{thres}}-1$, 
Then, we apply the localized graph attention mechanism to the augmented monomer graph. 

\subsubsection{Backbone Embedding}
Although message passing mechanism is powerful,  
capable of performing graph isomorphism tests with the same expressive power as the WL test~\cite{Xu2019GIN}, 
we will demonstrate that it often fails in the case of infinite polymer sequences with ring structures which are ubiquitous in practice. 

\begin{figure}[t]
  \captionsetup[subfigure]{justification=centering}
  \centering
  \begin{subfigure}[b]{0.96\columnwidth}
    \centering
    \centerline{\includegraphics[width=\textwidth]{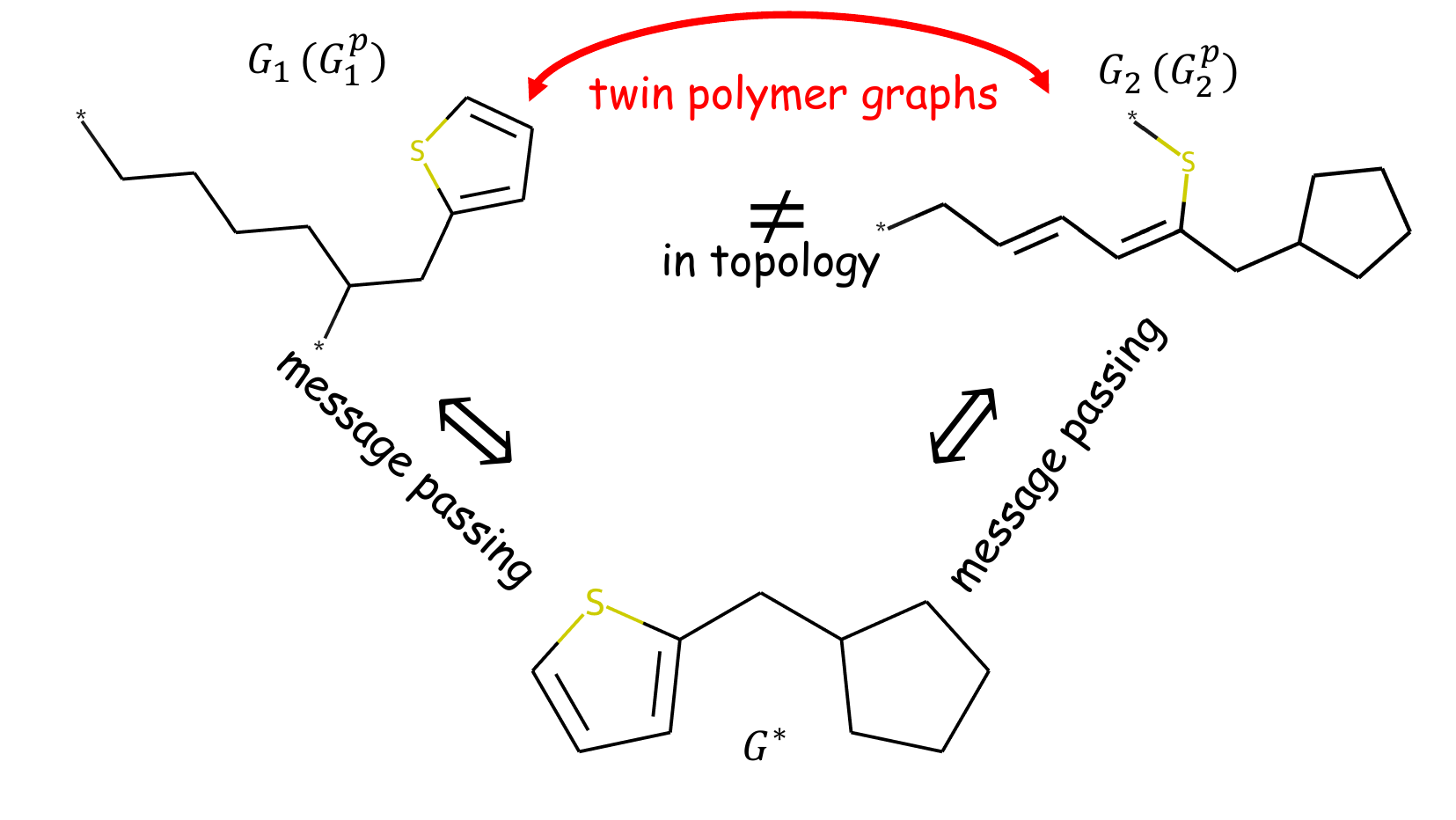}}
  \end{subfigure}
  \caption{An example of twin polymer graphs $\left(G_1^p, G_2^p\right)$. 
  By \cref{theorem:mpm_eq_star_linking}, message passing mechanism on $G_1^p$/$G_2^p$ is equivalent to that on $G^{*}$.}
  \label{fig:twin_polymer_graph}
\end{figure}

\begin{definition}[Twin Polymer Graphs]
Assume the monomer graphs of two polymers $G_1^p$ and $G_2^p$ are $G_1$ and $G_2$, respectively, with $G_1^p \neq G_2^p$. 
If their induced star-linking graphs $G_1^{*}$ and $G_2^{*}$ satisfies $G_1^{*} = G_2^{*}$, 
we call $\left(G_1^p, G_2^p\right)$ a pair of twin polymer graphs. 
\end{definition}
We show an example of twin polymer graphs in \cref{fig:twin_polymer_graph}. 
\begin{lemma}
  \label{lemma:polymer_wl_test}
  WL test can not distinguish twin polymer graphs $\left(G_1^p, G_2^p\right)$. 
\end{lemma}

\begin{theorem}
  \label{lemma:polymer_mpm_test}
  On twin polymer graphs $\left(G_1^p, G_2^p\right)$, massage passing mechanism~(MPM, \cref{eq:mpm}) or localized graph attention~(LGA, \cref{eq:local_attn}) will produce identical outputs. 
  In other words, MPM and LGA are unable to distinguish twin polymer graphs. 
\end{theorem}


To address the issue of message passing and localized attention mechanisms on infinite polymer sequences with ring structures, 
we propose \emph{backbone embedding} to improve the capability of message passing and localized attention mechanisms. 
We define the backbone of monomer graph as the atoms and rings on the shortest path between the boundary atoms. 
We add a learnable embedding, named backbone embedding, to the backbone atoms to distinguish the backbone and the rings on the side chain. 
The backbone bone embedding $\mathbf{b}_i$ of node $i$ is defined as: 
\[
  \mathbf{b}_i = 
  \begin{cases}
    \mathbf{b} & \text{if } i \text{ is a backbone atom} \\
    0 & \text{otherwise}
  \end{cases},
\]
where $\mathbf{b} \in \R^{d}$ is the learnable vector. 

\subsubsection{Localized Graph Transformer}
Assume we are given the induced star-linking graph $G^{*} = (V^{*}, E^{*}, \mathbf{X}^{*})$. 
We construct the backbone embedding matrix $\mathbf{B} \in \R^{d\times \abs{V}}$ by stacking the backbone embedding of all atoms. 
Next, we initialize the node representation for the graph as $\mathbf{X}^t_0 = \mathbf{X}^{*}+\mathbf{B}$. 
Localized graph transformer consists of $L$ localized graph attention layers. 
For the $l$-th layer, the node feature is updated as follows:
\[
  \mathbf{X}^t_{l+\frac{1}{2}} = \text{LayerNorm}\left( \text{LocalAttn}\left(V^{*}, E^{*}, \mathbf{X}^t_l \right) + \mathbf{X}^t_l \right),
\]
\[
  \mathbf{X}^t_{l+1} = \text{LayerNorm}\left( \text{FFN}(\mathbf{X}^t_{l+\frac{1}{2}}) + \mathbf{X}^t_{l+\frac{1}{2}} \right),
\]
where $\text{FFN}$ is a two-layer MLP with ReLU activation~\cite{Agarap2018DeepLU}. 
We adopt the same distance attention and path attention mechanisms as proposed in \citet{Ying2021Graphormer}. 

\begin{table*}[h]
  \centering
  \caption{$R^2$ Performance of GIN3-512 with "star keep", "star remove", "star substitution", 
  and our proposed "star linking" strategies w. and w.o RSIT with $T=5$ trials. RSIT Gap: Average performance drop under RSIT.}
  \begin{adjustbox}{width=\textwidth,center}
  \begin{tabular}{c|cccccccc|c}
  \toprule 
                      & Egc         & Egb         & Eea         & Ei          & Xc           & EPS         & Nc          & Eat         & RSIT Gap\\ \midrule
  Star Keep           & $0.866_{\pm0.013}$ & $0.901_{\pm0.012}$ & $0.890_{\pm{0.033}}$ & $0.762_{\pm{0.062}}$ & $0.252_{\pm{0.090}}$  & $0.723_{\pm{0.052}}$ & $0.814_{\pm{0.034}}$ & $0.956_{\pm{0.005}}$ & --                  \\
  Star Remove         & $0.857_{\pm0.015}$ & $0.881_{\pm{0.013}}$ & $0.871_{\pm{0.044}}$ & $0.772_{\pm{0.054}}$ & $0.243_{\pm{0.116}}$  & $0.714_{\pm{0.051}}$ & $0.805_{\pm{0.033}}$ & $0.939_{\pm{0.008}}$ & --                  \\
  Star Substitution   & $0.868_{\pm0.011}$ & $0.916_{\pm{0.012}}$ & $0.904_{\pm{0.035}}$ & $0.779_{\pm{0.056}}$ & $0.268_{\pm{0.108}}$  & $\mathbf{0.773}_{\pm{0.038}}$ & $0.845_{\pm{0.038}}$ & $0.963_{\pm{0.003}}$ & --                  \\
  Star Linking~(ours) &  $\mathbf{0.870}_{\pm{0.010}}$ &  $\mathbf{0.918}_{\pm{0.010}}$ &  $\mathbf{0.916}_{\pm{0.025}}$ &  $\mathbf{0.785}_{\pm{0.060}}$ &  $\mathbf{0.279}_{\pm{0.130}}$  & $0.772_{\pm{0.048}}$ &  $\mathbf{0.847}_{\pm{0.037}}$ & $\mathbf{0.966}_{\pm{0.006}}$ & --                  \\ \midrule
  Star Keep + RSIT              & $0.745_{\pm{0.020}}$ & $0.780_{\pm{0.017}}$ & $0.593_{\pm{0.110}}$ & $0.524_{\pm{0.154}}$ & $-0.287_{\pm{0.271}}$ & $0.466_{\pm{0.089}}$ & $0.627_{\pm{0.069}}$ & $0.928_{\pm{0.012}}$ & $0.233$               \\ 
  Star Remove + RSIT              & $0.765_{\pm{0.026}}$ & $0.792_{\pm{0.035}}$ & $0.696_{\pm{0.055}}$ & $0.468_{\pm{0.207}}$ & $-0.181_{\pm{0.291}}$ & $0.468_{\pm{0.085}}$ & $0.615_{\pm{0.073}}$ & $0.864_{\pm{0.016}}$ & $0.199$               \\ 
  Star Substitution + RSIT              & $0.796_{\pm{0.020}}$ & $0.871_{\pm{0.020}}$ & $0.821_{\pm{0.043}}$ & $0.645_{\pm{0.081}}$ & $0.041_{\pm{0.142}}$  & $0.638_{\pm{0.074}}$ & $0.729_{\pm{0.056}}$ & $0.922_{\pm{0.012}}$ & $0.107$               \\ 
  Star Linking~(ours) + RSIT              &  $\mathbf{0.870}_{\pm{0.010}}$ &  $\mathbf{0.918}_{\pm{0.010}}$ &  $\mathbf{0.916}_{\pm{0.025}}$ &  $\mathbf{0.785}_{\pm{0.060}}$ &  $\mathbf{0.279}_{\pm{0.130}}$  & $\mathbf{0.772}_{\pm{0.048}}$ &  $\mathbf{0.847}_{\pm{0.037}}$ &  $\mathbf{0.966}_{\pm{0.006}}$ &  $\mathbf{0.000}$                   \\ \bottomrule
  \end{tabular}
\end{adjustbox} 
\label{table:rsit}
\end{table*}

\begin{table*}[h]
  \centering
  \caption{$R^2$ Performance of GIN3-512 w./w.o backbone embedding~(BE).} 
  \begin{adjustbox}{width=\textwidth,center}
  \begin{tabular}{c|cccccccc}
  \toprule
  & Egc & Egb & Eea & Ei & Xc & EPS & Nc & Eat \\ \midrule
Star Linking & $0.870_{\pm0.010}$ & $0.918_{\pm0.010}$ & $0.916_{\pm0.025}$ & $\mathbf{0.785}_{\pm0.060}$ & $0.279_{\pm0.130}$ & $\mathbf{0.772}_{\pm0.048}$ & $0.847_{\pm0.037}$ & $0.966_{\pm0.006}$ \\
Star Linking + BE & $\mathbf{0.882}_{\pm0.007}$ & $\mathbf{0.920}_{\pm0.010}$ & $\mathbf{0.917}_{\pm0.029}$ & $\mathbf{0.785}_{\pm0.062}$ & $\mathbf{0.346}_{\pm0.083}$ & $\mathbf{0.772}_{\pm0.053}$ & $\mathbf{0.849}_{\pm0.043}$ & $\mathbf{0.967}_{\pm0.004}$ \\ \bottomrule
  \end{tabular}
\end{adjustbox} 
\label{table:be}
\end{table*}

\subsection{Spatial Structure Encoder}

We extract the spatial information of polymers by computing the 3D descriptors of monomers. 
3D descriptors are computed by the 3D coordinates of atoms in the monomer. 
We use the 3D molecular descriptors in \citet{Yang2019AnalyzingLM}, 
and atom pair 3D descriptors~\cite{Mahendra2014AtomPair,Landrum2016RDKit}.
Denote the 3D descriptors of the polymer as $\{ \mathbf{x}^{s}_{i}\}_{i=1}^{N_s}$, $\mathbf{x}^{s}_{i} \in \R^{d_i}$. 
We use a linear projection to encode the 3D descriptors to the same dimension as the topological structure encoder:
\[
  \hat{\mathbf{x}}^{s}_{i} = \mathbf{W}_i\mathbf{x}^{s}_{i}, \ \forall \  i \in \{1, \ldots, N_s\},
\]
where $\mathbf{W}_i \in \R^{d \times d_i}$ is the linear projection matrix. 
We stack $\hat{\mathbf{x}}^{s}_{i}$ as the spatial structure feature matrix $\mathbf{X}^s \in \R^{d \times N_s}$.

\subsection{Cross-modal Fusion}

With the topological structure feature matrix ${X}^t={X}^t_L$ and the spatial structure feature matrix ${X}^s$, 
we propose a cross-modal fusion mechanism to fuse the topological and spatial information of polymers. 
To this end, we use a cross-attention layer~\cite{Vaswani2017Transformer} 
to integrate the spatial information to each atom in the topological structure space.
$\mathbf{X}^{ts} = \text{CrossModalFusion}({X}^t, {X}^s)$ is computed as follows:

\[
  \mathbf{Q} = \mathbf{W}^Q \mathbf{X}^t, \ \mathbf{K} = \mathbf{W}^K \mathbf{X}^s, \ \mathbf{V}^s = \mathbf{W}^V \mathbf{X}^s,
\]
\[
  \mathbf{A}^{ts} = \text{Softmax}\left( \frac{\mathbf{K^T}\mathbf{Q}}{\sqrt{d}} \right),
\]
\[
  \mathbf{X}^{ts} = \text{LayerNorm}\left( \mathbf{X}^t + \mathbf{V^s}\mathbf{A}^{ts} \right).
\]

\subsection{Pre-training Objective}
We adopt masked-atom prediction as the pre-training objective~\cite{Rong2020GROVER,Li2023KPGT,Wang2025Fragformer}. 
Specifically, we mask the complete features of atoms within the polymer graph with a probability of $p_{\text{mask}}$. 
The type of the masked atoms is then predicted using the outputs of the cross-modal fusion layer, 
which are processed by a linear classifier.

\section{Experiments and Results}

\paragraph*{Dataset}

\begin{table}[h]
  \centering
  \caption{Brief statistics and descriptions of downstream datasets. E: Electronic, TP: Thermodynamic and Physical, OD: Optical and Dielectric.}
  \begin{adjustbox}{width=\columnwidth,center}
  \begin{tabular}{@{}cccccc@{}}
  \toprule
  Dataset & Property            & Task Type     & \# of Samples & Unit      & Data Range         \\ \midrule
  Egc     & bandgap (chain)         & E & 3380          & eV        & $\left[0.02, 8.30\right]$   \\
  Egb     & bandgap (bulk)        & E   & 561           & eV        & $\left[0.39, 10.05\right]$  \\
  Eea     & electron affinity     & E   & 368           & eV        & $\left[0.39, 4.61\right]$   \\
  Ei      & ionization energy     & E   & 370           & eV        & $\left[3.55, 9.61\right]$   \\ \midrule               
  Eat     & atomization energy    & TP   & 390           & eV·atom$^{-1}$ & $\left[6.83, 5.02\right]$ \\
  Xc      & crystallization tendency & TP & 432           & \%        & $\left[0.13, 98.41\right]$  \\ \midrule                           
  EPS     & dielectric constant    & OD  & 382           & 1         & $\left[2.61, 8.52\right]$   \\
  Nc      & refractive index       & OD  & 382           & 1         & $\left[1.48, 2.58\right]$   \\ \bottomrule
  \end{tabular}
\end{adjustbox} 
\label{table:dataset}
\end{table}

Following \citet{Wang2024MMPolymer}, we utilize PL1M~\cite{Ruimin2020PI1M} as the dataset for multimodal pre-training, 
which contains approximately $1$ million unlabeled polymer sequences. 
For the downstream tasks, we employ eight polymer property prediction datasets introduced in \citet{Kunneth2021Polymer,Pei2023PPP}, 
derived from the density functional theory calculations~\cite{Orio2009DFT}. 
A brief summary of downstream datasets is provided in \cref{table:dataset}, 
with detailed descriptions available in \SM. 
These datasets cover a wide variety of polymer properties, enabling a comprehensive evaluation of our model's performance. 
We employ root mean square error~(RMSE) or R-squared~($R^2$) over the five folds in cross-validation as the evaluation metrics. 
These metrics align with those used in previous studies~\cite{Wang2024MMPolymer,Pei2023PPP}, ensuring a fair comparison.

\begin{table*}[t]
  \centering
  \caption{The performance comparison of different methods on eight polymer property datasets, and the best result for each polymer property dataset has been bolded.} 
    
    \begin{adjustbox}{width=\textwidth,center}
      \begin{tabular}{c|c|cccccccc}
      \toprule
      Metric & Method & Egc   & Egb   & Eea   & Ei    & Xc    & EPS   & Nc    & Eat \\
      \midrule
      \multirow{9}[0]{*}{RMSE ($\downarrow$)} 
          & ChemBERTa~\cite{Ahmad2022ChemBERT} & 
          $0.539_{\pm{0.049}}$ & $0.664_{\pm{0.079}}$  & $0.350_{\pm{0.036}}$ & $0.485_{\pm{0.086}}$ & 
          $18.711_{\pm{1.396}}$ & $0.603_{\pm{0.083}}$ & $0.140_{\pm{0.010}}$ & $0.219_{\pm{0.056}}$  \\
  
          & MolCLR~\cite{Wang2022MolCLR} & 
          $0.587_{\pm{0.024}}$ & $0.644_{\pm{0.072}}$ & $0.404_{\pm{0.017}}$ & $0.533_{\pm{0.053}}$ & 
          $21.719_{\pm{1.631}}$ & $0.631_{\pm{0.045}}$ & $0.117_{\pm{0.015}}$ & $0.094_{\pm{0.033}}$ \\
  
          & 3D Infomax~\cite{Hannes2022Infomax} & 
          $0.494_{\pm{0.039}}$ & $0.553_{\pm{0.032}}$ & $0.335_{\pm{0.055}}$ & $0.449_{\pm{0.086}}$ & 
          $19.483_{\pm{2.491}}$ & $0.582_{\pm{0.054}}$ & $0.101_{\pm{0.018}}$ & $0.094_{\pm{0.039}}$ \\
  
          & Uni-Mol~\cite{Geng2023UniMol} & 
          $0.489_{\pm{0.028}}$ & $0.531_{\pm{0.055}}$  & $0.332_{\pm{0.027}}$ & $0.407_{\pm{0.080}}$ & 
          ${17.414}_{\pm{1.581}}$ & ${0.536}_{\pm{0.053}}$ & $0.095_{\pm{0.013}}$ & $0.084_{\pm{0.034}}$  \\
          
          & SML~\cite{Pei2023PPP} & 
          $0.489_{\pm{0.056}}$ & $0.547_{\pm{0.110}}$ & ${0.313}_{\pm{0.016}}$ & $0.432_{\pm{0.060}}$ & 
          $18.981_{\pm{1.258}}$ & $0.576_{\pm{0.020}}$ & $0.102_{\pm{0.010}}$ & $0.062_{\pm{0.014}}$ \\
      
          & PLM~\cite{Pei2023PPP} & 
          $0.459_{\pm{0.036}}$ & ${0.528}_{\pm{0.081}}$ & $0.322_{\pm{0.037}}$ & $0.444_{\pm{0.062}}$ & 
          $19.181_{\pm{1.308}}$ & $0.576_{\pm{0.060}}$ & $0.100_{\pm{0.010}}$ & $0.050_{\pm{0.010}}$ \\
          
          & polyBERT~\cite{Kuenneth2022polyBERT} & 
          $0.553_{\pm{0.011}}$ & $0.759_{\pm{0.042}}$ & $0.363_{\pm{0.037}}$ & $0.526_{\pm{0.068}}$ & 
          $18.437_{\pm{0.560}}$ & $0.618_{\pm{0.049}}$ & $0.113_{\pm{0.003}}$ & $0.172_{\pm{0.016}}$ \\
  
          & Transpolymer~\cite{Xu2022TransPolymer} & 
          ${0.453}_{\pm{0.007}}$ & $0.576_{\pm{0.021}}$ & ${0.326}_{\pm{0.040}}$ & ${0.397}_{\pm{0.061}}$ & 
          ${17.740}_{\pm{0.732}}$ & ${0.547}_{\pm{0.051}}$ & ${0.096}_{\pm{0.016}}$ & ${0.147}_{\pm{0.093}}$ \\
          
          & MMPolymer~\cite{Wang2024MMPolymer} & 
          $0.431_{\pm{0.017}}$ & $0.496_{\pm{0.031}}$  & $0.286_{\pm{0.029}}$  & $0.390_{\pm{0.057}}$ & 
          $16.814_{\pm{0.867}}$ & $0.511_{\pm{0.035}}$ & $0.087_{\pm{0.010}}$ & $0.061_{\pm{0.016}}$ \\

          & MIPS (ours) & 
            $\mathbf{0.429}_{\pm{0.014}}$ & $\mathbf{0.460}_{\pm{0.037}}$ & $\mathbf{0.267}_{\pm{0.018}}$ & $\mathbf{0.384}_{\pm{0.048}}$ & 
            $\mathbf{16.435}_{\pm{0.576}}$ & $\mathbf{0.484}_{\pm{0.049}}$ & $\mathbf{0.083}_{\pm{0.010}}$ & $\mathbf{0.038}_{\pm{0.006}}$ \\

      \midrule
      \multirow{9}[0]{*}{$R^2$ ($\uparrow$)} 
            & ChemBERTa~\cite{Ahmad2022ChemBERT} & 
            $0.880_{\pm{0.023}}$ & $0.881_{\pm{0.028}}$ & $0.888_{\pm{0.035}}$ & $0.745_{\pm{0.102}}$ & 
            $0.365_{\pm{0.098}}$ & $0.682_{\pm{0.123}}$ & $0.643_{\pm{0.076}}$ & $0.590_{\pm{0.078}}$ \\
  
            & MolCLR~\cite{Wang2022MolCLR} & 
            $0.858_{\pm{0.010}}$ & $0.882_{\pm{0.027}}$ & $0.854_{\pm{0.038}}$ & $0.689_{\pm{0.037}}$ & 
            $0.176_{\pm{0.026}}$ & $0.683_{\pm{0.020}}$ & $0.764_{\pm{0.037}}$ & $0.885_{\pm{0.104}}$ \\
  
            & 3D Infomax~\cite{Hannes2022Infomax} & 
            $0.900_{\pm{0.016}}$ & $0.898_{\pm{0.018}}$ & $0.891_{\pm{0.045}}$ & $0.766_{\pm{0.086}}$ & 
            $0.274_{\pm{0.122}}$ & $0.690_{\pm{0.063}}$ & $0.797_{\pm{0.086}}$ & $0.869_{\pm{0.097}}$ \\
  
            & Uni-Mol~\cite{Geng2023UniMol} & 
            $0.901_{\pm{0.013}}$ & $0.925_{\pm{0.011}}$  & $0.901_{\pm{0.027}}$ & $0.820_{\pm{0.075}}$ & 
            $0.454_{\pm{0.079}}$ & $0.751_{\pm{0.085}}$ & $0.828_{\pm{0.072}}$ & $0.937_{\pm{0.032}}$  \\
      
            & SML~\cite{Pei2023PPP} & 
            $0.901_{\pm{0.022}}$ & $0.920_{\pm{0.029}}$ & ${0.915}_{\pm{0.015}}$ & $0.802_{\pm{0.051}}$ & 
            $0.340_{\pm{0.125}}$ & $0.726_{\pm{0.038}}$ & $0.812_{\pm{0.058}}$ & ${0.967}_{\pm{0.015}}$ \\
      
            & PLM~\cite{Pei2023PPP} & 
            $0.911_{\pm{0.014}}$ & ${0.925}_{\pm{0.021}}$ & $0.910_{\pm{0.019}}$ & $0.791_{\pm{0.049}}$ & 
            $0.330_{\pm{0.105}}$ & $0.726_{\pm{0.058}}$ & $0.817_{\pm{0.056}}$ & $0.980_{\pm{0.008}}$ \\
                      
            & polyBERT~\cite{Kuenneth2022polyBERT} & 
            $0.875_{\pm{0.006}}$ & $0.844_{\pm{0.034}}$ & $0.880_{\pm{0.035}}$ & $0.705_{\pm{0.085}}$ & 
            $0.384_{\pm{0.066}}$ & $0.681_{\pm{0.058}}$ & $0.769_{\pm{0.034}}$ & $0.672_{\pm{0.119}}$ \\
  
            & Transpolymer~\cite{Xu2022TransPolymer} & 
            ${0.916}_{\pm{0.002}}$ & $0.911_{\pm{0.008}}$ & $0.902_{\pm{0.036}}$ & ${0.830}_{\pm{0.059}}$ & 
            ${0.430}_{\pm{0.058}}$ & ${0.744}_{\pm{0.075}}$ & ${0.826}_{\pm{0.071}}$ & $0.800_{\pm{0.172}}$ \\
            
            & MMPolymer~\cite{Wang2024MMPolymer} & 
            $0.924_{\pm{0.006}}$ & $0.934_{\pm{0.008}}$ & $0.925_{\pm{0.025}}$ & $0.836_{\pm{0.053}}$ & 
            $0.488_{\pm{0.072}}$ & $0.779_{\pm{0.052}}$ & $0.864_{\pm{0.036}}$ & $0.961_{\pm{0.018}}$ \\

            & MIPS (ours) & 
            $\mathbf{0.926}_{\pm{0.006}}$ & $\mathbf{0.945}_{\pm{0.007}}$ & $\mathbf{0.940}_{\pm{0.018}}$ & $\mathbf{0.846}_{\pm{0.051}}$ & 
            $\mathbf{0.506}_{\pm{0.074}}$ & $\mathbf{0.814}_{\pm{0.045}}$ & $\mathbf{0.877}_{\pm{0.046}}$ & $\mathbf{0.990}_{\pm{0.004}}$ \\
      \bottomrule
      \end{tabular}
    \end{adjustbox} 
    \label{table:main_exp}
\end{table*}%

\subsection{Preliminary Study}

\begin{figure}[h]
  \captionsetup[subfigure]{justification=centering}
  \centering
  \begin{subfigure}[b]{0.48\columnwidth}
    \centering
    \centerline{\includegraphics[width=\textwidth]{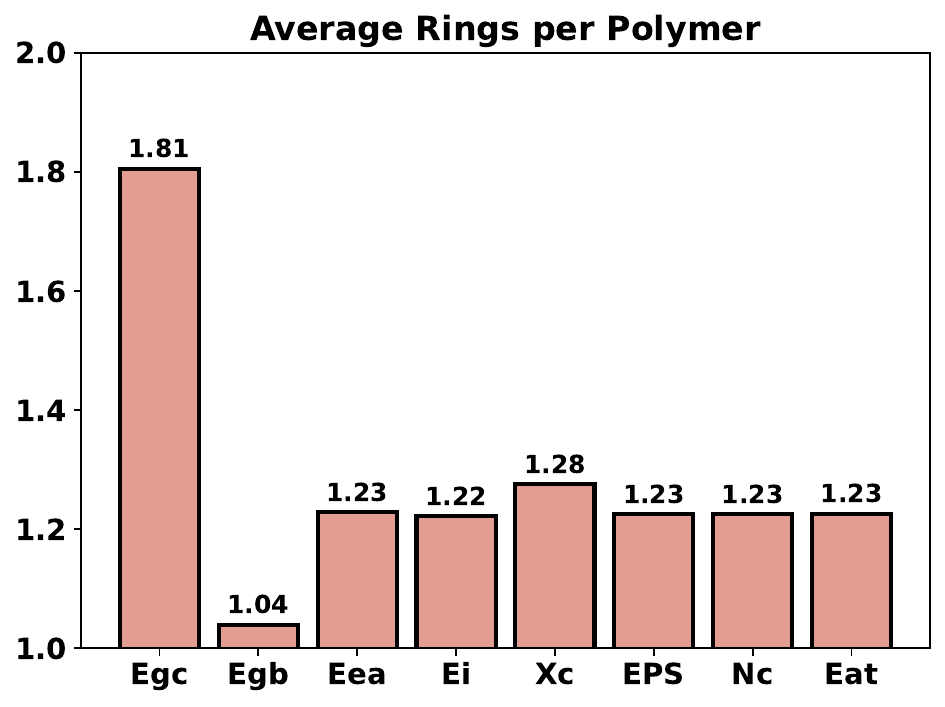}}
  \end{subfigure}
  \hfill
  \begin{subfigure}[b]{0.48\columnwidth}
    \centering
    \includegraphics[width=\textwidth]{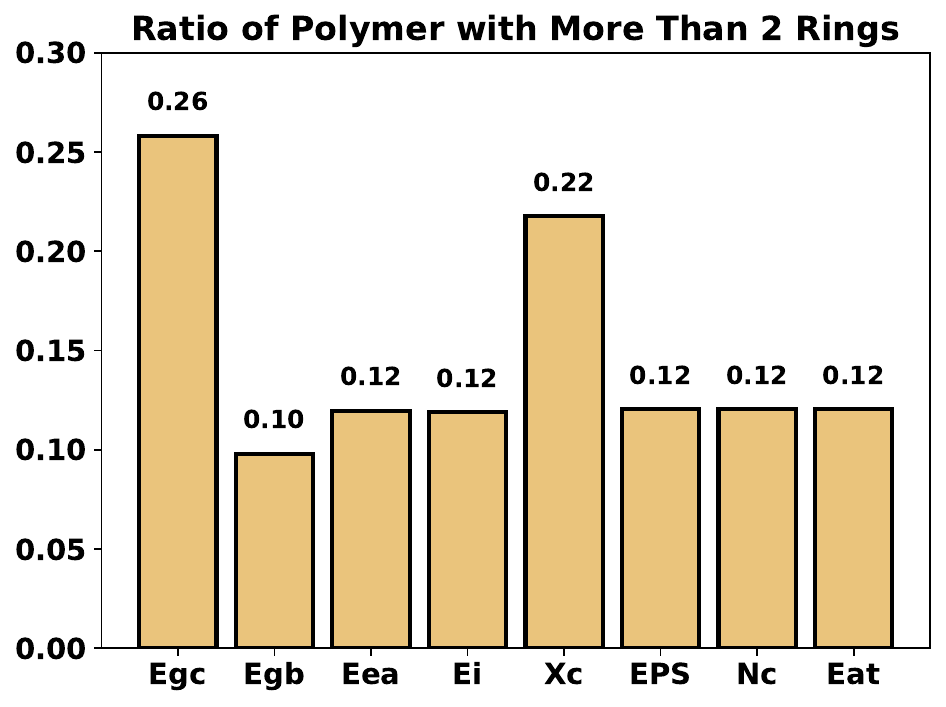}
  \end{subfigure}
  \caption{Ring statistics of eight polymer property datasets. 
  Left: Average number of rings per polymer. 
  Right: Ratio of polymers with more than two rings.}
  \label{fig:be_rings}
\end{figure}

We first conduct a preliminary study to demonstrate the effectiveness and robustness 
of our infinite polymer sequence modeling approach. 
For this purpose, we train a three-layer GIN model~\cite{Xu2019GIN} with 
"start keep", "star remove", "star substitution", and our proposed "star linking" strategies on eight polymer property prediction tasks. 
The model, referred to as GIN3-512, is configured with 512 hidden units. 
We use the same atom features $\mathbf{X}^{*}$ as in \citet{Li2023KPGT} to train GIN3-512 for 50 epochs with a batch size of 32 and learning rate of 0.001. 
Additionally, we evaluate the robustness of these strategie under RSIT, with results summarized in \cref{table:rsit}. 
We find that infinite polymer sequence modeling approach with "star linking" strategy 
outperforms the other three strategies on seven our of eight tasks without RSIT. 
Under RSIT, the performance of the three alternative strategies suffers significant degradation, 
while "star linking" strategy is resistant to RSIT, which suggests the robustness of our infinite polymer sequence modeling approach.
\paragraph*{Effectiveness of Backbone Embedding}
We further assess the effectiveness of backbone embedding (BE). 
Specifically, we compare GIN3-512 with and without backbone embedding across eight polymer property prediction tasks equiped with the "star linking" strategy. 
The results are shown in \cref{table:be}. 
Model with BE outperforms the model without BE on Egc and Xc datasets, 
and is on par with the model without BE on the other 6 datasets. 
To understand these results, we calcuate the average number of rings per polymer 
and the ratio of polymers with more than two rings across each dataset, as shown in \cref{fig:be_rings}. 
Notably, the Egc and Xc datasets exhibit the largest average number of rings per polymer 
and the largest ratio of polymers with more than two rings. 
Since backbone embedding is specifically designed to differentiate 
between the backbone structure and rings on the side chains, 
these ring statistics align with the observed model performance, 
highlighting the importance of backbone embedding for datasets with more ring structures.

\subsection{Multimodal Pre-training and Fine-tuning}
\paragraph*{Hyperparameters}
For pre-training, we use $L=6$ layers for the localized graph transformer and $d=512$ hidden units. 
The atom mask rate $p_{\text{mask}}$ is set to $0.3$. 
We optimize the model using Adam optimizer~\cite{Kingma2015Adam} with 
$\left(\beta_1, \beta_2\right)=\left(0.9, 0.999\right)$, a learning rate of $2e-4$, 
and a batch size of $1024$.
We pre-train the model for 20,000 steps with 2,000 warmup steps. 
In fine-tuning stage, the model is trained for $50$ epochs with a learning rate of $3e-5$ and a batch size of $32$. 
All experiments are conducted on a single NVIDIA $4090$ GPU. 

\begin{table*}[t]
  \centering
  \caption{2D: only use 2D graph with star linking; 
  2D + BE: add backbone embedding; 
  3D: only use 3D descriptors with a prediction head; 
  2D + BE + 3D: add backbone embedding and 3D descriptors. 
  $R^2$ is reported for each dataset.}
  \begin{adjustbox}{width=\textwidth,center}
  \begin{tabular}{@{}c|cccccccc@{}}
  \toprule
           & Egc         & Egb         & Eea         & Ei           & Xc          & EPS         & Nc          & Eat         \\ \midrule
  2D       & ${0.904}_{\pm 0.004}$ & ${0.932}_{\pm 0.007}$ & ${0.931}_{\pm 0.024}$ & ${0.835}_{\pm 0.044}$  & ${0.409}_{\pm 0.047}$ & ${0.803}_{\pm 0.070}$ & ${0.870}_{\pm 0.054}$ & ${0.980}_{\pm 0.011}$ \\
  2D + BE  & ${0.915}_{\pm 0.007}$ & ${0.934}_{\pm 0.011}$ & ${0.933}_{\pm 0.016}$ & ${0.836}_{\pm 0.056}$  & ${0.457}_{\pm 0.092}$ & ${0.806}_{\pm 0.065}$ & ${0.873}_{\pm 0.054}$ & ${0.988}_{\pm 0.003}$ \\
  3D       & ${0.886}_{\pm 0.010}$ & ${0.904}_{\pm 0.018}$ & ${0.868}_{\pm 0.027}$ & ${0.778}_{\pm 0.052}$  & ${0.419}_{\pm 0.044}$ & ${0.725}_{\pm 0.051}$ & ${0.822}_{\pm 0.031}$ & ${0.969}_{\pm 0.015}$ \\
  2D + BE + 3D & $\mathbf{0.926}_{\pm 0.006}$ & $\mathbf{0.945}_{\pm 0.007}$ & $\mathbf{0.940}_{\pm 0.018}$ & $\mathbf{0.846}_{\pm 0.051}$  & $\mathbf{0.506}_{\pm 0.074}$ & $\mathbf{0.814}_{\pm 0.045}$ & $\mathbf{0.877}_{\pm 0.046}$ & $\mathbf{0.990}_{\pm 0.004}$ \\ \bottomrule
  \end{tabular}
\end{adjustbox} 
\label{table:ablation_design}
\end{table*}

\paragraph*{Baseline Methods}
We compare our method against several state-of-the-art methods, including 
four molecular pre-training methods: ChemBERTa~\cite{Ahmad2022ChemBERT}, MolCLR~\cite{Wang2022MolCLR}, 
3D Infomax~\cite{Hannes2022Infomax}, Uni-Mol~\cite{Geng2023UniMol}, as well as
five polymer pre-training methods SML~\cite{Pei2023PPP}, PLM~\cite{Pei2023PPP}, polyBERT~\cite{Kuenneth2022polyBERT}, 
Transpolymer~\cite{Xu2022TransPolymer}, MMPolymer~\cite{Wang2024MMPolymer}. 
The baseline methods are set up following the default settings provided in their original references.
\paragraph*{Results}
The performance comparison of different methods across eight polymer property datasets is presented in \cref{table:main_exp}. 
It is clear that our MIPS outperforms the baseline methods on all datasets by considerable margins, 
especially on EPS and Eea datasets. 
Note that MMPolymer~\cite{Wang2024MMPolymer} also adopts the a multimodal pre-training strategy that 
integrates 1D P-SMILES and 3D geometric information. 
However, our MIPS framework stands out by directly modeling infinite polymer sequences, 
achieving superior performance. 
Additionally, we presente an analysis of the embedding generated by the pre-trained model in \SM, 
which reveals that the output embedding produced by our model show significant semantic 
alignment with the corresponding tasks. 


\section{Ablation and Interpretation Study}
In this section, 
we perform an ablation study to assess the contribution of various components in our model 
and explore the fragment-level interpretability.
Specifically, we evaluate the efficacy of the backbone embedding and the structure encoder. 
Due to space constraints, the analysis of "star linking" strategy, mask rate during the pre-training phase and the degree of localization in localized graph attention is provided in the \SM.

\paragraph*{Effectiveness of backbone embedding and 3D encoder} 
We evaluate the impact of backbone embedding and the 3D encoder 
by comparing model performance with and without these components, as shown in \cref{table:ablation_design}. 
Comparing the first two rows, we observe that incorporating backbone embedding leads to improved performance, 
particularly on the Egc and Xc datasets, which aligns with the observations in \cref{table:be}. 
Comparing the second and fourth rows, it is evident that the inclusion of the 3D encoder enhances performance across all datasets.
Furthermore, analyzing the last three rows reveals that combining both 2D and 3D encoders delivers the best overall performance, 
illustrating the synergistic role of the cross-modal fusion mechanism in MIPS.

\subsection{Model Interpretation by Fragment Analysis}
\begin{figure}[h]
  \captionsetup[subfigure]{justification=centering}
  \centering
  \begin{subfigure}[b]{0.98\columnwidth}
    \centering
    \centerline{\includegraphics[width=\textwidth]{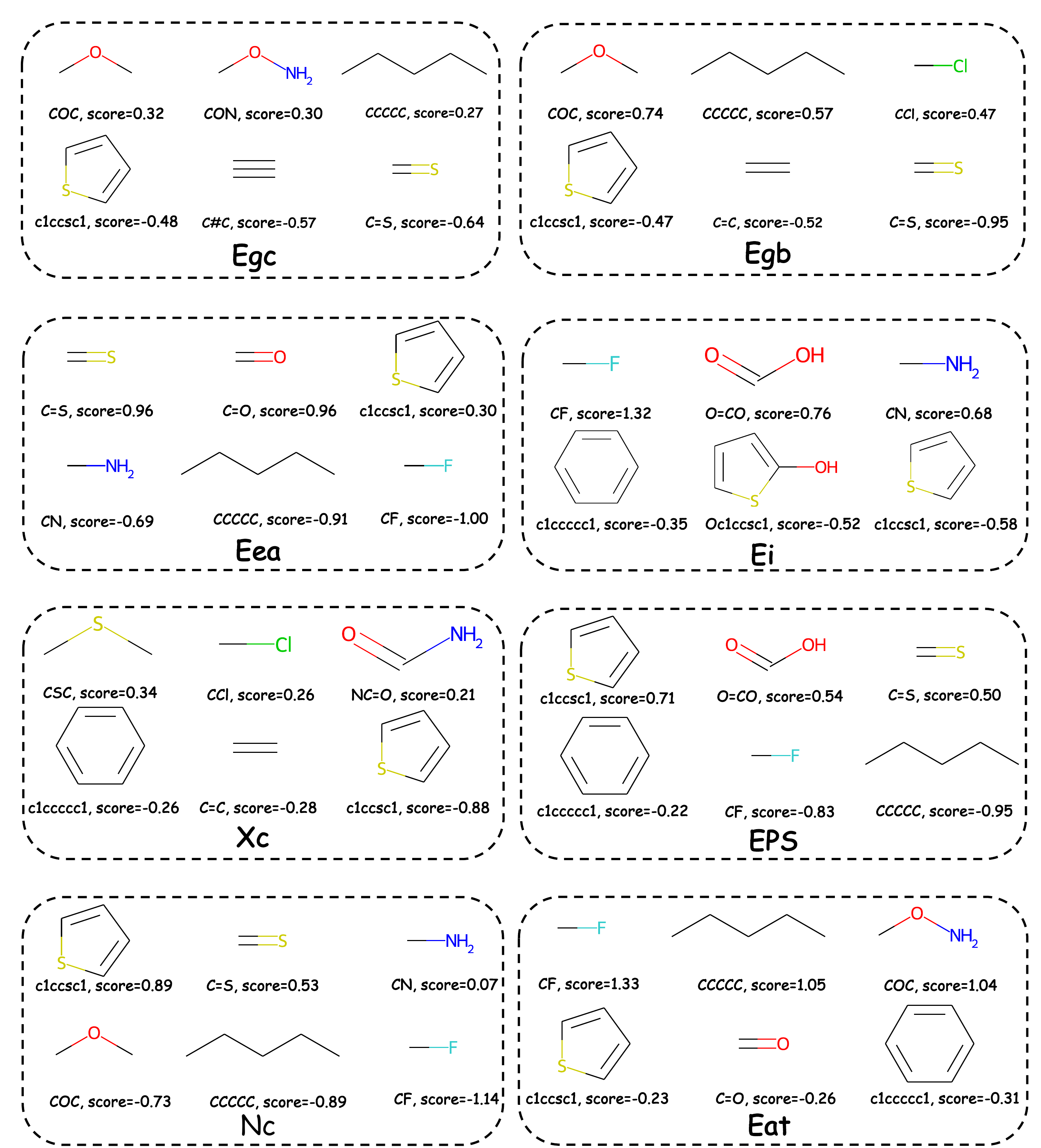}}
  \end{subfigure}
  \caption{Visualization of the top three and bottom three fragments with the highest and lowest importance scores on Egc, Ei, Xc, and Nc dataset.}
  \label{fig:fragment_analysis}
\end{figure}

We employ fragment analysis to interpret the model, 
identifying the relative importance of different fragments across each dataset. 
Principal subgraph mining with $1$-overlapping degree~\cite{Kong2022PSVAE,Wang2025Fragformer} 
with $100$ vocabulary size is used as the fragmentation method. 
The fragment library is constructed by the training set of each dataset. 
Additionally, we extend the fragment-level class activation mapping method~\cite{Wang2025Fragformer} to regression tasks, which 
determines the relative importance of fragments specific to each dataset. 
For visualization purposes, we illustrate the structure and contribution scores of the three fragments 
with the highest importance and the three with the lowest importance. 
Results are presented in \cref{fig:fragment_analysis}. 
We further analyze the consistency of fragment importance with chemical knowledge. 
we choose one representative dataset from each task type, namely the Egc, Xc, and Nc datasets. 
Due to the space limit, the details of fragment analysis and 
chemical investigation of Xc and Nc are provided in the \SM.

\paragraph*{Egc} The objective of this task is to predict the bandgap of polymer chains. 
Fragments such as "COC", "CON", and "CCCCC" show higher-than-average contribution scores to the bandgap (chain), 
whereas fragments like "c1ccsc1", "C\#C", and "C=S" exhibit lower-than-average scores. 
These findings align with chemical knowledge.
COC and CON typically exhibit localized electron density due to the electronegative oxygen or nitrogen atoms.
They stabilize the molecule but do not participate strongly in $\pi$-conjugation~\cite{Schwarze2016BandSE}. 
Consequently, their presence slightly increases the band gap.
"CCCCC"~(Linear Alkyl Chain) is nonpolar and non-conjugated~\cite{Hu2018EffectOA}. 
It does not contribute to the $\pi$-electron delocalization directly and slightly increases the band gap by reducing molecular excitability. 
On the other hand, "c1ccsc1", "C\#C, "C=S" actively contribute to $\pi$-conjugation~\cite{Liu2015ConjugatedPD}, 
which often leads to a reduction in the bandgap. 


\par 
From the analysis above, we conclude that the model can learn meaningful representations of polymers, 
which reflect the contributions of different functional groups.

\section{Conclusion} 
In this work, we introduce the Multimodal Infinite Polymer Sequence (MIPS) pre-training framework for polymer property prediction. 
Specifically, we propose a novel "star linking" strategy and backbone embedding to generalize the 
message passing and graph attention mechanisms to infinite polymer sequences with enhanced capability beyond Weisfeiler-Lehman test, 
which is provably robust under repeat and shift invariance test. 
Additionally, we design a cross-modal module to effectively fuse the topological and spatial information, 
enabling the model to learn multimodal representations via masked-atom prediction. 
Extensive experimental results across eight polymer property prediction tasks verify the efficacy of our method, 
which achieves the state-of-the-art performance. 
Ablation studies underscore the utility of the proposed modules, 
while the fragment analysis further provides evidence of the model's ability to capture chemically meaningful polymer representations.
\begin{acks}
This work was supported by Beijing Natural Science Foundation NO.L251072.
\end{acks}
\bibliographystyle{ACM-Reference-Format}
\bibliography{main}


\begin{thebibliography}{66}


\ifx \showCODEN    \undefined \def \showCODEN     #1{\unskip}     \fi
\ifx \showISBNx    \undefined \def \showISBNx     #1{\unskip}     \fi
\ifx \showISBNxiii \undefined \def \showISBNxiii  #1{\unskip}     \fi
\ifx \showISSN     \undefined \def \showISSN      #1{\unskip}     \fi
\ifx \showLCCN     \undefined \def \showLCCN      #1{\unskip}     \fi
\ifx \shownote     \undefined \def \shownote      #1{#1}          \fi
\ifx \showarticletitle \undefined \def \showarticletitle #1{#1}   \fi
\ifx \showURL      \undefined \def \showURL       {\relax}        \fi
\providecommand\bibfield[2]{#2}
\providecommand\bibinfo[2]{#2}
\providecommand\natexlab[1]{#1}
\providecommand\showeprint[2][]{arXiv:#2}

\bibitem[Agarap(2018)]%
        {Agarap2018DeepLU}
\bibfield{author}{\bibinfo{person}{Abien~Fred Agarap}.} \bibinfo{year}{2018}\natexlab{}.
\newblock \showarticletitle{Deep Learning using Rectified Linear Units (ReLU)}.
\newblock \bibinfo{journal}{\emph{CoRR}}  \bibinfo{volume}{abs/1803.08375} (\bibinfo{year}{2018}).
\newblock
\showeprint[arXiv]{1803.08375}
\urldef\tempurl%
\url{http://arxiv.org/abs/1803.08375}
\showURL{%
\tempurl}


\bibitem[Ahmad et~al\mbox{.}(2022)]%
        {Ahmad2022ChemBERT}
\bibfield{author}{\bibinfo{person}{Walid Ahmad}, \bibinfo{person}{Elana Simon}, \bibinfo{person}{Seyone Chithrananda}, \bibinfo{person}{Gabriel Grand}, {and} \bibinfo{person}{Bharath Ramsundar}.} \bibinfo{year}{2022}\natexlab{}.
\newblock \showarticletitle{ChemBERTa-2: Towards Chemical Foundation Models}.
\newblock \bibinfo{journal}{\emph{CoRR}}  \bibinfo{volume}{abs/2209.01712} (\bibinfo{year}{2022}).
\newblock
\href{https://doi.org/10.48550/ARXIV.2209.01712}{doi:\nolinkurl{10.48550/ARXIV.2209.01712}}
\showeprint[arXiv]{2209.01712}


\bibitem[Am{\'e}duri(2009)]%
        {Amduri2009FromVF}
\bibfield{author}{\bibinfo{person}{Bruno Am{\'e}duri}.} \bibinfo{year}{2009}\natexlab{}.
\newblock \showarticletitle{From vinylidene fluoride (VDF) to the applications of VDF-containing polymers and copolymers: recent developments and future trends.}
\newblock \bibinfo{journal}{\emph{Chemical reviews}}  \bibinfo{volume}{109 12} (\bibinfo{year}{2009}), \bibinfo{pages}{6632--86}.
\newblock
\urldef\tempurl%
\url{https://api.semanticscholar.org/CorpusID:206901877}
\showURL{%
\tempurl}


\bibitem[Awale and Reymond(2014)]%
        {Mahendra2014AtomPair}
\bibfield{author}{\bibinfo{person}{Mahendra Awale} {and} \bibinfo{person}{Jean{-}Louis Reymond}.} \bibinfo{year}{2014}\natexlab{}.
\newblock \showarticletitle{Atom Pair 2D-Fingerprints Perceive 3D-Molecular Shape and Pharmacophores for Very Fast Virtual Screening of {ZINC} and {GDB-17}}.
\newblock \bibinfo{journal}{\emph{J. Chem. Inf. Model.}} \bibinfo{volume}{54}, \bibinfo{number}{7} (\bibinfo{year}{2014}), \bibinfo{pages}{1892--1907}.
\newblock
\href{https://doi.org/10.1021/CI500232G}{doi:\nolinkurl{10.1021/CI500232G}}


\bibitem[Bagal et~al\mbox{.}(2022)]%
        {Bagal2022MolGPT}
\bibfield{author}{\bibinfo{person}{Viraj Bagal}, \bibinfo{person}{Rishal Aggarwal}, \bibinfo{person}{P.~K. Vinod}, {and} \bibinfo{person}{U.~Deva Priyakumar}.} \bibinfo{year}{2022}\natexlab{}.
\newblock \showarticletitle{MolGPT: Molecular Generation Using a Transformer-Decoder Model}.
\newblock \bibinfo{journal}{\emph{J. Chem. Inf. Model.}} \bibinfo{volume}{62}, \bibinfo{number}{9} (\bibinfo{year}{2022}), \bibinfo{pages}{2064--2076}.
\newblock
\href{https://doi.org/10.1021/ACS.JCIM.1C00600}{doi:\nolinkurl{10.1021/ACS.JCIM.1C00600}}


\bibitem[Bicerano(1996)]%
        {Bicerano1996PredictionOP}
\bibfield{author}{\bibinfo{person}{Jozef Bicerano}.} \bibinfo{year}{1996}\natexlab{}.
\newblock \showarticletitle{Prediction of Polymer Properties}.
\newblock
\urldef\tempurl%
\url{https://api.semanticscholar.org/CorpusID:138697246}
\showURL{%
\tempurl}


\bibitem[Bouritsas et~al\mbox{.}(2023)]%
        {Giorgos2023GNNCounting}
\bibfield{author}{\bibinfo{person}{Giorgos Bouritsas}, \bibinfo{person}{Fabrizio Frasca}, \bibinfo{person}{Stefanos Zafeiriou}, {and} \bibinfo{person}{Michael~M. Bronstein}.} \bibinfo{year}{2023}\natexlab{}.
\newblock \showarticletitle{Improving Graph Neural Network Expressivity via Subgraph Isomorphism Counting}.
\newblock \bibinfo{journal}{\emph{{IEEE} Trans. Pattern Anal. Mach. Intell.}} \bibinfo{volume}{45}, \bibinfo{number}{1} (\bibinfo{year}{2023}), \bibinfo{pages}{657--668}.
\newblock
\href{https://doi.org/10.1109/TPAMI.2022.3154319}{doi:\nolinkurl{10.1109/TPAMI.2022.3154319}}


\bibitem[Breiman(2001)]%
        {Leo2001RF}
\bibfield{author}{\bibinfo{person}{Leo Breiman}.} \bibinfo{year}{2001}\natexlab{}.
\newblock \showarticletitle{Random Forests}.
\newblock \bibinfo{journal}{\emph{Mach. Learn.}} \bibinfo{volume}{45}, \bibinfo{number}{1} (\bibinfo{year}{2001}), \bibinfo{pages}{5--32}.
\newblock
\href{https://doi.org/10.1023/A:1010933404324}{doi:\nolinkurl{10.1023/A:1010933404324}}


\bibitem[Chen et~al\mbox{.}(2020)]%
        {Ting2020SimCLR}
\bibfield{author}{\bibinfo{person}{Ting Chen}, \bibinfo{person}{Simon Kornblith}, \bibinfo{person}{Mohammad Norouzi}, {and} \bibinfo{person}{Geoffrey~E. Hinton}.} \bibinfo{year}{2020}\natexlab{}.
\newblock \showarticletitle{A Simple Framework for Contrastive Learning of Visual Representations}. In \bibinfo{booktitle}{\emph{Proceedings of the 37th International Conference on Machine Learning, {ICML} 2020, 13-18 July 2020, Virtual Event}} \emph{(\bibinfo{series}{Proceedings of Machine Learning Research}, Vol.~\bibinfo{volume}{119})}. \bibinfo{publisher}{{PMLR}}, \bibinfo{pages}{1597--1607}.
\newblock
\urldef\tempurl%
\url{http://proceedings.mlr.press/v119/chen20j.html}
\showURL{%
\tempurl}


\bibitem[Cortes and Vapnik(1995)]%
        {Cortes1995SVM}
\bibfield{author}{\bibinfo{person}{Corinna Cortes} {and} \bibinfo{person}{Vladimir Vapnik}.} \bibinfo{year}{1995}\natexlab{}.
\newblock \showarticletitle{Support-Vector Networks}.
\newblock \bibinfo{journal}{\emph{Mach. Learn.}} \bibinfo{volume}{20}, \bibinfo{number}{3} (\bibinfo{year}{1995}), \bibinfo{pages}{273--297}.
\newblock
\href{https://doi.org/10.1007/BF00994018}{doi:\nolinkurl{10.1007/BF00994018}}


\bibitem[Devlin et~al\mbox{.}(2019)]%
        {Devlin2019Bert}
\bibfield{author}{\bibinfo{person}{Jacob Devlin}, \bibinfo{person}{Ming{-}Wei Chang}, \bibinfo{person}{Kenton Lee}, {and} \bibinfo{person}{Kristina Toutanova}.} \bibinfo{year}{2019}\natexlab{}.
\newblock \showarticletitle{{BERT:} Pre-training of Deep Bidirectional Transformers for Language Understanding}. In \bibinfo{booktitle}{\emph{Proceedings of the 2019 Conference of the North American Chapter of the Association for Computational Linguistics: Human Language Technologies, {NAACL-HLT} 2019, Minneapolis, MN, USA, June 2-7, 2019, Volume 1 (Long and Short Papers)}}, \bibfield{editor}{\bibinfo{person}{Jill Burstein}, \bibinfo{person}{Christy Doran}, {and} \bibinfo{person}{Thamar Solorio}} (Eds.). \bibinfo{publisher}{Association for Computational Linguistics}, \bibinfo{pages}{4171--4186}.
\newblock
\href{https://doi.org/10.18653/V1/N19-1423}{doi:\nolinkurl{10.18653/V1/N19-1423}}


\bibitem[Fabian et~al\mbox{.}(2020)]%
        {Fabian2020MolBERT}
\bibfield{author}{\bibinfo{person}{Benedek Fabian}, \bibinfo{person}{Thomas Edlich}, \bibinfo{person}{H{\'{e}}l{\'{e}}na Gaspar}, \bibinfo{person}{Marwin H.~S. Segler}, \bibinfo{person}{Joshua Meyers}, \bibinfo{person}{Marco Fiscato}, {and} \bibinfo{person}{Mohamed Ahmed}.} \bibinfo{year}{2020}\natexlab{}.
\newblock \showarticletitle{Molecular representation learning with language models and domain-relevant auxiliary tasks}.
\newblock \bibinfo{journal}{\emph{CoRR}}  \bibinfo{volume}{abs/2011.13230} (\bibinfo{year}{2020}).
\newblock
\showeprint[arXiv]{2011.13230}
\urldef\tempurl%
\url{https://arxiv.org/abs/2011.13230}
\showURL{%
\tempurl}


\bibitem[Fried(2014)]%
        {Fried2014Polymer}
\bibfield{author}{\bibinfo{person}{Joel~R Fried}.} \bibinfo{year}{2014}\natexlab{}.
\newblock \bibinfo{booktitle}{\emph{Polymer science and technology}}.
\newblock \bibinfo{publisher}{Pearson Education}.
\newblock


\bibitem[Hamilton et~al\mbox{.}(2017)]%
        {Hamilton2017GraphSAGE}
\bibfield{author}{\bibinfo{person}{William~L. Hamilton}, \bibinfo{person}{Zhitao Ying}, {and} \bibinfo{person}{Jure Leskovec}.} \bibinfo{year}{2017}\natexlab{}.
\newblock \showarticletitle{Inductive Representation Learning on Large Graphs}. In \bibinfo{booktitle}{\emph{Advances in Neural Information Processing Systems 30: Annual Conference on Neural Information Processing Systems 2017, December 4-9, 2017, Long Beach, CA, {USA}}}, \bibfield{editor}{\bibinfo{person}{Isabelle Guyon}, \bibinfo{person}{Ulrike von Luxburg}, \bibinfo{person}{Samy Bengio}, \bibinfo{person}{Hanna~M. Wallach}, \bibinfo{person}{Rob Fergus}, \bibinfo{person}{S.~V.~N. Vishwanathan}, {and} \bibinfo{person}{Roman Garnett}} (Eds.). \bibinfo{pages}{1024--1034}.
\newblock
\urldef\tempurl%
\url{https://proceedings.neurips.cc/paper/2017/hash/5dd9db5e033da9c6fb5ba83c7a7ebea9-Abstract.html}
\showURL{%
\tempurl}


\bibitem[Higashihara and Ueda(2015)]%
        {Higashihara2015RecentPI}
\bibfield{author}{\bibinfo{person}{Tomoya Higashihara} {and} \bibinfo{person}{Mitsuru Ueda}.} \bibinfo{year}{2015}\natexlab{}.
\newblock \showarticletitle{Recent Progress in High Refractive Index Polymers}.
\newblock \bibinfo{journal}{\emph{Macromolecules}}  \bibinfo{volume}{48} (\bibinfo{year}{2015}), \bibinfo{pages}{1915--1929}.
\newblock
\urldef\tempurl%
\url{https://api.semanticscholar.org/CorpusID:102023070}
\showURL{%
\tempurl}


\bibitem[Hu et~al\mbox{.}(2018)]%
        {Hu2018EffectOA}
\bibfield{author}{\bibinfo{person}{Yuanyuan Hu}, \bibinfo{person}{David~Xi Cao}, \bibinfo{person}{Alexander~T Lill}, \bibinfo{person}{Lang Jiang}, \bibinfo{person}{Chong‐an Di}, \bibinfo{person}{Xike Gao}, \bibinfo{person}{Henning Sirringhaus}, {and} \bibinfo{person}{Thuc‐Quyen Nguyen}.} \bibinfo{year}{2018}\natexlab{}.
\newblock \showarticletitle{Effect of Alkyl‐Chain Length on Charge Transport Properties of Organic Semiconductors and Organic Field‐Effect Transistors}.
\newblock \bibinfo{journal}{\emph{Advanced Electronic Materials}}  \bibinfo{volume}{4} (\bibinfo{year}{2018}).
\newblock
\urldef\tempurl%
\url{https://api.semanticscholar.org/CorpusID:103005072}
\showURL{%
\tempurl}


\bibitem[Jolliffe(1982)]%
        {Jolliffe1982PCR}
\bibfield{author}{\bibinfo{person}{Ian~T. Jolliffe}.} \bibinfo{year}{1982}\natexlab{}.
\newblock \showarticletitle{A Note on the Use of Principal Components in Regression}.
\newblock \bibinfo{journal}{\emph{Journal of The Royal Statistical Society Series C-applied Statistics}}  \bibinfo{volume}{31} (\bibinfo{year}{1982}), \bibinfo{pages}{300--303}.
\newblock
\urldef\tempurl%
\url{https://api.semanticscholar.org/CorpusID:126082585}
\showURL{%
\tempurl}


\bibitem[Kingma and Ba(2015)]%
        {Kingma2015Adam}
\bibfield{author}{\bibinfo{person}{Diederik~P. Kingma} {and} \bibinfo{person}{Jimmy Ba}.} \bibinfo{year}{2015}\natexlab{}.
\newblock \showarticletitle{Adam: {A} Method for Stochastic Optimization}. In \bibinfo{booktitle}{\emph{3rd International Conference on Learning Representations, {ICLR} 2015, San Diego, CA, USA, May 7-9, 2015, Conference Track Proceedings}}, \bibfield{editor}{\bibinfo{person}{Yoshua Bengio} {and} \bibinfo{person}{Yann LeCun}} (Eds.).
\newblock
\urldef\tempurl%
\url{http://arxiv.org/abs/1412.6980}
\showURL{%
\tempurl}


\bibitem[Kipf and Welling(2017)]%
        {Kipf2017GCN}
\bibfield{author}{\bibinfo{person}{Thomas~N. Kipf} {and} \bibinfo{person}{Max Welling}.} \bibinfo{year}{2017}\natexlab{}.
\newblock \showarticletitle{Semi-Supervised Classification with Graph Convolutional Networks}. In \bibinfo{booktitle}{\emph{5th International Conference on Learning Representations, {ICLR} 2017, Toulon, France, April 24-26, 2017, Conference Track Proceedings}}. \bibinfo{publisher}{OpenReview.net}.
\newblock
\urldef\tempurl%
\url{https://openreview.net/forum?id=SJU4ayYgl}
\showURL{%
\tempurl}


\bibitem[Kong et~al\mbox{.}(2022)]%
        {Kong2022PSVAE}
\bibfield{author}{\bibinfo{person}{Xiangzhe Kong}, \bibinfo{person}{Wenbing Huang}, \bibinfo{person}{Zhixing Tan}, {and} \bibinfo{person}{Yang Liu}.} \bibinfo{year}{2022}\natexlab{}.
\newblock \showarticletitle{Molecule Generation by Principal Subgraph Mining and Assembling}. In \bibinfo{booktitle}{\emph{Advances in Neural Information Processing Systems 35: Annual Conference on Neural Information Processing Systems 2022, NeurIPS 2022, New Orleans, LA, USA, November 28 - December 9, 2022}}, \bibfield{editor}{\bibinfo{person}{Sanmi Koyejo}, \bibinfo{person}{S.~Mohamed}, \bibinfo{person}{A.~Agarwal}, \bibinfo{person}{Danielle Belgrave}, \bibinfo{person}{K.~Cho}, {and} \bibinfo{person}{A.~Oh}} (Eds.).
\newblock
\urldef\tempurl%
\url{http://papers.nips.cc/paper\_files/paper/2022/hash/1160792eab11de2bbaf9e71fce191e8c-Abstract-Conference.html}
\showURL{%
\tempurl}


\bibitem[Kuenneth and Ramprasad(2022)]%
        {Kuenneth2022polyBERT}
\bibfield{author}{\bibinfo{person}{Christopher Kuenneth} {and} \bibinfo{person}{Rampi Ramprasad}.} \bibinfo{year}{2022}\natexlab{}.
\newblock \showarticletitle{polyBERT: a chemical language model to enable fully machine-driven ultrafast polymer informatics}.
\newblock \bibinfo{journal}{\emph{Nature Communications}}  \bibinfo{volume}{14} (\bibinfo{year}{2022}).
\newblock
\urldef\tempurl%
\url{https://api.semanticscholar.org/CorpusID:252595691}
\showURL{%
\tempurl}


\bibitem[K{\"{u}}nneth et~al\mbox{.}(2021)]%
        {Kunneth2021Polymer}
\bibfield{author}{\bibinfo{person}{Christopher K{\"{u}}nneth}, \bibinfo{person}{Arunkumar~Chitteth Rajan}, \bibinfo{person}{Huan Tran}, \bibinfo{person}{Lihua Chen}, \bibinfo{person}{Chiho Kim}, {and} \bibinfo{person}{Rampi Ramprasad}.} \bibinfo{year}{2021}\natexlab{}.
\newblock \showarticletitle{Polymer informatics with multi-task learning}.
\newblock \bibinfo{journal}{\emph{Patterns}} \bibinfo{volume}{2}, \bibinfo{number}{4} (\bibinfo{year}{2021}), \bibinfo{pages}{100238}.
\newblock
\href{https://doi.org/10.1016/J.PATTER.2021.100238}{doi:\nolinkurl{10.1016/J.PATTER.2021.100238}}


\bibitem[Landrum(2016)]%
        {Landrum2016RDKit}
\bibfield{author}{\bibinfo{person}{Greg Landrum}.} \bibinfo{year}{2016}\natexlab{}.
\newblock \showarticletitle{RDKit: Open-Source Cheminformatics Software}.
\newblock  (\bibinfo{year}{2016}).
\newblock
\urldef\tempurl%
\url{https://github.com/rdkit/rdkit/releases/tag/Release_2016_09_4}
\showURL{%
\tempurl}


\bibitem[Le et~al\mbox{.}(2012)]%
        {Le2012QSPR}
\bibfield{author}{\bibinfo{person}{Tu~C. Le}, \bibinfo{person}{Vidana~Chandana Epa}, \bibinfo{person}{Frank~R. Burden}, {and} \bibinfo{person}{Dave Winkler}.} \bibinfo{year}{2012}\natexlab{}.
\newblock \showarticletitle{Quantitative structure-property relationship modeling of diverse materials properties.}
\newblock \bibinfo{journal}{\emph{Chemical reviews}}  \bibinfo{volume}{112 5} (\bibinfo{year}{2012}), \bibinfo{pages}{2889--919}.
\newblock
\urldef\tempurl%
\url{https://api.semanticscholar.org/CorpusID:30285330}
\showURL{%
\tempurl}


\bibitem[LeCun et~al\mbox{.}(2015)]%
        {LeCun2015DL}
\bibfield{author}{\bibinfo{person}{Yann LeCun}, \bibinfo{person}{Yoshua Bengio}, {and} \bibinfo{person}{Geoffrey~E. Hinton}.} \bibinfo{year}{2015}\natexlab{}.
\newblock \showarticletitle{Deep learning}.
\newblock \bibinfo{journal}{\emph{Nat.}} \bibinfo{volume}{521}, \bibinfo{number}{7553} (\bibinfo{year}{2015}), \bibinfo{pages}{436--444}.
\newblock
\href{https://doi.org/10.1038/NATURE14539}{doi:\nolinkurl{10.1038/NATURE14539}}


\bibitem[Li et~al\mbox{.}(2023)]%
        {Li2023KPGT}
\bibfield{author}{\bibinfo{person}{Han Li}, \bibinfo{person}{Ruotian Zhang}, \bibinfo{person}{Min Yaosen}, \bibinfo{person}{Dacheng Ma}, \bibinfo{person}{Dan Zhao}, {and} \bibinfo{person}{Jianyang Zeng}.} \bibinfo{year}{2023}\natexlab{}.
\newblock \showarticletitle{A knowledge-guided pre-training framework for improving molecular representation learning}.
\newblock \bibinfo{journal}{\emph{Nature Communications}}  \bibinfo{volume}{14} (\bibinfo{date}{11} \bibinfo{year}{2023}).
\newblock
\href{https://doi.org/10.1038/s41467-023-43214-1}{doi:\nolinkurl{10.1038/s41467-023-43214-1}}


\bibitem[Lin et~al\mbox{.}(2024)]%
        {Lin2024CrossView}
\bibfield{author}{\bibinfo{person}{Junyu Lin}, \bibinfo{person}{Yan Zheng}, \bibinfo{person}{Xinyue Chen}, \bibinfo{person}{Yazhou Ren}, \bibinfo{person}{Xiaorong Pu}, {and} \bibinfo{person}{Jing He}.} \bibinfo{year}{2024}\natexlab{}.
\newblock \showarticletitle{Cross-view Contrastive Unification Guides Generative Pretraining for Molecular Property Prediction}. In \bibinfo{booktitle}{\emph{Proceedings of the 32nd {ACM} International Conference on Multimedia, {MM} 2024, Melbourne, VIC, Australia, 28 October 2024 - 1 November 2024}}, \bibfield{editor}{\bibinfo{person}{Jianfei Cai}, \bibinfo{person}{Mohan~S. Kankanhalli}, \bibinfo{person}{Balakrishnan Prabhakaran}, \bibinfo{person}{Susanne Boll}, \bibinfo{person}{Ramanathan Subramanian}, \bibinfo{person}{Liang Zheng}, \bibinfo{person}{Vivek~K. Singh}, \bibinfo{person}{Pablo C{\'{e}}sar}, \bibinfo{person}{Lexing Xie}, {and} \bibinfo{person}{Dong Xu}} (Eds.). \bibinfo{publisher}{{ACM}}, \bibinfo{pages}{2108--2116}.
\newblock
\href{https://doi.org/10.1145/3664647.3681193}{doi:\nolinkurl{10.1145/3664647.3681193}}


\bibitem[Liu et~al\mbox{.}(2015)]%
        {Liu2015ConjugatedPD}
\bibfield{author}{\bibinfo{person}{Yajing Liu}, \bibinfo{person}{Jacky Wing~Yip Lam}, {and} \bibinfo{person}{Ben~Zhong Tang}.} \bibinfo{year}{2015}\natexlab{}.
\newblock \showarticletitle{Conjugated polymers developed from alkynes}.
\newblock \bibinfo{journal}{\emph{National Science Review}}  \bibinfo{volume}{2} (\bibinfo{year}{2015}), \bibinfo{pages}{493--509}.
\newblock
\urldef\tempurl%
\url{https://api.semanticscholar.org/CorpusID:88351303}
\showURL{%
\tempurl}


\bibitem[Luo et~al\mbox{.}(2018)]%
        {Luo2018PolymerEnergy1}
\bibfield{author}{\bibinfo{person}{Hang Luo}, \bibinfo{person}{Sheng Chen}, \bibinfo{person}{Lihong Liu}, \bibinfo{person}{Xuefan Zhou}, \bibinfo{person}{Chao Ma}, \bibinfo{person}{Weiwei Liu}, {and} \bibinfo{person}{Dou Zhang}.} \bibinfo{year}{2018}\natexlab{}.
\newblock \showarticletitle{Core–Shell Nanostructure Design in Polymer Nanocomposite Capacitors for Energy Storage Applications}.
\newblock \bibinfo{journal}{\emph{ACS Sustainable Chemistry \& Engineering}} (\bibinfo{year}{2018}).
\newblock
\urldef\tempurl%
\url{https://api.semanticscholar.org/CorpusID:104416054}
\showURL{%
\tempurl}


\bibitem[Ma and Luo(2020)]%
        {Ruimin2020PI1M}
\bibfield{author}{\bibinfo{person}{Ruimin Ma} {and} \bibinfo{person}{Tengfei Luo}.} \bibinfo{year}{2020}\natexlab{}.
\newblock \showarticletitle{{PI1M:} {A} Benchmark Database for Polymer Informatics}.
\newblock \bibinfo{journal}{\emph{J. Chem. Inf. Model.}} \bibinfo{volume}{60}, \bibinfo{number}{10} (\bibinfo{year}{2020}), \bibinfo{pages}{4684--4690}.
\newblock
\href{https://doi.org/10.1021/ACS.JCIM.0C00726}{doi:\nolinkurl{10.1021/ACS.JCIM.0C00726}}


\bibitem[Mark(2007)]%
        {Mark2007PhysicalPO}
\bibfield{author}{\bibinfo{person}{James~E. Mark}.} \bibinfo{year}{2007}\natexlab{}.
\newblock \showarticletitle{Physical properties of polymers handbook}.
\newblock
\urldef\tempurl%
\url{https://api.semanticscholar.org/CorpusID:135747818}
\showURL{%
\tempurl}


\bibitem[McInnes and Healy(2018)]%
        {McInnes2018UMAP}
\bibfield{author}{\bibinfo{person}{Leland McInnes} {and} \bibinfo{person}{John Healy}.} \bibinfo{year}{2018}\natexlab{}.
\newblock \showarticletitle{{UMAP:} Uniform Manifold Approximation and Projection for Dimension Reduction}.
\newblock \bibinfo{journal}{\emph{CoRR}}  \bibinfo{volume}{abs/1802.03426} (\bibinfo{year}{2018}).
\newblock
\showeprint[arXiv]{1802.03426}
\urldef\tempurl%
\url{http://arxiv.org/abs/1802.03426}
\showURL{%
\tempurl}


\bibitem[Morris et~al\mbox{.}(2019)]%
        {Morris2019HigherOrderGNN}
\bibfield{author}{\bibinfo{person}{Christopher Morris}, \bibinfo{person}{Martin Ritzert}, \bibinfo{person}{Matthias Fey}, \bibinfo{person}{William~L. Hamilton}, \bibinfo{person}{Jan~Eric Lenssen}, \bibinfo{person}{Gaurav Rattan}, {and} \bibinfo{person}{Martin Grohe}.} \bibinfo{year}{2019}\natexlab{}.
\newblock \showarticletitle{Weisfeiler and Leman Go Neural: Higher-Order Graph Neural Networks}. In \bibinfo{booktitle}{\emph{The Thirty-Third {AAAI} Conference on Artificial Intelligence, {AAAI} 2019, The Thirty-First Innovative Applications of Artificial Intelligence Conference, {IAAI} 2019, The Ninth {AAAI} Symposium on Educational Advances in Artificial Intelligence, {EAAI} 2019, Honolulu, Hawaii, USA, January 27 - February 1, 2019}}. \bibinfo{publisher}{{AAAI} Press}, \bibinfo{pages}{4602--4609}.
\newblock
\href{https://doi.org/10.1609/AAAI.V33I01.33014602}{doi:\nolinkurl{10.1609/AAAI.V33I01.33014602}}


\bibitem[Munshi et~al\mbox{.}(2021)]%
        {Munshi2021PolymerOptic}
\bibfield{author}{\bibinfo{person}{Joydeep Munshi}, \bibinfo{person}{Wei Chen}, \bibinfo{person}{TeYu Chien}, {and} \bibinfo{person}{Ganesh Balasubramanian}.} \bibinfo{year}{2021}\natexlab{}.
\newblock \showarticletitle{Transfer Learned Designer Polymers For Organic Solar Cells}.
\newblock \bibinfo{journal}{\emph{Journal of chemical information and modeling}} (\bibinfo{year}{2021}).
\newblock
\urldef\tempurl%
\url{https://api.semanticscholar.org/CorpusID:230824254}
\showURL{%
\tempurl}


\bibitem[Namazi(2017)]%
        {Namazi2017PolymerLife}
\bibfield{author}{\bibinfo{person}{Hassan Namazi}.} \bibinfo{year}{2017}\natexlab{}.
\newblock \showarticletitle{Polymers in our daily life}.
\newblock \bibinfo{journal}{\emph{BioImpacts : BI}}  \bibinfo{volume}{7} (\bibinfo{year}{2017}), \bibinfo{pages}{73 -- 74}.
\newblock
\urldef\tempurl%
\url{https://api.semanticscholar.org/CorpusID:29251347}
\showURL{%
\tempurl}


\bibitem[Odian(2004)]%
        {Odian2004Polymer}
\bibfield{author}{\bibinfo{person}{George Odian}.} \bibinfo{year}{2004}\natexlab{}.
\newblock \bibinfo{booktitle}{\emph{Principles of polymerization}}.
\newblock \bibinfo{publisher}{John Wiley \& Sons}.
\newblock


\bibitem[Orio et~al\mbox{.}(2009)]%
        {Orio2009DFT}
\bibfield{author}{\bibinfo{person}{Maylis Orio}, \bibinfo{person}{Dimitrios~A. Pantazis}, {and} \bibinfo{person}{Frank Neese}.} \bibinfo{year}{2009}\natexlab{}.
\newblock \showarticletitle{Density functional theory}.
\newblock \bibinfo{journal}{\emph{Photosynthesis Research}}  \bibinfo{volume}{102} (\bibinfo{year}{2009}), \bibinfo{pages}{443 -- 453}.
\newblock
\urldef\tempurl%
\url{https://api.semanticscholar.org/CorpusID:5677752}
\showURL{%
\tempurl}


\bibitem[Painter and Coleman(2008)]%
        {Painter2008EssentialsOP}
\bibfield{author}{\bibinfo{person}{Paul~C. Painter} {and} \bibinfo{person}{Michael~M. Coleman}.} \bibinfo{year}{2008}\natexlab{}.
\newblock \showarticletitle{Essentials of Polymer Science and Engineering}.
\newblock
\urldef\tempurl%
\url{https://api.semanticscholar.org/CorpusID:60462767}
\showURL{%
\tempurl}


\bibitem[Pasut and Veronese(2007)]%
        {Pasut2007PolymerDrug}
\bibfield{author}{\bibinfo{person}{Gianfranco Pasut} {and} \bibinfo{person}{F.~M. Veronese}.} \bibinfo{year}{2007}\natexlab{}.
\newblock \showarticletitle{Polymer-drug conjugation, recent achievements and general strategies}.
\newblock \bibinfo{journal}{\emph{Progress in Polymer Science}}  \bibinfo{volume}{32} (\bibinfo{year}{2007}), \bibinfo{pages}{933--961}.
\newblock
\urldef\tempurl%
\url{https://api.semanticscholar.org/CorpusID:95522376}
\showURL{%
\tempurl}


\bibitem[Rahman et~al\mbox{.}(2021)]%
        {Rahman2021AML}
\bibfield{author}{\bibinfo{person}{Aowabin Rahman}, \bibinfo{person}{Prathamesh~Prashant Deshpande}, \bibinfo{person}{Matthew~S. Radue}, \bibinfo{person}{Gregory~M. Odegard}, \bibinfo{person}{S. Gowtham}, \bibinfo{person}{Susanta Ghosh}, {and} \bibinfo{person}{Ashley~D. Spear}.} \bibinfo{year}{2021}\natexlab{}.
\newblock \showarticletitle{A machine learning framework for predicting the shear strength of carbon nanotube-polymer interfaces based on molecular dynamics simulation data}.
\newblock \bibinfo{journal}{\emph{Composites Science and Technology}}  \bibinfo{volume}{207} (\bibinfo{year}{2021}), \bibinfo{pages}{108627}.
\newblock
\urldef\tempurl%
\url{https://api.semanticscholar.org/CorpusID:233545965}
\showURL{%
\tempurl}


\bibitem[Richtering(2003)]%
        {Richtering2003PolymerP}
\bibfield{author}{\bibinfo{person}{Walter Richtering}.} \bibinfo{year}{2003}\natexlab{}.
\newblock \showarticletitle{Polymer Physics}.
\newblock \bibinfo{journal}{\emph{Applied Rheology}}  \bibinfo{volume}{13} (\bibinfo{year}{2003}), \bibinfo{pages}{172 -- 173}.
\newblock
\urldef\tempurl%
\url{https://api.semanticscholar.org/CorpusID:202047447}
\showURL{%
\tempurl}


\bibitem[Rong et~al\mbox{.}(2020)]%
        {Rong2020GROVER}
\bibfield{author}{\bibinfo{person}{Yu Rong}, \bibinfo{person}{Yatao Bian}, \bibinfo{person}{Tingyang Xu}, \bibinfo{person}{Weiyang Xie}, \bibinfo{person}{Ying Wei}, \bibinfo{person}{Wenbing Huang}, {and} \bibinfo{person}{Junzhou Huang}.} \bibinfo{year}{2020}\natexlab{}.
\newblock \showarticletitle{Self-Supervised Graph Transformer on Large-Scale Molecular Data}. In \bibinfo{booktitle}{\emph{Advances in Neural Information Processing Systems 33: Annual Conference on Neural Information Processing Systems 2020, NeurIPS 2020, December 6-12, 2020, virtual}}, \bibfield{editor}{\bibinfo{person}{Hugo Larochelle}, \bibinfo{person}{Marc'Aurelio Ranzato}, \bibinfo{person}{Raia Hadsell}, \bibinfo{person}{Maria{-}Florina Balcan}, {and} \bibinfo{person}{Hsuan{-}Tien Lin}} (Eds.).
\newblock
\urldef\tempurl%
\url{https://proceedings.neurips.cc/paper/2020/hash/94aef38441efa3380a3bed3faf1f9d5d-Abstract.html}
\showURL{%
\tempurl}


\bibitem[Satorras et~al\mbox{.}(2021)]%
        {Satorras2021EGNN}
\bibfield{author}{\bibinfo{person}{Victor~Garcia Satorras}, \bibinfo{person}{Emiel Hoogeboom}, {and} \bibinfo{person}{Max Welling}.} \bibinfo{year}{2021}\natexlab{}.
\newblock \showarticletitle{E(n) Equivariant Graph Neural Networks}.
\newblock \bibinfo{journal}{\emph{CoRR}}  \bibinfo{volume}{abs/2102.09844} (\bibinfo{year}{2021}).
\newblock
\showeprint[arXiv]{2102.09844}
\urldef\tempurl%
\url{https://arxiv.org/abs/2102.09844}
\showURL{%
\tempurl}


\bibitem[Schwarze et~al\mbox{.}(2016)]%
        {Schwarze2016BandSE}
\bibfield{author}{\bibinfo{person}{Martin Schwarze}, \bibinfo{person}{Wolfgang~R. Tress}, \bibinfo{person}{Beatrice Beyer}, \bibinfo{person}{Feng Gao}, \bibinfo{person}{Reinhard Scholz}, \bibinfo{person}{Carl Poelking}, \bibinfo{person}{Katrin Ortstein}, \bibinfo{person}{Alrun~A. G{\"u}nther}, \bibinfo{person}{Daniel Kasemann}, \bibinfo{person}{Denis Andrienko}, {and} \bibinfo{person}{Karl Leo}.} \bibinfo{year}{2016}\natexlab{}.
\newblock \showarticletitle{Band structure engineering in organic semiconductors}.
\newblock \bibinfo{journal}{\emph{Science}}  \bibinfo{volume}{352} (\bibinfo{year}{2016}), \bibinfo{pages}{1446 -- 1449}.
\newblock
\urldef\tempurl%
\url{https://api.semanticscholar.org/CorpusID:27736779}
\showURL{%
\tempurl}


\bibitem[Shi et~al\mbox{.}(2022)]%
        {Shi2022Graphormer}
\bibfield{author}{\bibinfo{person}{Yu Shi}, \bibinfo{person}{Shuxin Zheng}, \bibinfo{person}{Guolin Ke}, \bibinfo{person}{Yifei Shen}, \bibinfo{person}{Jiacheng You}, \bibinfo{person}{Jiyan He}, \bibinfo{person}{Shengjie Luo}, \bibinfo{person}{Chang Liu}, \bibinfo{person}{Di He}, {and} \bibinfo{person}{Tie{-}Yan Liu}.} \bibinfo{year}{2022}\natexlab{}.
\newblock \showarticletitle{Benchmarking Graphormer on Large-Scale Molecular Modeling Datasets}.
\newblock \bibinfo{journal}{\emph{CoRR}}  \bibinfo{volume}{abs/2203.04810} (\bibinfo{year}{2022}).
\newblock
\href{https://doi.org/10.48550/ARXIV.2203.04810}{doi:\nolinkurl{10.48550/ARXIV.2203.04810}}
\showeprint[arXiv]{2203.04810}


\bibitem[Simine et~al\mbox{.}(2020)]%
        {Simine2020PredictingOS}
\bibfield{author}{\bibinfo{person}{Lena Simine}, \bibinfo{person}{Thomas~C. Allen}, {and} \bibinfo{person}{Peter~J. Rossky}.} \bibinfo{year}{2020}\natexlab{}.
\newblock \showarticletitle{Predicting optical spectra for optoelectronic polymers using coarse-grained models and recurrent neural networks}.
\newblock \bibinfo{journal}{\emph{Proceedings of the National Academy of Sciences}}  \bibinfo{volume}{117} (\bibinfo{year}{2020}), \bibinfo{pages}{13945 -- 13948}.
\newblock
\urldef\tempurl%
\url{https://api.semanticscholar.org/CorpusID:219553008}
\showURL{%
\tempurl}


\bibitem[St{\"{a}}rk et~al\mbox{.}(2022)]%
        {Hannes2022Infomax}
\bibfield{author}{\bibinfo{person}{Hannes St{\"{a}}rk}, \bibinfo{person}{Dominique Beaini}, \bibinfo{person}{Gabriele Corso}, \bibinfo{person}{Prudencio Tossou}, \bibinfo{person}{Christian Dallago}, \bibinfo{person}{Stephan G{\"{u}}nnemann}, {and} \bibinfo{person}{Pietro Li{\'{o}}}.} \bibinfo{year}{2022}\natexlab{}.
\newblock \showarticletitle{3D Infomax improves GNNs for Molecular Property Prediction}. In \bibinfo{booktitle}{\emph{International Conference on Machine Learning, {ICML} 2022, 17-23 July 2022, Baltimore, Maryland, {USA}}} \emph{(\bibinfo{series}{Proceedings of Machine Learning Research}, Vol.~\bibinfo{volume}{162})}, \bibfield{editor}{\bibinfo{person}{Kamalika Chaudhuri}, \bibinfo{person}{Stefanie Jegelka}, \bibinfo{person}{Le~Song}, \bibinfo{person}{Csaba Szepesv{\'{a}}ri}, \bibinfo{person}{Gang Niu}, {and} \bibinfo{person}{Sivan Sabato}} (Eds.). \bibinfo{publisher}{{PMLR}}, \bibinfo{pages}{20479--20502}.
\newblock
\urldef\tempurl%
\url{https://proceedings.mlr.press/v162/stark22a.html}
\showURL{%
\tempurl}


\bibitem[Szegedy et~al\mbox{.}(2014)]%
        {Szegedy2014Intriguing}
\bibfield{author}{\bibinfo{person}{Christian Szegedy}, \bibinfo{person}{Wojciech Zaremba}, \bibinfo{person}{Ilya Sutskever}, \bibinfo{person}{Joan Bruna}, \bibinfo{person}{Dumitru Erhan}, \bibinfo{person}{Ian~J. Goodfellow}, {and} \bibinfo{person}{Rob Fergus}.} \bibinfo{year}{2014}\natexlab{}.
\newblock \showarticletitle{Intriguing properties of neural networks}. In \bibinfo{booktitle}{\emph{2nd International Conference on Learning Representations, {ICLR} 2014, Banff, AB, Canada, April 14-16, 2014, Conference Track Proceedings}}, \bibfield{editor}{\bibinfo{person}{Yoshua Bengio} {and} \bibinfo{person}{Yann LeCun}} (Eds.).
\newblock
\urldef\tempurl%
\url{http://arxiv.org/abs/1312.6199}
\showURL{%
\tempurl}


\bibitem[Vaswani et~al\mbox{.}(2017)]%
        {Vaswani2017Transformer}
\bibfield{author}{\bibinfo{person}{Ashish Vaswani}, \bibinfo{person}{Noam Shazeer}, \bibinfo{person}{Niki Parmar}, \bibinfo{person}{Jakob Uszkoreit}, \bibinfo{person}{Llion Jones}, \bibinfo{person}{Aidan~N. Gomez}, \bibinfo{person}{Lukasz Kaiser}, {and} \bibinfo{person}{Illia Polosukhin}.} \bibinfo{year}{2017}\natexlab{}.
\newblock \showarticletitle{Attention is All you Need}. In \bibinfo{booktitle}{\emph{Advances in Neural Information Processing Systems 30: Annual Conference on Neural Information Processing Systems 2017, December 4-9, 2017, Long Beach, CA, {USA}}}, \bibfield{editor}{\bibinfo{person}{Isabelle Guyon}, \bibinfo{person}{Ulrike von Luxburg}, \bibinfo{person}{Samy Bengio}, \bibinfo{person}{Hanna~M. Wallach}, \bibinfo{person}{Rob Fergus}, \bibinfo{person}{S.~V.~N. Vishwanathan}, {and} \bibinfo{person}{Roman Garnett}} (Eds.). \bibinfo{pages}{5998--6008}.
\newblock
\urldef\tempurl%
\url{https://proceedings.neurips.cc/paper/2017/hash/3f5ee243547dee91fbd053c1c4a845aa-Abstract.html}
\showURL{%
\tempurl}


\bibitem[Velickovic et~al\mbox{.}(2018)]%
        {Velickovic2018GAT}
\bibfield{author}{\bibinfo{person}{Petar Velickovic}, \bibinfo{person}{Guillem Cucurull}, \bibinfo{person}{Arantxa Casanova}, \bibinfo{person}{Adriana Romero}, \bibinfo{person}{Pietro Li{\`{o}}}, {and} \bibinfo{person}{Yoshua Bengio}.} \bibinfo{year}{2018}\natexlab{}.
\newblock \showarticletitle{Graph Attention Networks}. In \bibinfo{booktitle}{\emph{6th International Conference on Learning Representations, {ICLR} 2018, Vancouver, BC, Canada, April 30 - May 3, 2018, Conference Track Proceedings}}. \bibinfo{publisher}{OpenReview.net}.
\newblock
\urldef\tempurl%
\url{https://openreview.net/forum?id=rJXMpikCZ}
\showURL{%
\tempurl}


\bibitem[Wang et~al\mbox{.}(2024)]%
        {Wang2024MMPolymer}
\bibfield{author}{\bibinfo{person}{Fanmeng Wang}, \bibinfo{person}{Wentao Guo}, \bibinfo{person}{Minjie Cheng}, \bibinfo{person}{Shen Yuan}, \bibinfo{person}{Hongteng Xu}, {and} \bibinfo{person}{Zhifeng Gao}.} \bibinfo{year}{2024}\natexlab{}.
\newblock \showarticletitle{MMPolymer: {A} Multimodal Multitask Pretraining Framework for Polymer Property Prediction}. In \bibinfo{booktitle}{\emph{Proceedings of the 33rd {ACM} International Conference on Information and Knowledge Management, {CIKM} 2024, Boise, ID, USA, October 21-25, 2024}}, \bibfield{editor}{\bibinfo{person}{Edoardo Serra} {and} \bibinfo{person}{Francesca Spezzano}} (Eds.). \bibinfo{publisher}{{ACM}}, \bibinfo{pages}{2336--2346}.
\newblock
\href{https://doi.org/10.1145/3627673.3679684}{doi:\nolinkurl{10.1145/3627673.3679684}}


\bibitem[Wang et~al\mbox{.}(2025)]%
        {Wang2025Fragformer}
\bibfield{author}{\bibinfo{person}{Jiaxi Wang}, \bibinfo{person}{Yaosen Min}, \bibinfo{person}{Miao Li}, {and} \bibinfo{person}{Ji Wu}.} \bibinfo{year}{2025}\natexlab{}.
\newblock \showarticletitle{FragFormer: A Fragment-based Representation Learning Framework for Molecular Property Prediction}.
\newblock \bibinfo{journal}{\emph{Transactions on Machine Learning Research}} (\bibinfo{year}{2025}).
\newblock
\showISSN{2835-8856}
\urldef\tempurl%
\url{https://openreview.net/forum?id=9aiuB3kIjd}
\showURL{%
\tempurl}


\bibitem[Wang et~al\mbox{.}(2023)]%
        {Wang2023A3PT}
\bibfield{author}{\bibinfo{person}{Xu Wang}, \bibinfo{person}{Huan Zhao}, \bibinfo{person}{Wei{-}Wei Tu}, {and} \bibinfo{person}{Quanming Yao}.} \bibinfo{year}{2023}\natexlab{}.
\newblock \showarticletitle{Automated 3D Pre-Training for Molecular Property Prediction}. In \bibinfo{booktitle}{\emph{Proceedings of the 29th {ACM} {SIGKDD} Conference on Knowledge Discovery and Data Mining, {KDD} 2023, Long Beach, CA, USA, August 6-10, 2023}}, \bibfield{editor}{\bibinfo{person}{Ambuj~K. Singh}, \bibinfo{person}{Yizhou Sun}, \bibinfo{person}{Leman Akoglu}, \bibinfo{person}{Dimitrios Gunopulos}, \bibinfo{person}{Xifeng Yan}, \bibinfo{person}{Ravi Kumar}, \bibinfo{person}{Fatma Ozcan}, {and} \bibinfo{person}{Jieping Ye}} (Eds.). \bibinfo{publisher}{{ACM}}, \bibinfo{pages}{2419--2430}.
\newblock
\href{https://doi.org/10.1145/3580305.3599252}{doi:\nolinkurl{10.1145/3580305.3599252}}


\bibitem[Wang et~al\mbox{.}(2022)]%
        {Wang2022MolCLR}
\bibfield{author}{\bibinfo{person}{Yuyang Wang}, \bibinfo{person}{Jianren Wang}, \bibinfo{person}{Zhonglin Cao}, {and} \bibinfo{person}{Amir~Barati Farimani}.} \bibinfo{year}{2022}\natexlab{}.
\newblock \showarticletitle{Molecular contrastive learning of representations via graph neural networks}.
\newblock \bibinfo{journal}{\emph{Nat. Mach. Intell.}} \bibinfo{volume}{4}, \bibinfo{number}{3} (\bibinfo{year}{2022}), \bibinfo{pages}{279--287}.
\newblock
\href{https://doi.org/10.1038/S42256-022-00447-X}{doi:\nolinkurl{10.1038/S42256-022-00447-X}}


\bibitem[Watanabe et~al\mbox{.}(2022)]%
        {Watanabe2022TranscendingTT}
\bibfield{author}{\bibinfo{person}{Seigo Watanabe}, \bibinfo{person}{Terufumi Takayama}, {and} \bibinfo{person}{Kenichi Oyaizu}.} \bibinfo{year}{2022}\natexlab{}.
\newblock \showarticletitle{Transcending the Trade-off in Refractive Index and Abbe Number for Highly Refractive Polymers: Synergistic Effect of Polarizable Skeletons and Robust Hydrogen Bonds}.
\newblock \bibinfo{journal}{\emph{ACS Polymers Au}}  \bibinfo{volume}{2} (\bibinfo{year}{2022}), \bibinfo{pages}{458 -- 466}.
\newblock
\urldef\tempurl%
\url{https://api.semanticscholar.org/CorpusID:251707888}
\showURL{%
\tempurl}


\bibitem[Weininger(1988)]%
        {Weininger1988SMILES}
\bibfield{author}{\bibinfo{person}{David Weininger}.} \bibinfo{year}{1988}\natexlab{}.
\newblock \showarticletitle{SMILES, a chemical language and information system. 1. Introduction to methodology and encoding rules}.
\newblock \bibinfo{journal}{\emph{J. Chem. Inf. Comput. Sci.}}  \bibinfo{volume}{28} (\bibinfo{year}{1988}), \bibinfo{pages}{31--36}.
\newblock
\urldef\tempurl%
\url{https://api.semanticscholar.org/CorpusID:5445756}
\showURL{%
\tempurl}


\bibitem[Wollschl{\"{a}}ger et~al\mbox{.}(2024)]%
        {Tom2024FragNet}
\bibfield{author}{\bibinfo{person}{Tom Wollschl{\"{a}}ger}, \bibinfo{person}{Niklas Kemper}, \bibinfo{person}{Leon Hetzel}, \bibinfo{person}{Johanna Sommer}, {and} \bibinfo{person}{Stephan G{\"{u}}nnemann}.} \bibinfo{year}{2024}\natexlab{}.
\newblock \showarticletitle{Expressivity and Generalization: Fragment-Biases for Molecular GNNs}. In \bibinfo{booktitle}{\emph{Forty-first International Conference on Machine Learning, {ICML} 2024, Vienna, Austria, July 21-27, 2024}}. \bibinfo{publisher}{OpenReview.net}.
\newblock
\urldef\tempurl%
\url{https://openreview.net/forum?id=rPm5cKb1VB}
\showURL{%
\tempurl}


\bibitem[Xu et~al\mbox{.}(2022)]%
        {Xu2022TransPolymer}
\bibfield{author}{\bibinfo{person}{Changwen Xu}, \bibinfo{person}{Yuyang Wang}, {and} \bibinfo{person}{Amir~Barati Farimani}.} \bibinfo{year}{2022}\natexlab{}.
\newblock \showarticletitle{TransPolymer: a Transformer-based language model for polymer property predictions}.
\newblock \bibinfo{journal}{\emph{npj Computational Materials}}  \bibinfo{volume}{9} (\bibinfo{year}{2022}), \bibinfo{pages}{1--14}.
\newblock
\urldef\tempurl%
\url{https://api.semanticscholar.org/CorpusID:252090395}
\showURL{%
\tempurl}


\bibitem[Xu et~al\mbox{.}(2019)]%
        {Xu2019GIN}
\bibfield{author}{\bibinfo{person}{Keyulu Xu}, \bibinfo{person}{Weihua Hu}, \bibinfo{person}{Jure Leskovec}, {and} \bibinfo{person}{Stefanie Jegelka}.} \bibinfo{year}{2019}\natexlab{}.
\newblock \showarticletitle{How Powerful are Graph Neural Networks?}. In \bibinfo{booktitle}{\emph{7th International Conference on Learning Representations, {ICLR} 2019, New Orleans, LA, USA, May 6-9, 2019}}. \bibinfo{publisher}{OpenReview.net}.
\newblock
\urldef\tempurl%
\url{https://openreview.net/forum?id=ryGs6iA5Km}
\showURL{%
\tempurl}


\bibitem[Yang et~al\mbox{.}(2019)]%
        {Yang2019AnalyzingLM}
\bibfield{author}{\bibinfo{person}{Kevin Yang}, \bibinfo{person}{Kyle Swanson}, \bibinfo{person}{Wengong Jin}, \bibinfo{person}{Connor~W. Coley}, \bibinfo{person}{Philipp Eiden}, \bibinfo{person}{Hua Gao}, \bibinfo{person}{Angel Guzman{-}Perez}, \bibinfo{person}{Timothy Hopper}, \bibinfo{person}{Brian Kelley}, \bibinfo{person}{Miriam Mathea}, \bibinfo{person}{Andrew Palmer}, \bibinfo{person}{Volker Settels}, \bibinfo{person}{Tommi~S. Jaakkola}, \bibinfo{person}{Klavs~F. Jensen}, {and} \bibinfo{person}{Regina Barzilay}.} \bibinfo{year}{2019}\natexlab{}.
\newblock \showarticletitle{Analyzing Learned Molecular Representations for Property Prediction}.
\newblock \bibinfo{journal}{\emph{J. Chem. Inf. Model.}} \bibinfo{volume}{59}, \bibinfo{number}{8} (\bibinfo{year}{2019}), \bibinfo{pages}{3370--3388}.
\newblock
\href{https://doi.org/10.1021/ACS.JCIM.9B00237}{doi:\nolinkurl{10.1021/ACS.JCIM.9B00237}}


\bibitem[Ying et~al\mbox{.}(2021)]%
        {Ying2021Graphormer}
\bibfield{author}{\bibinfo{person}{Chengxuan Ying}, \bibinfo{person}{Tianle Cai}, \bibinfo{person}{Shengjie Luo}, \bibinfo{person}{Shuxin Zheng}, \bibinfo{person}{Guolin Ke}, \bibinfo{person}{Di He}, \bibinfo{person}{Yanming Shen}, {and} \bibinfo{person}{Tie{-}Yan Liu}.} \bibinfo{year}{2021}\natexlab{}.
\newblock \showarticletitle{Do Transformers Really Perform Badly for Graph Representation?}. In \bibinfo{booktitle}{\emph{Advances in Neural Information Processing Systems 34: Annual Conference on Neural Information Processing Systems 2021, NeurIPS 2021, December 6-14, 2021, virtual}}, \bibfield{editor}{\bibinfo{person}{Marc'Aurelio Ranzato}, \bibinfo{person}{Alina Beygelzimer}, \bibinfo{person}{Yann~N. Dauphin}, \bibinfo{person}{Percy Liang}, {and} \bibinfo{person}{Jennifer~Wortman Vaughan}} (Eds.). \bibinfo{pages}{28877--28888}.
\newblock
\urldef\tempurl%
\url{https://proceedings.neurips.cc/paper/2021/hash/f1c1592588411002af340cbaedd6fc33-Abstract.html}
\showURL{%
\tempurl}


\bibitem[Zaidi et~al\mbox{.}(2023)]%
        {Zaidi2023Denoising}
\bibfield{author}{\bibinfo{person}{Sheheryar Zaidi}, \bibinfo{person}{Michael Schaarschmidt}, \bibinfo{person}{James Martens}, \bibinfo{person}{Hyunjik Kim}, \bibinfo{person}{Yee~Whye Teh}, \bibinfo{person}{Alvaro Sanchez{-}Gonzalez}, \bibinfo{person}{Peter~W. Battaglia}, \bibinfo{person}{Razvan Pascanu}, {and} \bibinfo{person}{Jonathan Godwin}.} \bibinfo{year}{2023}\natexlab{}.
\newblock \showarticletitle{Pre-training via Denoising for Molecular Property Prediction}. In \bibinfo{booktitle}{\emph{The Eleventh International Conference on Learning Representations, {ICLR} 2023, Kigali, Rwanda, May 1-5, 2023}}. \bibinfo{publisher}{OpenReview.net}.
\newblock
\urldef\tempurl%
\url{https://openreview.net/forum?id=tYIMtogyee}
\showURL{%
\tempurl}


\bibitem[Zhang et~al\mbox{.}(2025)]%
        {Zhang2025Expressive}
\bibfield{author}{\bibinfo{person}{Bingxu Zhang}, \bibinfo{person}{Changjun Fan}, \bibinfo{person}{Shixuan Liu}, \bibinfo{person}{Kuihua Huang}, \bibinfo{person}{Xiang Zhao}, \bibinfo{person}{Jincai Huang}, {and} \bibinfo{person}{Zhong Liu}.} \bibinfo{year}{2025}\natexlab{}.
\newblock \showarticletitle{The Expressive Power of Graph Neural Networks: {A} Survey}.
\newblock \bibinfo{journal}{\emph{{IEEE} Trans. Knowl. Data Eng.}} \bibinfo{volume}{37}, \bibinfo{number}{3} (\bibinfo{year}{2025}), \bibinfo{pages}{1455--1474}.
\newblock
\href{https://doi.org/10.1109/TKDE.2024.3523700}{doi:\nolinkurl{10.1109/TKDE.2024.3523700}}


\bibitem[Zhang et~al\mbox{.}(2023)]%
        {Pei2023PPP}
\bibfield{author}{\bibinfo{person}{Pei Zhang}, \bibinfo{person}{Logan~T. Kearney}, \bibinfo{person}{Debsindhu Bhowmik}, \bibinfo{person}{Zachary~R. Fox}, \bibinfo{person}{Amit~K. Naskar}, {and} \bibinfo{person}{John Gounley}.} \bibinfo{year}{2023}\natexlab{}.
\newblock \showarticletitle{Transferring a Molecular Foundation Model for Polymer Property Predictions}.
\newblock \bibinfo{journal}{\emph{J. Chem. Inf. Model.}} \bibinfo{volume}{63}, \bibinfo{number}{24} (\bibinfo{year}{2023}), \bibinfo{pages}{7689--7698}.
\newblock
\href{https://doi.org/10.1021/ACS.JCIM.3C01650}{doi:\nolinkurl{10.1021/ACS.JCIM.3C01650}}


\bibitem[Zhou et~al\mbox{.}(2023)]%
        {Geng2023UniMol}
\bibfield{author}{\bibinfo{person}{Gengmo Zhou}, \bibinfo{person}{Zhifeng Gao}, \bibinfo{person}{Qiankun Ding}, \bibinfo{person}{Hang Zheng}, \bibinfo{person}{Hongteng Xu}, \bibinfo{person}{Zhewei Wei}, \bibinfo{person}{Linfeng Zhang}, {and} \bibinfo{person}{Guolin Ke}.} \bibinfo{year}{2023}\natexlab{}.
\newblock \showarticletitle{Uni-Mol: {A} Universal 3D Molecular Representation Learning Framework}. In \bibinfo{booktitle}{\emph{The Eleventh International Conference on Learning Representations, {ICLR} 2023, Kigali, Rwanda, May 1-5, 2023}}. \bibinfo{publisher}{OpenReview.net}.
\newblock
\urldef\tempurl%
\url{https://openreview.net/forum?id=6K2RM6wVqKu}
\showURL{%
\tempurl}


\bibitem[Zhu et~al\mbox{.}(2023)]%
        {Zhu2023GNNIR}
\bibfield{author}{\bibinfo{person}{Jinhua Zhu}, \bibinfo{person}{Kehan Wu}, \bibinfo{person}{Bohan Wang}, \bibinfo{person}{Yingce Xia}, \bibinfo{person}{Shufang Xie}, \bibinfo{person}{Qi Meng}, \bibinfo{person}{Lijun Wu}, \bibinfo{person}{Tao Qin}, \bibinfo{person}{Wengang Zhou}, \bibinfo{person}{Houqiang Li}, {and} \bibinfo{person}{Tie{-}Yan Liu}.} \bibinfo{year}{2023}\natexlab{}.
\newblock \showarticletitle{O-GNN: incorporating ring priors into molecular modeling}. In \bibinfo{booktitle}{\emph{The Eleventh International Conference on Learning Representations, {ICLR} 2023, Kigali, Rwanda, May 1-5, 2023}}. \bibinfo{publisher}{OpenReview.net}.
\newblock
\urldef\tempurl%
\url{https://openreview.net/forum?id=5cFfz6yMVPU}
\showURL{%
\tempurl}


\end{thebibliography}

\appendix

\section{Proof of Proposition 1}

\begin{proposition}
  \label{proposition:appendix_mpm_eq_star_linking}
  Under the same message passing mechanism, 
  the node feature of the $i$-th atom in the polymer graph $G^p$ 
  is the same as the node feature of the $i$-th atom in the induced star linking graph $G^{*}$, $\forall 0 \leq i < \abs{V}$.
  \begin{proof}
    The message passing mechanism is stated as:
    \begin{equation}
      \mathbf{x}_v \gets \text{UPDATE}\left(\mathbf{x}_v, \text{AGG} \left( \{\mathbf{x}_u\}_{u \in \mathcal{N}(v)} \right) \right),
        \label{eq:appendix_mpm}
    \end{equation}

    For $0<i<\abs{V}-1$, $v_i$ is not the endpoints, so $v_i^p$ share the same neighborhood in $G^p$ with $v_i^{*}$ in $G^{*}$. 
    Therefore, the feature update of $v_i^p$ in $G^p$ is the same as that of $v_i^{*}$ in $G^{*}$. \newline 
    For $i=0$ or $i=\abs{V}-1$, $v_i$ is the boundary atom in $G^p$. 
    $v_0^p$~($v_{\abs{V}-1}^p$) is connected to $v_{-1}^p$~($v_{\abs{V}}^p$), while 
    $v_0^{*}$~($v_{\abs{V}-1}^{*}$) is connected to $v_{\abs{V}-1}^{*}$~($v_{0}^{*}$).
    By periodic symmetry, $v_{-1}^p$ and $v_{\abs{V}-1}^p$ share the same feature, as well as $v_{\abs{V}}^p$ and $v_{0}^p$. 
    Thus, the feature update of $v_0^p$ or $v_{\abs{V}-1}^p$ in $G^p$ is the same as that of $v_{0}^{*}$ and $v_{\abs{V}-1}^{*}$ in $G^{*}$. 
  \end{proof}
\end{proposition}
\section{Proof of Theorem 1}
\begin{theorem}
  \label{theorem:appendix_mpm_eq_star_linking}
  If the network is composed of message passing layers, nodewise transformations, and end with a mean pooling layer, 
  Then the output of the network on the polymer graph $G^p$ is the same as the output of the network on the induced star linking graph $G^{*}$. 
  \begin{proof}
    Due to the periodic symmetry of features in polymer graph $G^p$, it is sufficient to prove that
    the node features of $v_i^p$ is are identical to that of $v_i^{*}$, $\forall 0 \leq i < \abs{V}$, for each forward module. 
    This holds true for nodewise transformation, as they are independent of the graph structure. 
    Furthermore, by \cref{proposition:appendix_mpm_eq_star_linking}, it also holds for message passing layers. 
  \end{proof}
\end{theorem}

\section{Proof of Theorem 2}
\begin{theorem}
  \label{theorem:appendix_attn_eq_star_linking}
  Assume that the distance between boundary atoms in monomer graph $G$ is larger than $2d_{{thres}}-1$.
  If the network consists of localized graph attention layers, nodewise transformations, and concludes with a mean pooling layer, 
  then its output on the polymer graph $G^p$ will be identical to its output on the induced star-linking graph $G^{*}$. 
  \begin{proof}
    Similar to \cref{theorem:appendix_mpm_eq_star_linking}, 
    we only need to prove that the node features $v_i^p$ are identical to $v_i^{*}$, $\forall 0 \leq i < \abs{V}$, 
    for localized graph attention layers. \newline 
    Define the rceptive field of $v$, $R\left(v\right)$, as the set of nodes reachable from $v$ within $d_{{thres}}-1$ hops. 
    Since the distance between boundary atoms in monomer graph $G$ is larger than $2d_{{thres}}-1$, 
    $R\left(v_i^{p}\right)$ can not contain a pair of nodes $\left(v_{k_1}^p, v_{k_2}^p \right)$, 
    where $k_1-k_2= k \abs{V}$ for some $k \in \integers$. 
    By periodic symmetry, $R\left(v_i^{p}\right)$ is the same as $R\left(v_i^{*}\right)$. 
    Therefore, the localized attention update of $v_i^p$ in $G^p$ is the same as that of $v_i^{*}$ in $G^{*}$. 
  \end{proof}
\end{theorem}

\begin{figure*}[t]
  \captionsetup[subfigure]{justification=centering}
  \centering
  \begin{subfigure}[b]{0.24\textwidth}
    \centering
    \centerline{\includegraphics[width=\textwidth]{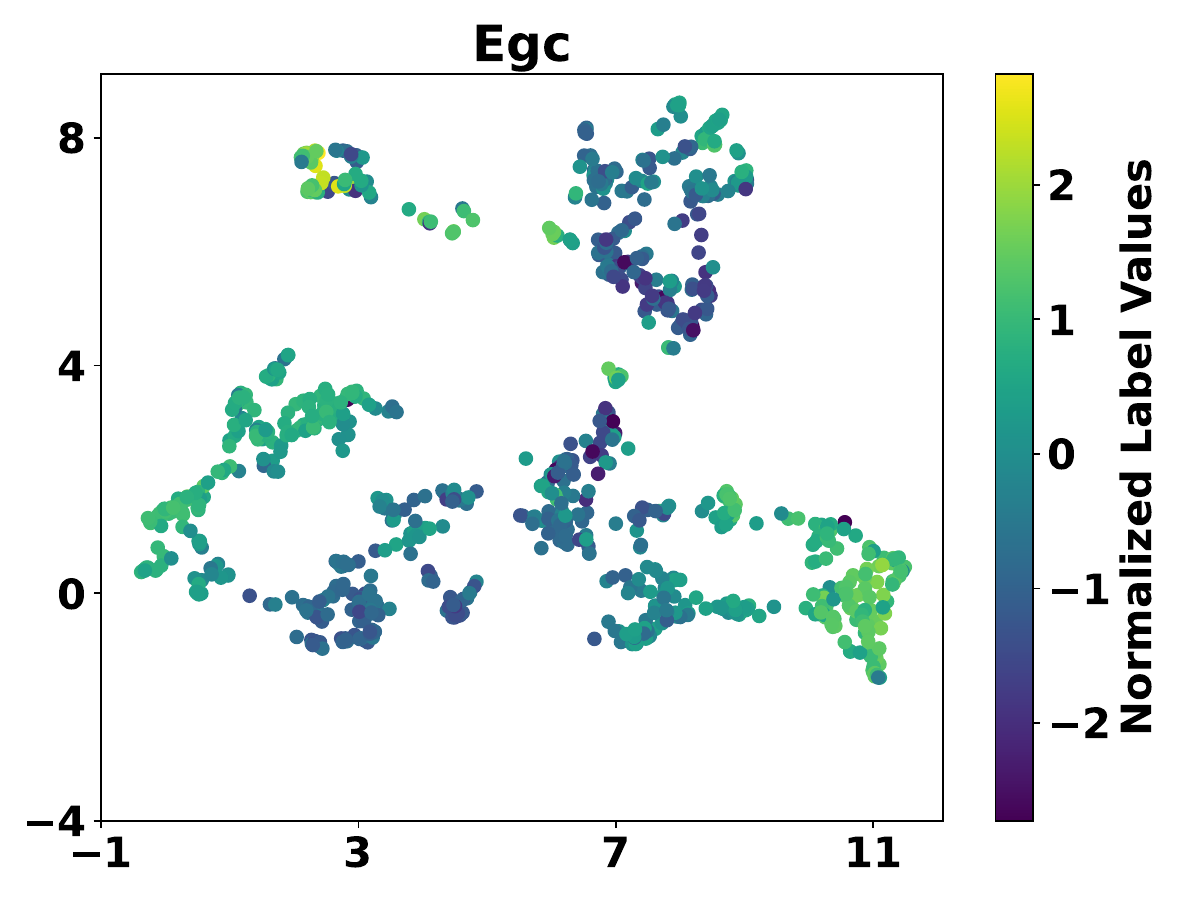}}
  \end{subfigure}
  \hfill
  \begin{subfigure}[b]{0.24\textwidth}
    \centering
    \includegraphics[width=\textwidth]{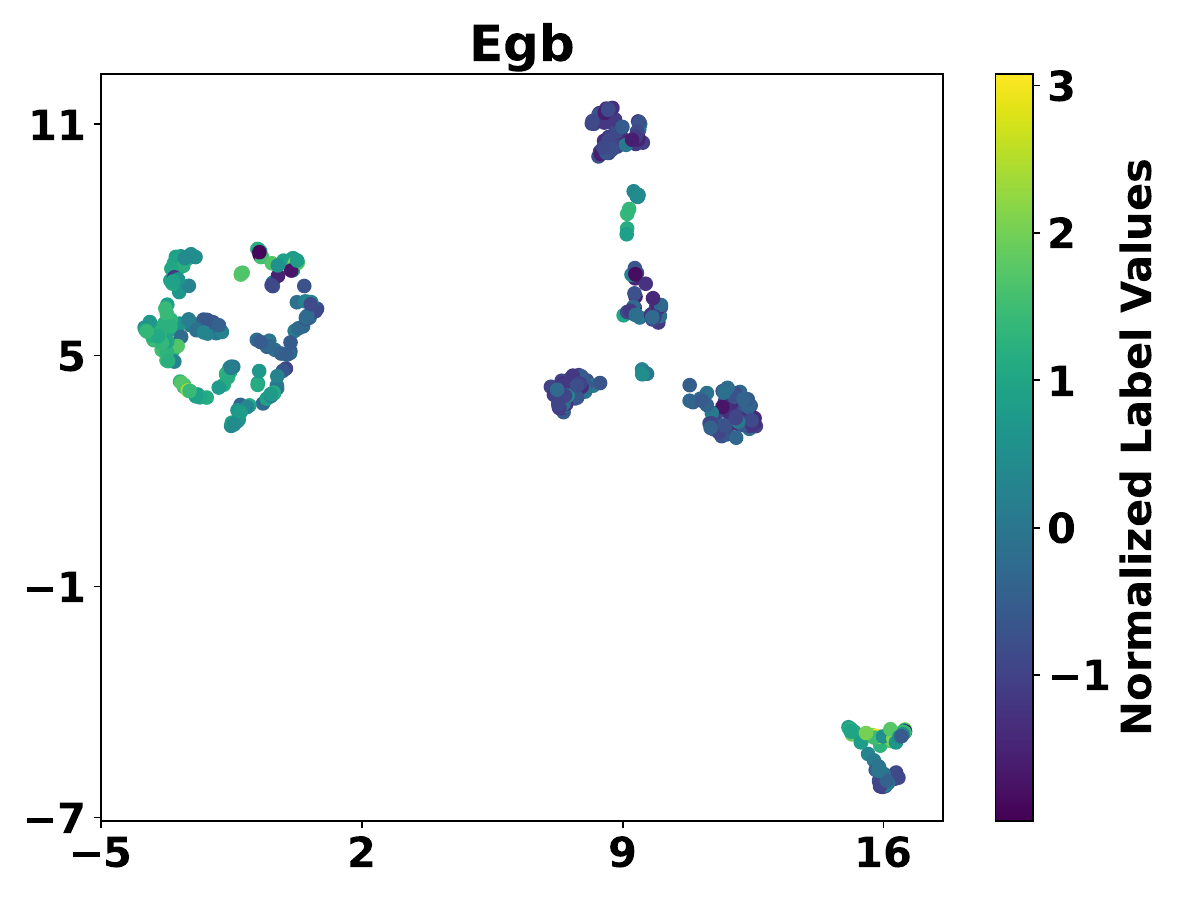}
  \end{subfigure}
  \hfill
  \begin{subfigure}[b]{0.24\textwidth}
    \centering
    \includegraphics[width=\textwidth]{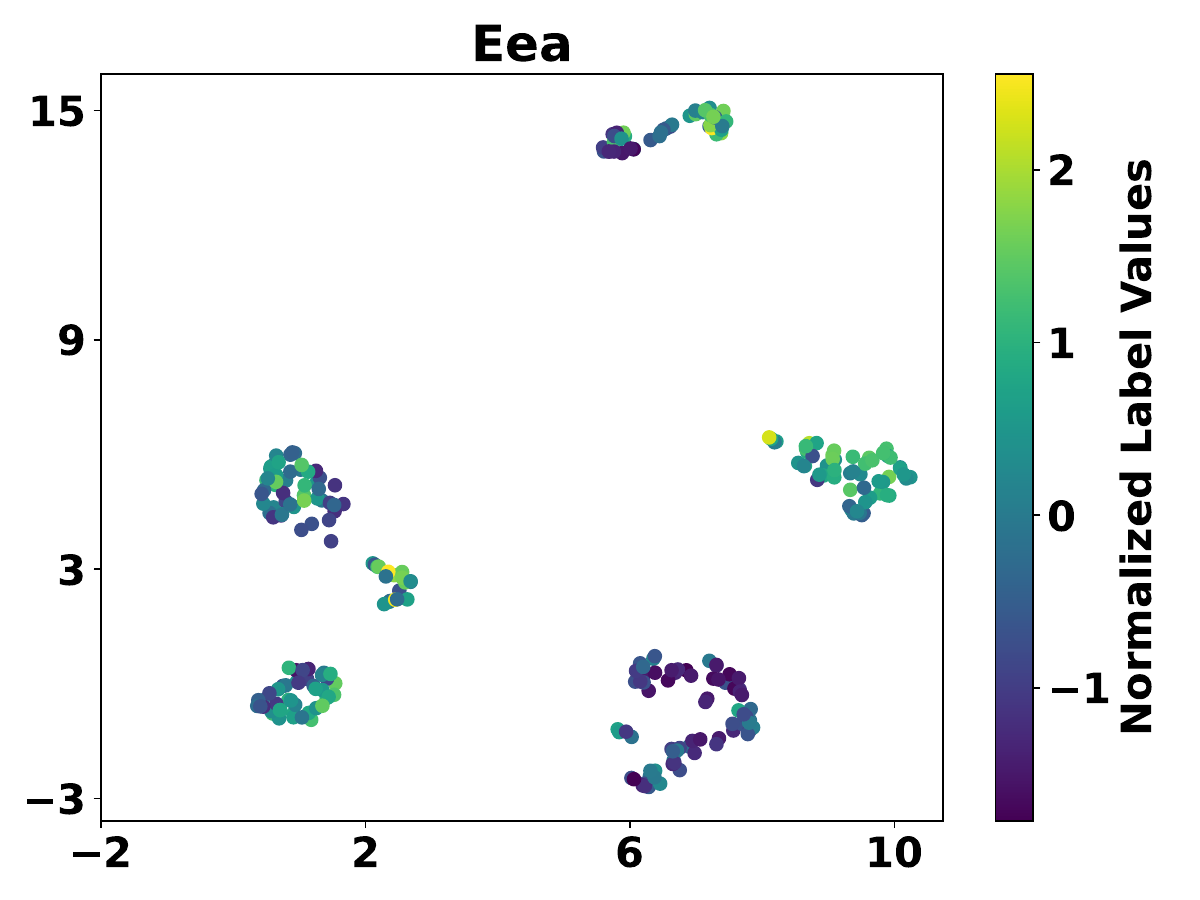}
  \end{subfigure}
  \hfill
  \begin{subfigure}[b]{0.24\textwidth}
    \centering
    \includegraphics[width=\textwidth]{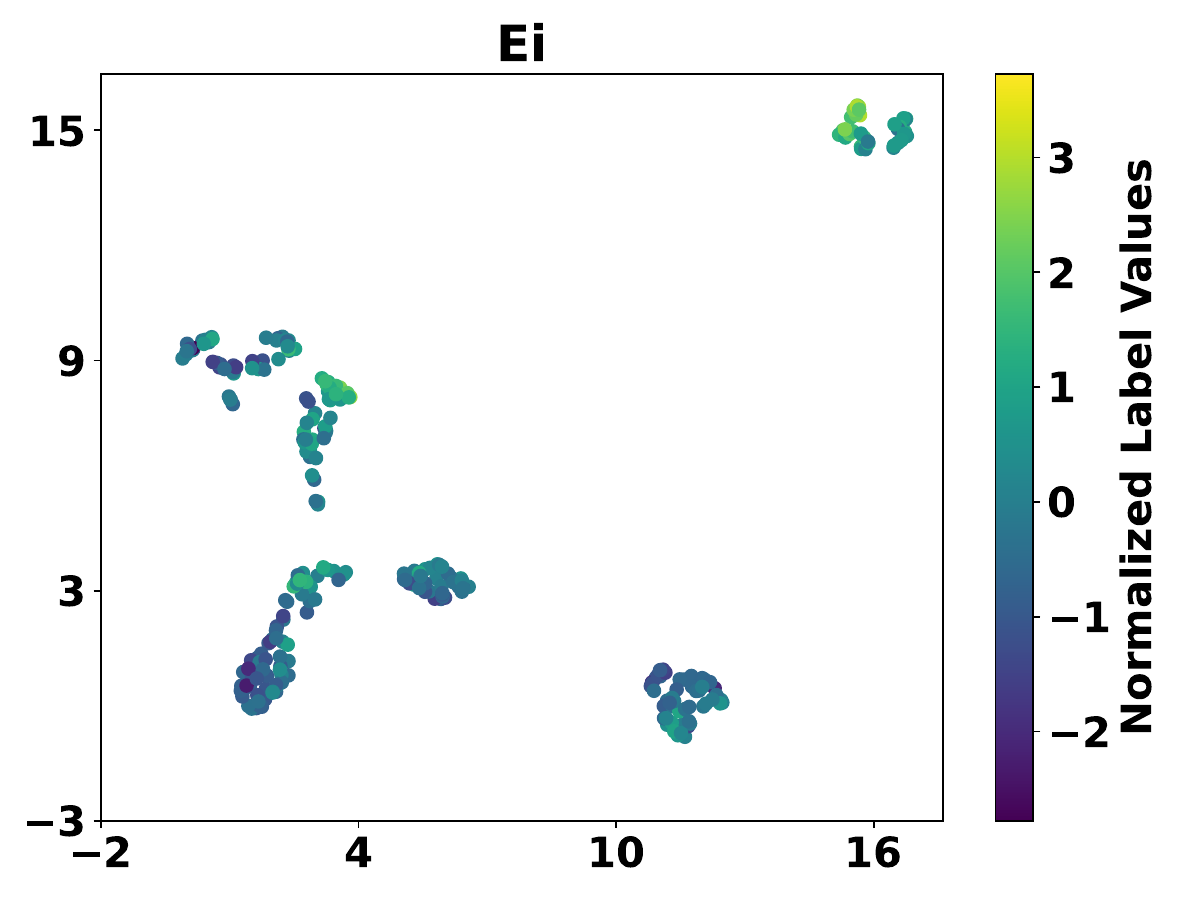}
  \end{subfigure}

  \begin{subfigure}[b]{0.24\textwidth}
    \centering
    \centerline{\includegraphics[width=\textwidth]{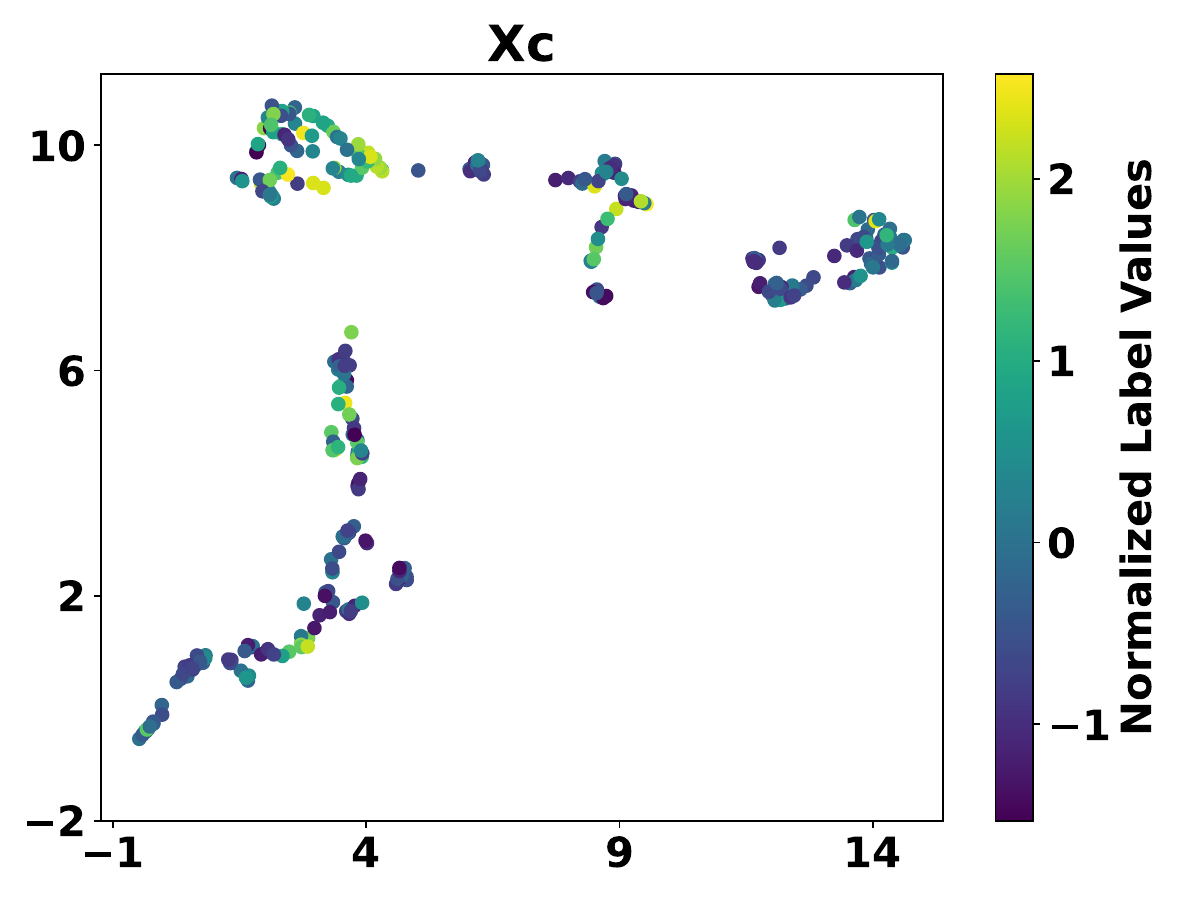}}
  \end{subfigure}
  \hfill
  \begin{subfigure}[b]{0.24\textwidth}
    \centering
    \includegraphics[width=\textwidth]{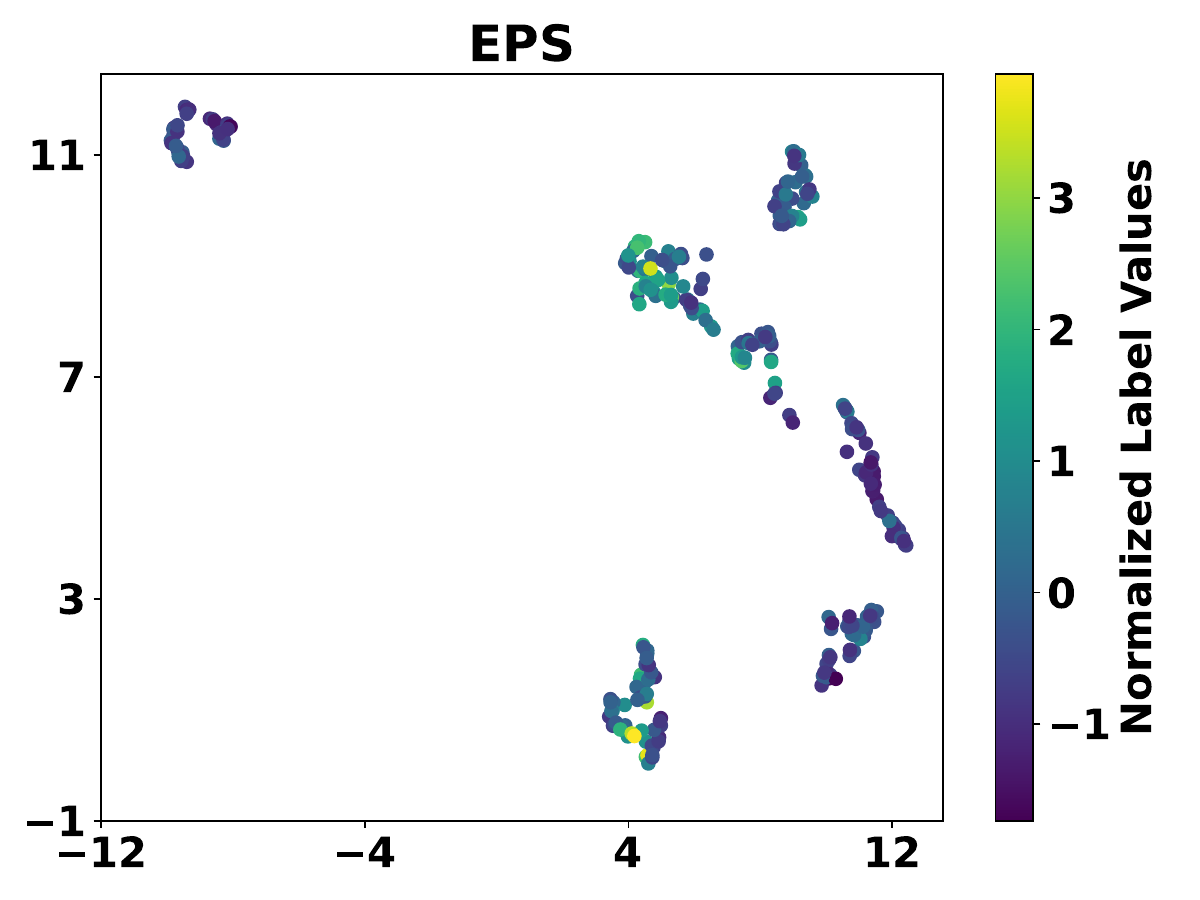}
  \end{subfigure}
  \hfill
  \begin{subfigure}[b]{0.24\textwidth}
    \centering
    \includegraphics[width=\textwidth]{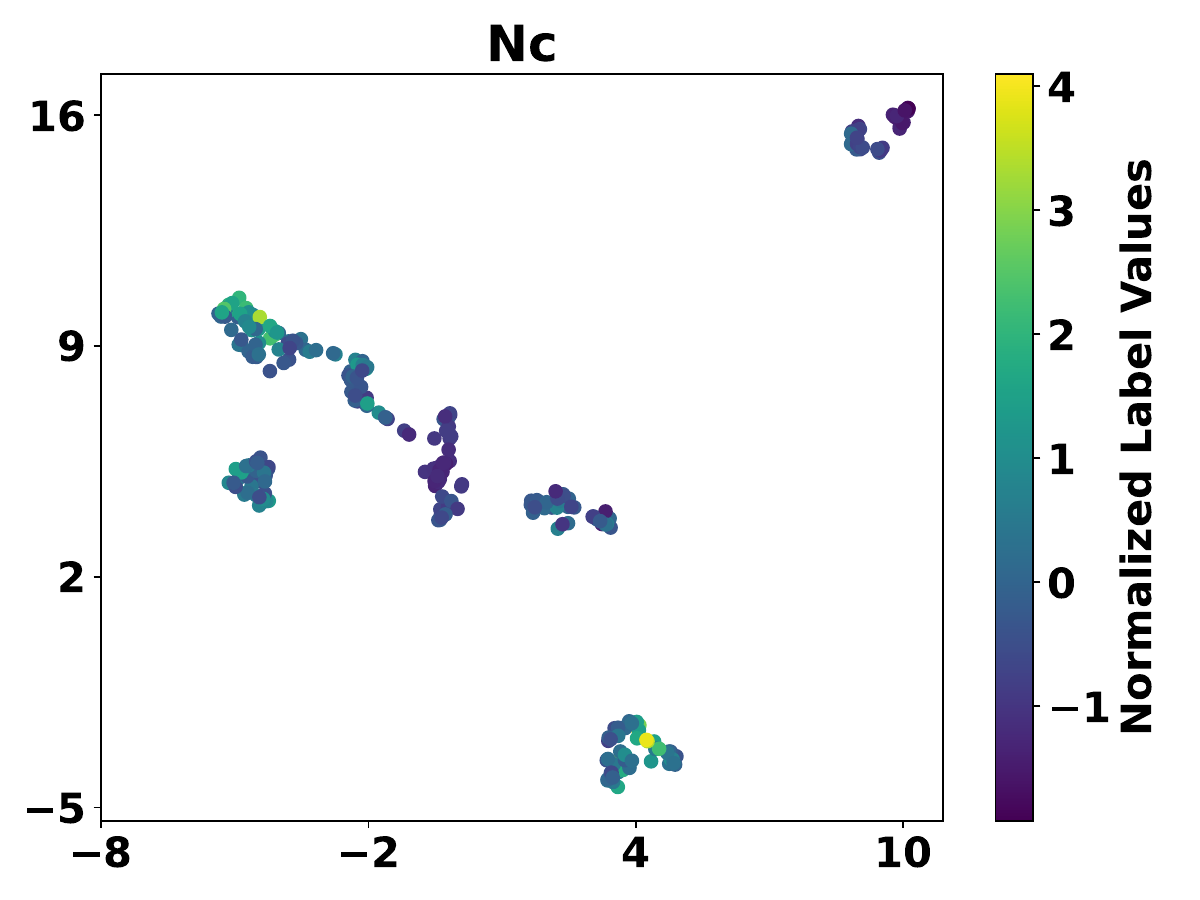}
  \end{subfigure}
  \hfill
  \begin{subfigure}[b]{0.24\textwidth}
    \centering
    \includegraphics[width=\textwidth]{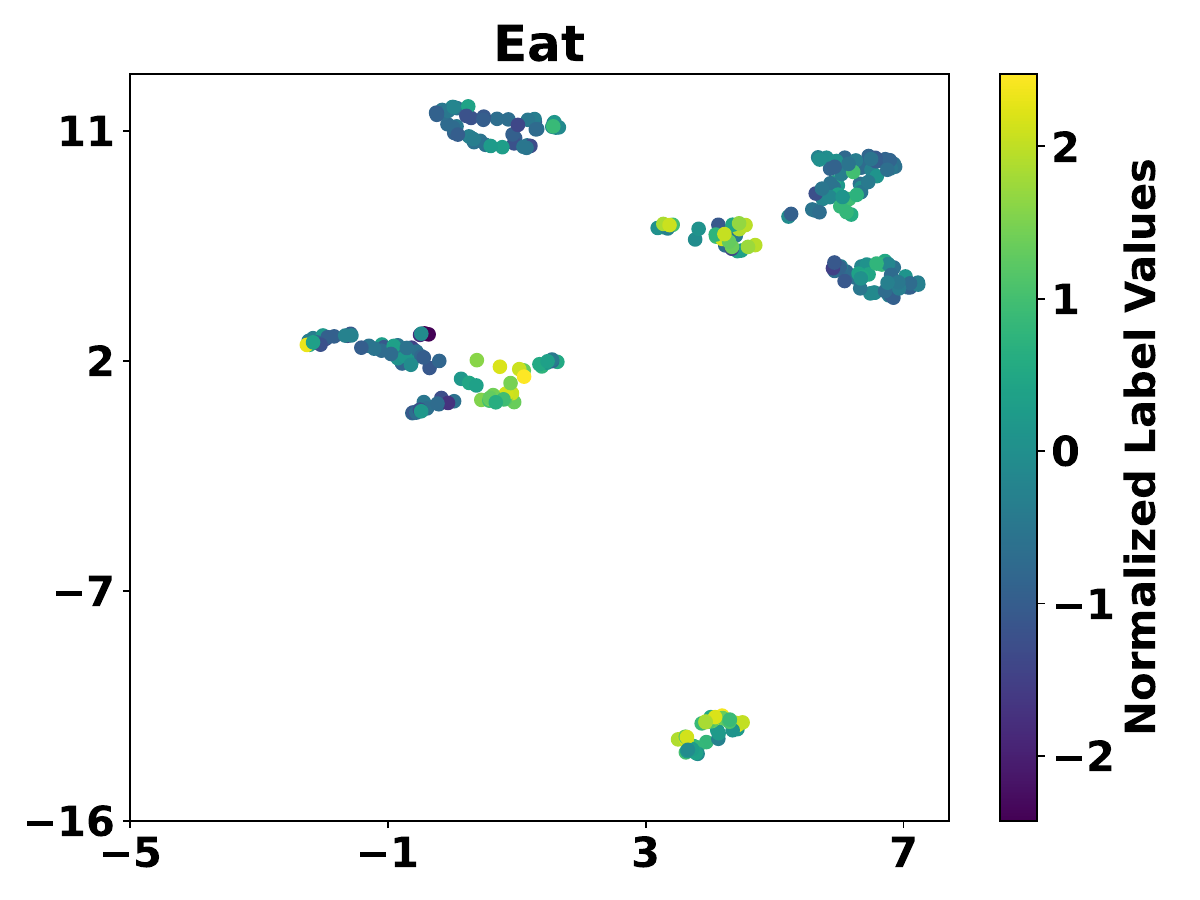}
  \end{subfigure}

  \caption{UMAP visualization of the embedding generated by MIPS on eight polymer property datasets.}
  \label{fig:visualize_embedding}
\end{figure*}

\begin{figure*}[t]
  \captionsetup[subfigure]{justification=centering}
  \centering
  \begin{subfigure}[b]{0.98\textwidth}
    \centering
    \centerline{\includegraphics[width=\textwidth]{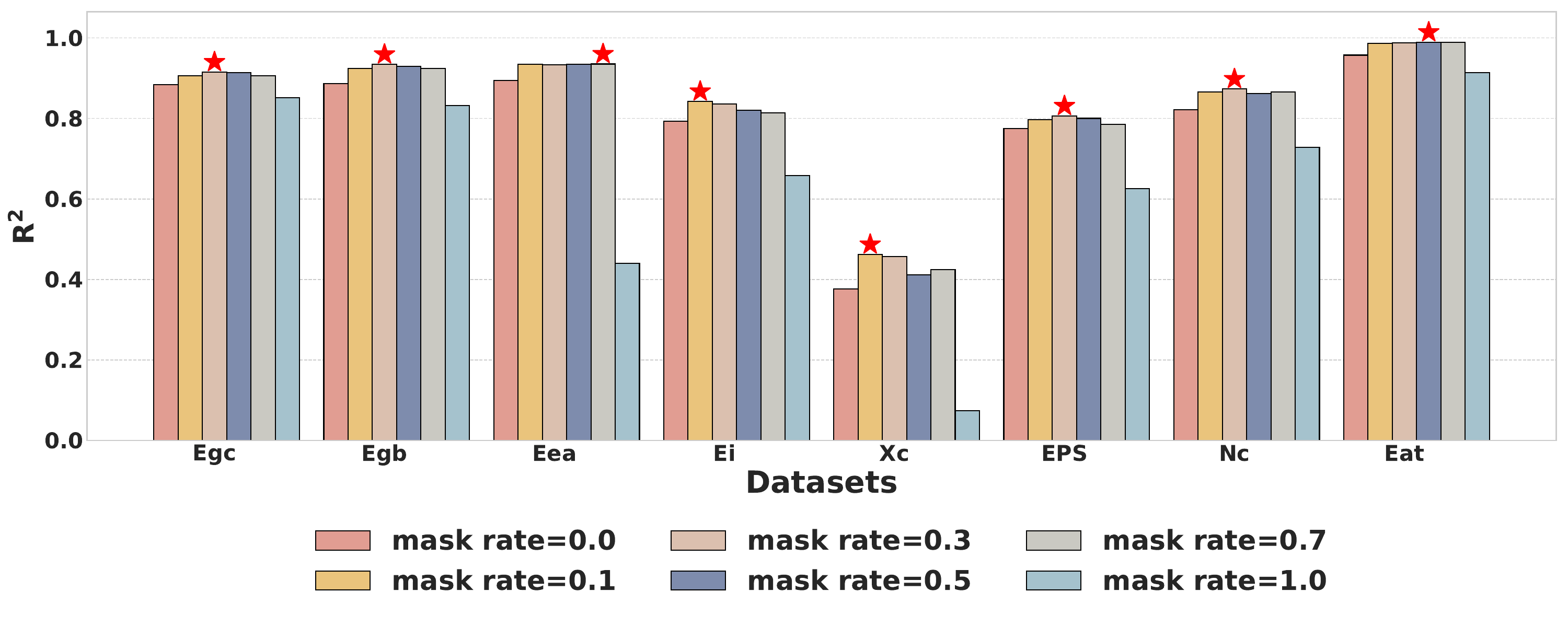}}
  \end{subfigure}
  \caption{Comparison of $R^2$ performance across different mask rates during pre-training. 
  The mask rate yieding the best result is highlighted with a red star for each dataset.}
  \label{fig:mask_rate}
\end{figure*}

\section{Proof of Lemma 1}
\begin{lemma}
  \label{lemma:appendix_polymer_wl_test}
  WL test can not distinguish twin polymer graphs $\left(G_1^p, G_2^p\right)$. 
  \begin{proof}
    WL test iteratively colors the nodes of the graph based on the colors of their neighbors. 
    By \cref{theorem:appendix_mpm_eq_star_linking}, the coloring results of $G^{p}$ on $v_i^{p}$, 
    $i=0, \ldots, \abs{V}-1$, is the same as that of $G^{*}$ on $v_i^{*}$. 
    Since $G_1^{*} = G_2^{*}$, the coloring results of $G_2^{*}$ is the same as $G_2^{p}$.
  \end{proof}
\end{lemma}

\section{Proof of Theorem 3}
\begin{theorem}
  \label{lemma:appendix_polymer_mpm_test}
  On twin polymer graphs $\left(G_1^p, G_2^p\right)$, massage passing mechanism~(MPM) or localized graph attention~(LGA) will produce identical outputs. 
  In other words, MPM and LGA are unable to distinguish twin polymer graphs. 
  \begin{proof}
    By \cref{lemma:appendix_polymer_wl_test} and the fact that the ability of MPM to distinguish non-isomorphic graphs is upper bounded by WL test~\cite{Xu2019GIN}, 
    MPM can not distinguish twin polymer graphs. \newline 
    For LGA, by \cref{theorem:appendix_attn_eq_star_linking} and the equivalence $G_1^{*} = G_2^{*}$, 
    LGA will produce identical outputs on $G_1^p$ and $G_2^p$. 
  \end{proof}
\end{theorem}

\begin{figure*}[h]
  \captionsetup[subfigure]{justification=centering}
  \centering
  \begin{subfigure}[b]{0.24\textwidth}
    \centering
    \centerline{\includegraphics[width=\textwidth]{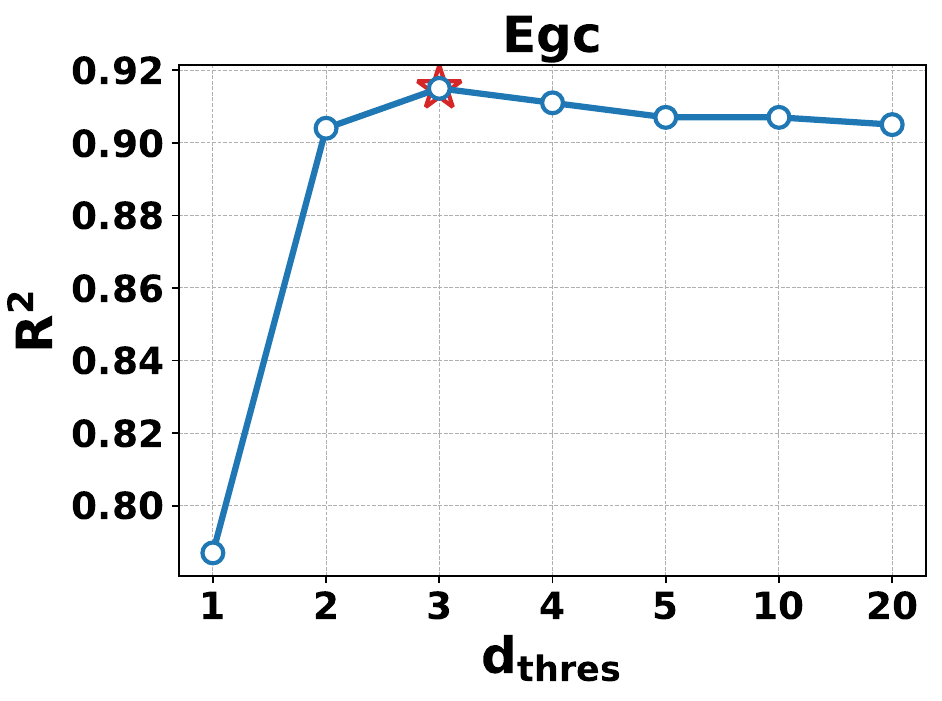}}
  \end{subfigure}
  \hfill
  \begin{subfigure}[b]{0.24\textwidth}
    \centering
    \includegraphics[width=\textwidth]{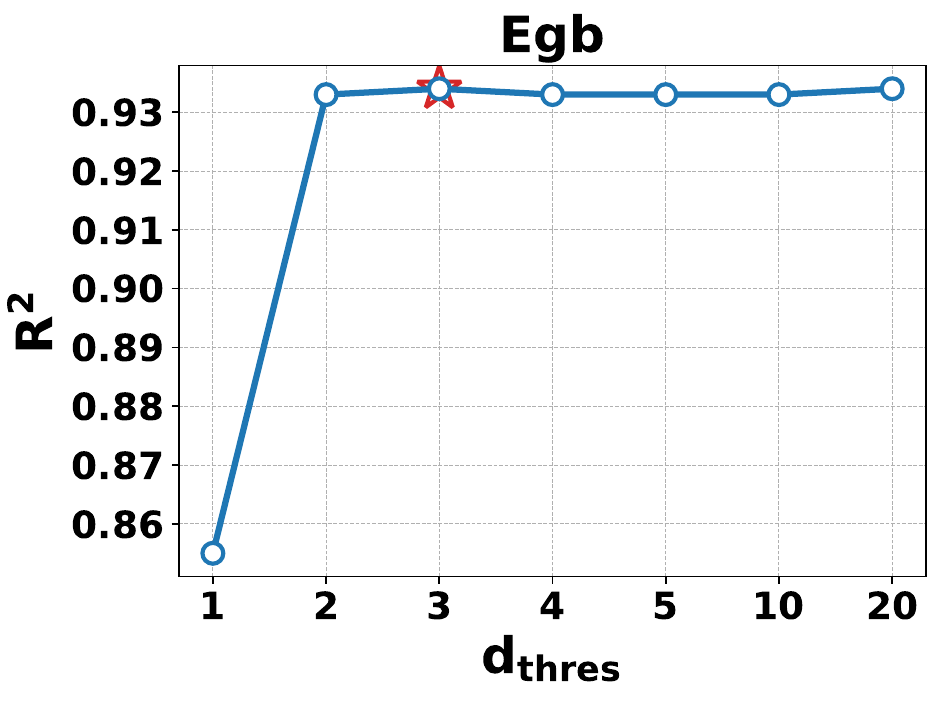}
  \end{subfigure}
  \hfill
  \begin{subfigure}[b]{0.24\textwidth}
    \centering
    \includegraphics[width=\textwidth]{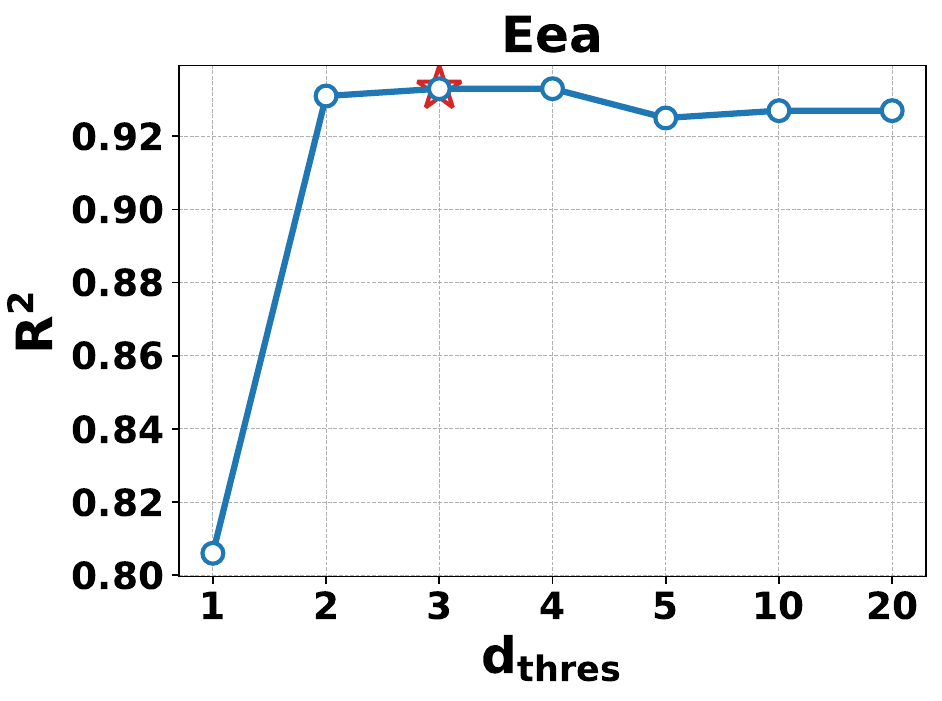}
  \end{subfigure}
  \hfill
  \begin{subfigure}[b]{0.24\textwidth}
    \centering
    \includegraphics[width=\textwidth]{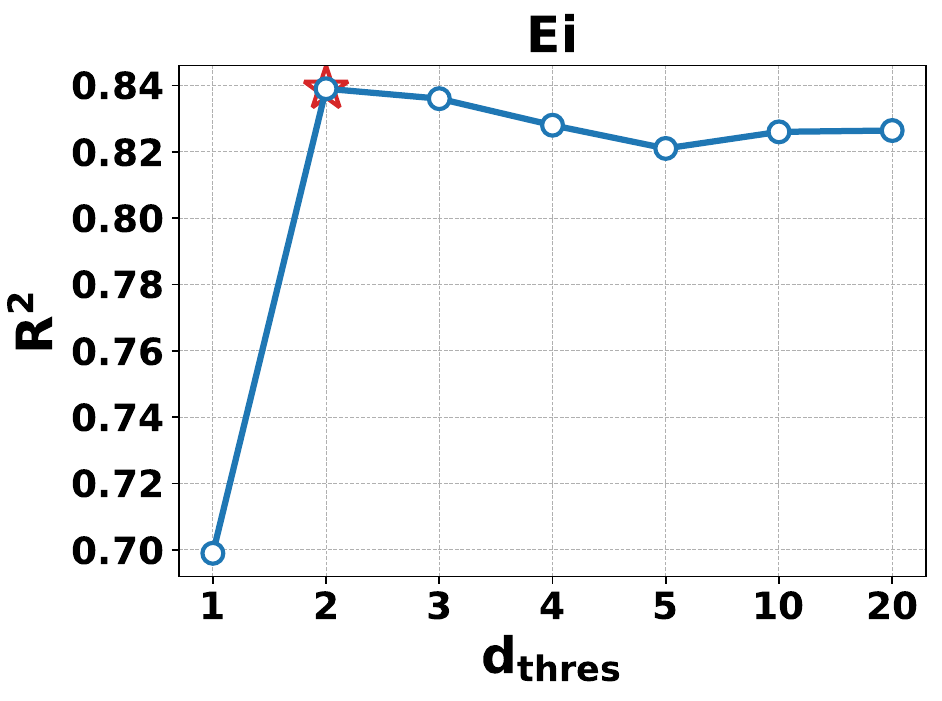}
  \end{subfigure}

  \begin{subfigure}[b]{0.24\textwidth}
    \centering
    \centerline{\includegraphics[width=\textwidth]{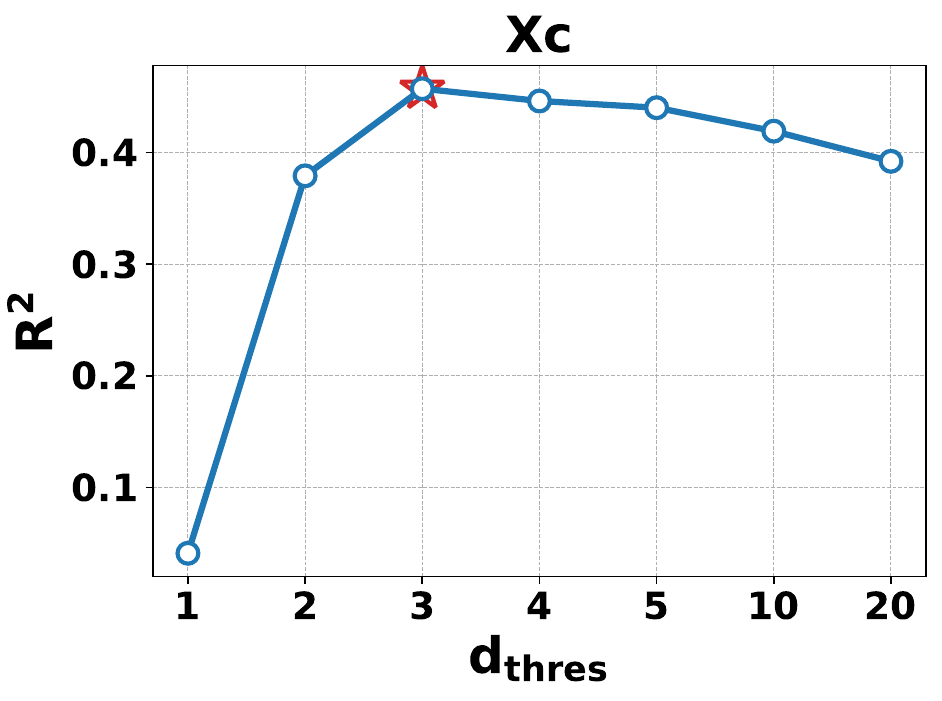}}
  \end{subfigure}
  \hfill
  \begin{subfigure}[b]{0.24\textwidth}
    \centering
    \includegraphics[width=\textwidth]{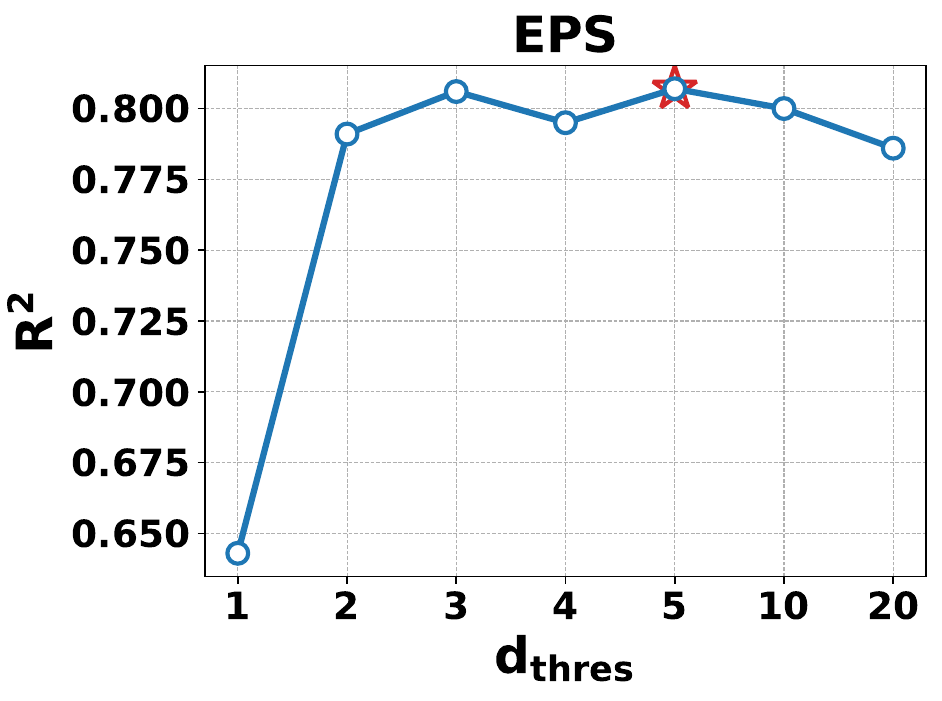}
  \end{subfigure}
  \hfill
  \begin{subfigure}[b]{0.24\textwidth}
    \centering
    \includegraphics[width=\textwidth]{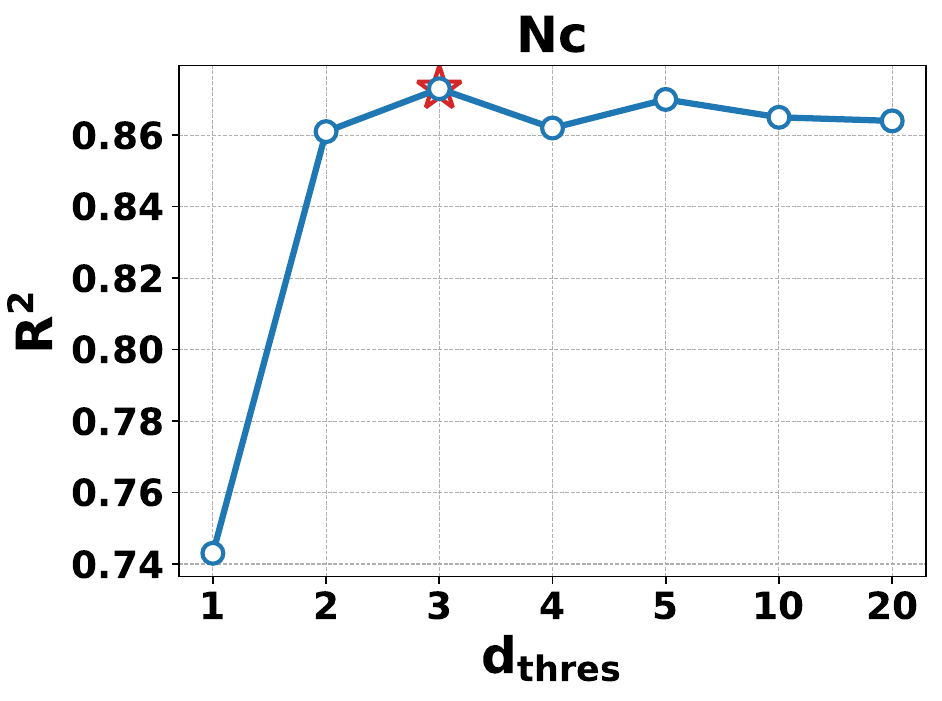}
  \end{subfigure}
  \hfill
  \begin{subfigure}[b]{0.24\textwidth}
    \centering
    \includegraphics[width=\textwidth]{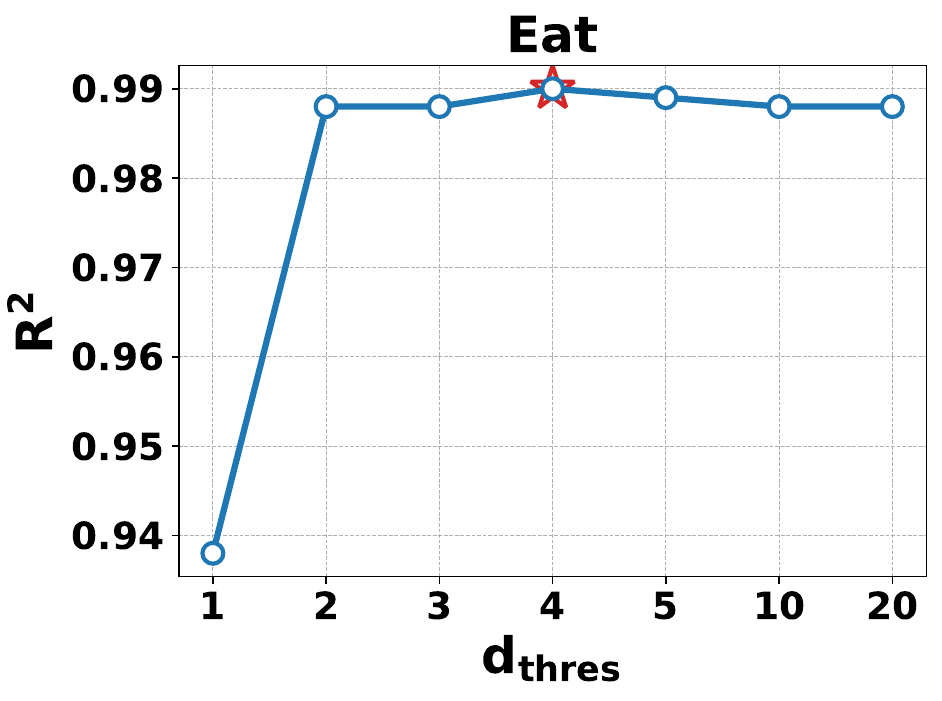}
  \end{subfigure}

  \caption{Influence of $d_{{thres}}$ in localized graph attention. 
  The red star indicates the best $d_{{thres}}$ for each dataset.}
  \label{fig:max_length}
\end{figure*}

\begin{table*}[t]
  \centering
  \caption{$R^2$ Performance of MIPS with different polymer modeling approaches on eight polymer property datasets w./w.o RSIT with $T=5$ trials. 
  The best result for each dataset has been bolded. RSIT Gap: Average performance drop under RSIT.}
  \begin{adjustbox}{width=\textwidth,center}
  \begin{tabular}{@{}c|cccccccc|c@{}}
  \toprule
              & Egc         & Egb         & Eea         & Ei          & Xc           & EPS         & Nc          & Eat         & RSIT Gap \\ \midrule
  Star Keep   & ${0.905}_{\pm 0.003}$ & ${0.931}_{\pm 0.008}$ & ${0.925}_{\pm 0.020}$ & ${0.823}_{\pm 0.061}$ & ${0.404}_{\pm 0.077}$  & ${0.783}_{\pm 0.066}$ & ${0.848}_{\pm 0.053}$ & ${0.979}_{\pm 0.005}$ & --       \\
  Star Remove & ${0.901}_{\pm 0.007}$ & ${0.920}_{\pm 0.013}$ & ${0.919}_{\pm 0.022}$ & ${0.818}_{\pm 0.068}$ & ${0.394}_{\pm 0.049}$  & ${0.773}_{\pm 0.075}$ & ${0.839}_{\pm 0.057}$ & ${0.978}_{\pm 0.006}$ & --       \\
  Star Substitution    & ${0.907}_{\pm 0.005}$ & ${0.932}_{\pm 0.010}$ & ${0.931}_{\pm 0.021}$ & ${0.822}_{\pm 0.051}$ & ${0.427}_{\pm 0.071}$  & $\mathbf{0.808}_{\pm 0.043}$ & ${0.870}_{\pm 0.043}$ & ${0.986}_{\pm 0.002}$ & --       \\
  Star Linking~(ours)   & $\mathbf{0.915}_{\pm 0.007}$ & $\mathbf{0.934}_{\pm 0.011}$ & $\mathbf{0.933}_{\pm 0.016}$ & $\mathbf{0.836}_{\pm 0.056}$ & $\mathbf{0.457}_{\pm 0.092}$  & ${0.806}_{\pm 0.065}$ & $\mathbf{0.873}_{\pm 0.054}$ & $\mathbf{0.988}_{\pm 0.003}$ & --       \\
  \midrule
  Star Keep + RSIT       & ${0.612}_{\pm 0.065}$ & ${0.801}_{\pm 0.043}$ & ${0.755}_{\pm 0.075}$ & ${0.640}_{\pm 0.099}$ & ${-0.165}_{\pm 0.235}$ & ${0.562}_{\pm 0.086}$ & ${0.701}_{\pm 0.073}$ & ${0.952}_{\pm 0.018}$ & $0.218$    \\
  Star Remove + RSIT       & ${0.853}_{\pm 0.011}$ & ${0.849}_{\pm 0.012}$ & ${0.746}_{\pm 0.054}$ & ${0.700}_{\pm 0.053}$ & ${0.088}_{\pm 0.113}$  & ${0.438}_{\pm 0.216}$ & ${0.586}_{\pm 0.195}$ & ${0.909}_{\pm 0.026}$ & $0.172$    \\
  Star Substitution + RSIT       & ${0.854}_{\pm 0.005}$ & ${0.855}_{\pm 0.020}$ & ${0.787}_{\pm 0.083}$ & ${0.716}_{\pm 0.070}$ & ${0.029}_{\pm 0.190}$  & ${0.441}_{\pm 0.242}$ & ${0.573}_{\pm 0.184}$ & ${0.920}_{\pm 0.021}$ & $0.189$    \\
  Star Linking~(ours) + RSIT       & $\mathbf{0.915}_{\pm 0.007}$ & $\mathbf{0.934}_{\pm 0.011}$ & $\mathbf{0.933}_{\pm 0.016}$ & $\mathbf{0.836}_{\pm 0.056}$ & $\mathbf{0.457}_{\pm 0.092}$  & $\mathbf{0.806}_{\pm 0.065}$ & $\mathbf{0.873}_{\pm 0.054}$ & $\mathbf{0.988}_{\pm 0.003}$ & $\mathbf{0.000}$        \\
  \bottomrule  
\end{tabular}
\end{adjustbox} 
\label{table:gt_op_compare}
\end{table*}

\section{Visualization of MIPS Embedding on Downstream Tasks}

we visualize the embedding generated of pre-trained model on each polymer property prediction tasks by UMAP~\cite{McInnes2018UMAP}. 
For the datasets containing more than 1,000 samples, a random subset of 1,000 samples is selected for visualization.
The results, shown in \cref{fig:visualize_embedding}, 
reveal that samples with similar labels tend to form distinct clusters in the embedding space. 
This demonstrates that the output embeddings produced by our model exhibit 
semantic alignment with the respective tasks to a notable degree.

\section{Effectiveness and Robustness of Star Linking in MIPS} 

To assess the impact of the "star linking" strategy in MIPS, 
we replace it with alternative operations, including "star keep", "star remove", and "star substitution"~\cite{Wang2024MMPolymer}, 
as summarized in \cref{table:gt_op_compare}. 
To isolate the effect of the 3D information, the 3D encoder is excluded during this comparison. 
The results show that the "star linking" strategy consistently 
delivers the best performance on seven out of eight datasets. 
Additionally, "star linking" strategy demonstrates robustness against RSIT, 
whereas other three strategies suffer significant performance degradation when subjected to RSIT. 

\section{Influence of Mask Rate in Pre-training}

We investigate the impact of different mask rates $p_{\text{mask}}$ during pre-training. 
The results are presented in \cref{fig:mask_rate}. 
The model demonstrates robust performance across mask rates ranging from 0.1 to 0.7, 
with the optimal average rank achieved at a mask rate of 0.3. 
A mask rate of zero, which is equivalent to no pre-training, results in comparatively lower performance, 
highlighting the importance and effectiveness of pre-training.
Conversely, an excessively high mask rate (e.g., 1.0) 
leads to significant performance degradation, 
yielding results that are even worse than those observed with a zero mask rate.

\section{Influence of $d_{thres}$ on Localized Graph Attention}

We explore the influence of $d_{{thres}}$ in localized graph attention 
by trying different values of $d_{{thres}}$ from $\{1, 2, 3, 4, 5, 10, 20\}$. 
The results are shown in \cref{fig:max_length}. 
Setting $d_{{thres}}$ greater than 1 is crucial to achieving optimal model performance. 
However, excessively large values of $d_{{thres}}$ can result in a slight performance decline, 
which may attribute to the noise from long-range interactions. 
Values of $d_{{thres}}$ between 2 to 5 consistently yield the best performance across all datasets.

\section{Details of Fragment Analysis for Model Interpretation}
We apply DOVE-1~\cite{Wang2025Fragformer} with $100$ vocabulary size as the fragmentation method, 
Additionally, we extend the FragCAM method~\cite{Wang2025Fragformer} to regression tasks to 
determine the relative importance of fragments specific to each dataset.  
The fragmentation $F$ of a monomer from DOVE-1 is a set of sets: 
$F=\{ F_i\}_{i=1}^{N_F}$, where $F_i=\{f_{ij}\}_{j=1}^{N_{F_i}}$, $f_{ij}$ represents an atom.  
To obtain the fragment representation, 
We apply a fragment pooling on the atom representations $\mathbf{X}^{ts}$ to 
get the fragment representation $\{\mathbf{h}_i\}_{i=1}^{N_F}$, where
\[
  \mathbf{h}_i = \sum_{j=1}^{N_{F_i}} \mathbf{X}^{ts}_{f_{ij}}
\].
Next, mean pooling is used to compute the polymer representation: 
\[
  \mathbf{h} = \frac{1}{N_F}\sum\limits_{i=1}^{N_F} h_i
\]
Finally, a linear transformation is applied to produce the prediction: 
\[
\hat{y} = \mathbf{w}^T \mathbf{h}, 
\]
where $\mathbf{w}$ represents the weight of the linear transformation. 
Since the target value have been normalized to zero mean and unit variance, 
no bias term is required in the linear prediction. 
After training, the importance of fragment $F_i$ is calcuated as: 
\[
a_i = \frac{\mathbf{w}^T \mathbf{h}_i}{N_F}.
\]
The relative importance of each fragment is determined by averaging its importance across all samples in the dataset. 
We haved illustrated the structure and contribution scores of the three fragments 
with the highest importance and the three with the lowest importance in the main text.  
For completeness, we also present the results in \cref{fig:fragment_analysis}. 
We further analyze the consistency of fragment importance 
with chemical knowledge of Xc, and Nc datasets.


\paragraph*{Xc} This task aims at predicting the crystallization tendency of polyermers. 
CSC, CCl, and NC=O exhibit higher-than-average contribution scores to crystallization tendency, 
while "c1ccccc1", "C=C" and "c1ccsc1" show lower-than-average scores. 
These observations align with chemical knowledge.
CSC, CCl, and NC=O generally introduce polar or weakly polar interactions. 
These interactions enhance cohesive forces between polymer chains, 
favoring more ordered molecular arrangements and thus promoting crystallization\cite{Richtering2003PolymerP}.
Aromatic~("c1ccccc1") or heterocyclic~("c1ccsc1") ring structures generally display rigidity in polymer chains.
However, this rigidity often impedes regular stacking and crystallization~\cite{Painter2008EssentialsOP}. 
The presence of carbon-carbon double bonds increases structural rigidity and 
introduces conformational constraints into polymer chains, 
Carbon-Carbon double bonds~("C=C")s often lead to cis/trans conformational isomerism and structural irregularities, 
reducing the ability of polymer chains to achieve orderly packing~\cite{Bicerano1996PredictionOP}. 

\paragraph*{Nc} This task focuses on forecasting the refractive index of polymers.  
"c1ccsc1" and "C=S" exhibit higher-than-average contribution scores to the refractive index 
(CN is ignored here due to its much smaller score relative to "c1ccsc1" and "C=S"), 
while "COC", "CCCCC" and "CF" show lower-than-average scores. 
These findings are consistent with chemical knowledge. 
Aromatic systems such as "c1ccsc1" are well-known to exhibit strong electron delocalization and high polarizability, 
thus positively contributing to a higher refractive index~\cite{Higashihara2015RecentPI}. 
Sulfur-containing polymers (polysulfides, polythiourea derivatives, etc.) 
typically display higher refractive indexes due to stronger electronic polarization originating from sulfur atoms~\cite{Watanabe2022TranscendingTT}. 
In contrast, "COC" fragments often provide flexible molecular backbone structures and introduces increased free-volume into polymers, 
leading to reduced polymer density and electronic polarizability~\cite{Mark2007PhysicalPO}. 
"CCCCC" fragments lack polarizable functional groups or aromatic conjugation, 
resulting in relatively low electronic polarization and thus lower refractive indices~\cite{Richtering2003PolymerP}. 
Due to fluorine’s small atomic radius and strong electronegativity, 
fluorine substitution is widely employed to significantly 
lower refractive indices and dielectric constants of polymeric materials~\cite{Amduri2009FromVF}. 
\par 
From the analysis above, we conclude that the model can learn meaningful representations of polymers, 
which reflect the contributions of different functional groups. 

\end{document}